\documentclass[twoside]{article}

\usepackage[accepted]{aistats2025}
\usepackage[round]{natbib}

\def\safedef#1{%
   \ifx#1\undefined
      \expandafter\def\expandafter#1%
   \else
      \errmessage{The \string#1 is defined already}%
      \expandafter\def\expandafter\tmp
   \fi
}

\usepackage{amsthm,amsmath,bbm,amsfonts,amssymb}
\usepackage{dsfont} 
\usepackage{mathtools}
\usepackage{commath}
\usepackage{eqnarray}
\mathtoolsset{showonlyrefs=false}
\allowdisplaybreaks 

\usepackage{xspace}
\usepackage[T1]{fontenc}
\usepackage[utf8]{inputenc}
\usepackage[usenames,dvipsnames]{xcolor} 

\usepackage{hyperref,url}

\usepackage{wrapfig}
\usepackage[export]{adjustbox}
\usepackage{graphicx,tabularx}
\usepackage{multirow,hhline}
\usepackage{booktabs}
\usepackage{pbox} 
\usepackage{algorithm,algorithmic}

\usepackage[export]{adjustbox}

\usepackage{enumitem}
\usepackage[normalem]{ulem}

\newcommand{\kj}[1]{{\color{RedOrange}[KJ: #1]}}

\newcommand{\gray}[1]{{ \color[rgb]{.6,.6,.6} #1 }}

{\color{kjsaved}}%

\usepackage{framed}
\usepackage[most]{tcolorbox}
\definecolor{kjgray}{rgb}{.7,.7,.7}

\newtheoremstyle{kjstyle}
{1ex} 
{\topsep} 
{\itshape} 
{} 
{\bfseries} 
{.} 
{.5em} 
{} 

\newtheoremstyle{kjstyle2}
{.0em} 
{.0em} 
{\itshape} 
{} 
{\bfseries} 
{.} 
{.5em} 
{} 

\newtheoremstyle{kjstylenoitalic}
{1ex} 
{\topsep} 
{} 
{} 
{\bfseries} 
{.} 
{.5em} 
{} 

\tcbset{kjboxstyle/.style={title={},breakable,colback=white,enhanced jigsaw,boxrule=1.3pt,sharp corners,colframe=kjgray,boxsep=0pt,coltitle={black},attach title to upper={},left=.8ex,bottom=.4em}}    

\tcbset{kjboxstylec/.style={title={},breakable,colback=white,enhanced jigsaw,boxrule=1.3pt,sharp corners,colframe=kjgray,boxsep=0pt,coltitle={black},attach title to upper={},before skip=1.5ex,left=.8ex,bottom=.4em,top=0.4em,enlarge top by=-0.3em,enlarge bottom by=-0.3em}}    

\newtheorem{theorem}{Theorem}

\theoremstyle{kjstyle}
\AtBeginEnvironment{thm}{\begin{tcolorbox}[kjboxstyle]}%
  \AtEndEnvironment{thm}{\end{tcolorbox}}%
\theoremstyle{kjstyle}
\AtBeginEnvironment{lem}{\begin{tcolorbox}[kjboxstyle]}%
  \AtEndEnvironment{lem}{\end{tcolorbox}}%
\theoremstyle{kjstyle}
\AtBeginEnvironment{prop}{\begin{tcolorbox}[kjboxstyle]}%
  \AtEndEnvironment{prop}{\end{tcolorbox}}%
\theoremstyle{kjstyle}
\AtBeginEnvironment{cor}{\begin{tcolorbox}[kjboxstyle]}%
  \AtEndEnvironment{cor}{\end{tcolorbox}}%

\theoremstyle{kjstyle}
\AtBeginEnvironment{defn}{\begin{tcolorbox}[kjboxstyle]}%
  \AtEndEnvironment{defn}{\end{tcolorbox}}%

\theoremstyle{kjstyle}
\AtBeginEnvironment{ass}{\begin{tcolorbox}[kjboxstyle]}%
  \AtEndEnvironment{ass}{\end{tcolorbox}}%

\theoremstyle{kjstyle}
\AtBeginEnvironment{conj}{\begin{tcolorbox}[kjboxstyle]}%
  \AtEndEnvironment{conj}{\end{tcolorbox}}%

\theoremstyle{kjstyle}
\AtBeginEnvironment{question}{\begin{tcolorbox}[kjboxstyle]}%
  \AtEndEnvironment{question}{\end{tcolorbox}}%

\theoremstyle{kjstylenoitalic}\newtheorem{remark}{Remark}
\AtBeginEnvironment{remark}{\begin{tcolorbox}[kjboxstyle]}%
  \AtEndEnvironment{remark}{\end{tcolorbox}}%
\theoremstyle{kjstylenoitalic}
\AtBeginEnvironment{openp}{\begin{tcolorbox}[kjboxstyle]}%
  \AtEndEnvironment{openp}{\end{tcolorbox}}%

\theoremstyle{kjstylenoitalic}
\AtBeginEnvironment{ex}{\begin{tcolorbox}[kjboxstyle]}%
  \AtEndEnvironment{ex}{\end{tcolorbox}}%

\theoremstyle{kjstylenoitalic}

\usepackage{mdframed}
\usepackage{lipsum}
\definecolor{kjgray}{rgb}{.7,.7,.7}
\makeatletter



\makeatletter
\renewcommand{\paragraph}{%
  \@startsection{paragraph}{4}%
  {\z@}{0.50ex \@plus 1ex \@minus .2ex}{-1em}%
  {\normalfont\normalsize\bfseries}%
}
\makeatother

\usepackage[customcolors,shade]{hf-tikz} 

\usepackage{tabularx} 
\newcolumntype{P}[1]{>{\centering\arraybackslash}p{#1}}
\newcolumntype{M}[1]{>{\centering\arraybackslash}m{#1}}

\def\ddefloop#1{\ifx\ddefloop#1\else\ddef{#1}\expandafter\ddefloop\fi}

\def\ddef#1{\expandafter\def\csname #1#1\endcsname{\ensuremath{\mathbb{#1}}}}
\ddefloop ABCDFGHIJKLMNORSTUWXYZ\ddefloop 

\def\ddef#1{\expandafter\def\csname c#1\endcsname{\ensuremath{\mathcal{#1}}}}
\ddefloop ABCDEFGHIJKLMNOPQRSTUVWXYZ\ddefloop

\def\ddef#1{\expandafter\def\csname b#1\endcsname{\ensuremath{{\mathbf{#1}}}}}
\ddefloop ABCDEFGHIJKLMNOPQRSTUVWXYZ\ddefloop  
\def\ddef#1{\expandafter\def\csname b#1\endcsname{\ensuremath{{\boldsymbol{#1}}}}}
\ddefloop abcdeghijklmnopqrtsuvwxyz\ddefloop  

\def\ddef#1{\expandafter\def\csname h#1\endcsname{\ensuremath{\hat{#1}}}}
\ddefloop ABCDEFGHIJKLMNOPQRSTUVWXYZabcdefghijklmnopqrsuvwxyz\ddefloop 
\def\ddef#1{\expandafter\def\csname hc#1\endcsname{\ensuremath{\hat{\mathcal{#1}}}}}
\ddefloop ABCDEFGHIJKLMNOPQRSTUVWXYZ\ddefloop
\def\ddef#1{\expandafter\def\csname hb#1\endcsname{\ensuremath{\hat{\mathbf{#1}}}}}
\ddefloop ABCDEFGHIJKLMNOPQRSTUVWXYZ\ddefloop %
\def\ddef#1{\expandafter\def\csname hb#1\endcsname{\ensuremath{\hat{\boldsymbol{#1}}}}}
\ddefloop abcdefghijklmnopqrstuvwxyz\ddefloop %

\def\ddef#1{\expandafter\def\csname t#1\endcsname{\ensuremath{\tilde{#1}}}}
\ddefloop ABCDEFGHIJKLMNOPQRSTUVWXYZabcdefgijklmnpqtsuvwxyz\ddefloop 
\def\ddef#1{\expandafter\def\csname tc#1\endcsname{\ensuremath{\tilde{\mathcal{#1}}}}}
\ddefloop ABCDEFGHIJKLMNOPQRSTUVWXYZ\ddefloop
\def\ddef#1{\expandafter\def\csname tb#1\endcsname{\ensuremath{\tilde{\mathbf{#1}}}}}
\ddefloop ABCDEFGHIJKLMNOPQRSTUVWXYZ\ddefloop
\def\ddef#1{\expandafter\def\csname tb#1\endcsname{\ensuremath{\tilde{\boldsymbol{#1}}}}}
\ddefloop abcdefghijklmnopqrstuvwxyz\ddefloop %

\def\ddef#1{\expandafter\def\csname bar#1\endcsname{\ensuremath{\bar{#1}}}}
\ddefloop ABCDEFGHIJKLMNOPQRSTUVWXYZabcdefghijklmnopqrtsuvwxyz\ddefloop
\def\ddef#1{\expandafter\def\csname barc#1\endcsname{\ensuremath{\bar{\mathcal{#1}}}}}
\ddefloop ABCDEFGHIJKLMNOPQRSTUVWXYZ\ddefloop
\def\ddef#1{\expandafter\def\csname barb#1\endcsname{\ensuremath{\bar{\mathbf{#1}}}}}
\ddefloop ABCDEFGHIJKLMNOPQRSTUVWXYZ\ddefloop
\def\ddef#1{\expandafter\def\csname barb#1\endcsname{\ensuremath{\bar{\boldsymbol{#1}}}}}
\ddefloop abcdefghijklmnopqrstuvwxyz\ddefloop %

\def\ddef#1{\expandafter\def\csname war#1\endcsname{\ensuremath{\overline{#1}}}}
\ddefloop ABCDEFGHIJKLMNOPQRSTUVWXYZabcdefghijklmnopqrtsuvwxyz\ddefloop
\def\ddef#1{\expandafter\def\csname warc#1\endcsname{\ensuremath{\overline{\mathcal{#1}}}}}
\ddefloop ABCDEFGHIJKLMNOPQRSTUVWXYZ\ddefloop
\def\ddef#1{\expandafter\def\csname warb#1\endcsname{\ensuremath{\overline{\mathbf{#1}}}}}
\ddefloop ABCDEFGHIJKLMNOPQRSTUVWXYZ\ddefloop
\def\ddef#1{\expandafter\def\csname warb#1\endcsname{\ensuremath{\overline{\boldsymbol{#1}}}}}
\ddefloop abcdefghijklmnopqrstuvwxyz\ddefloop %

\def\dt{\delta}
\def\gam{\gamma}
\def\lam{\lambda}

\def\eps{\varepsilon}
\def\epsilon{\varepsilon}

\usepackage{pgffor}
\def\greeksymbols{alpha,beta,gamma,gam,delta,dt,eps,epsilon,zeta,eta,theta,th,iota,kappa,kap,lambda,lam,mu,nu,xi,pi,rho,sigma,sig,tau,phi,chi,psi,omega,om,Gamma,Gam,Delta,Dt,Theta,Th,Lambda,Lam,Pi,Sigma,Sig,Phi,Psi,Omega,Om}
\def\greeksymbolsnoeta{alpha,beta,gamma,gam,delta,dt,eps,epsilon,zeta,theta,th,iota,kappa,kap,lambda,lam,mu,nu,xi,pi,rho,sigma,sig,tau,phi,chi,psi,omega,om,Gamma,Gam,Delta,Dt,Theta,Th,Lambda,Lam,Pi,Sigma,Sig,Phi,Psi,Omega,Om} 

\foreach \x in \greeksymbolsnoeta{\expandafter\xdef\csname b\x\endcsname{\noexpand\ensuremath{\noexpand\boldsymbol{\csname \x\endcsname}}}}

\foreach \x in \greeksymbols{\expandafter\xdef\csname h\x\endcsname{\noexpand\ensuremath{\noexpand\hat{\csname \x\endcsname}}}}
\foreach \x in \greeksymbolsnoeta{\expandafter\xdef\csname hb\x\endcsname{\noexpand\ensuremath{\noexpand\hat{\noexpand\boldsymbol{ \csname \x\endcsname}}}}}

\foreach \x in \greeksymbols{\expandafter\xdef\csname bar\x\endcsname{\noexpand\ensuremath{\noexpand\bar{\csname \x\endcsname}}}}
\foreach \x in \greeksymbolsnoeta{%
\expandafter\xdef\csname barb\x\endcsname{\noexpand\ensuremath{\noexpand\bar{\noexpand\boldsymbol{ \csname \x\endcsname}}}}
}

\foreach \x in \greeksymbols{\expandafter\xdef\csname t\x\endcsname{\noexpand\ensuremath{\noexpand\tilde{\csname \x\endcsname}}}}
\foreach \x in \greeksymbolsnoeta{\expandafter\xdef\csname tb\x\endcsname{\noexpand\ensuremath{\noexpand\tilde{\noexpand\boldsymbol{ \csname \x\endcsname}}}}}

\providecommand{\normz}[2][-1]{
\ensuremath{\mathinner{
\ifthenelse{\equal{#1}{-1}}{ 
\!\left\|#2\right\|}{}
\ifthenelse{\equal{#1}{0}}{ 
\|#2\|}{}
\ifthenelse{\equal{#1}{1}}{ 
\bigl\|#2\bigr\|}{}
\ifthenelse{\equal{#1}{2}}{ 
\Bigl\|#2\Bigr\|}{}
\ifthenelse{\equal{#1}{3}}{ 
\biggl\|#2\biggr\|}{}
\ifthenelse{\equal{#1}{4}}{ 
\Biggl\|#2\Biggr\|}{}
}} 
}  

\providecommand{\floor}[2][-1]{
\ensuremath{\mathinner{
\ifthenelse{\equal{#1}{-1}}{ 
\!\left\lfloor#2\right\rfloor}{}
\ifthenelse{\equal{#1}{0}}{ 
\lfloor#2\rfloor}{}
\ifthenelse{\equal{#1}{1}}{ 
\!\bigl\lfloor#2\bigr\rfloor}{}
\ifthenelse{\equal{#1}{2}}{ 
\!\Bigl\lfloor#2\Bigr\rfloor}{}
\ifthenelse{\equal{#1}{3}}{ 
\!\biggl\lfloor#2\biggr\rfloor}{}
\ifthenelse{\equal{#1}{4}}{ 
\!\Biggl\lfloor#2\Biggr\rfloor}{}
}} 
}

\providecommand{\ceil}[2][-1]{
\ensuremath{\mathinner{
\ifthenelse{\equal{#1}{-1}}{ 
\!\left\lceil#2\right\rceil}{}
\ifthenelse{\equal{#1}{0}}{ 
\lceil#2\rceil}{}
\ifthenelse{\equal{#1}{1}}{ 
\!\bigl\lceil#2\bigr\rceil}{}
\ifthenelse{\equal{#1}{2}}{ 
\!\Bigl\lceil#2\Bigr\rceil}{}
\ifthenelse{\equal{#1}{3}}{ 
\!\biggl\lceil#2\biggr\rceil}{}
\ifthenelse{\equal{#1}{4}}{ 
\!\Biggl\lceil#2\Biggr\rceil}{}
}} 
}

\newcommand{\fr}[2]{ { \frac{#1}{#2} }}

\def\cd{\cdot}

\def\rarrow{\ensuremath{\rightarrow}} 
\def\sm{{\ensuremath{\setminus}}}

\definecolor{mygrn}{rgb}{0,.8,0}
\definecolor{myred}{rgb}{.8,0,0}

\DeclareMathOperator{\EE}{\mathbb{E}} 

\DeclareMathOperator{\PP}{\mathbb{P}}

\DeclareMathOperator{\Var}{{\mathrm{Var}}}

\DeclareMathOperator*{\argmax}{arg~max}

\DeclarePairedDelimiterX{\inp}[2]{\langle}{\rangle}{#1, #2}

\newcommand\declareop[3]{%
  \newcommand#1{%
    \mskip\muexpr\medmuskip*#2\relax
    {#3}%
    \mskip\muexpr\medmuskip*#2\relax
}}
\declareop\capprox{1}{{\sr{\const}{\approx}}} 
\declareop\logapprox{1}{{\sr{\mathsf{log}}{\approx}}} 

\newcommand{\lsim}{\mathop{}\!\lesssim}
\newcommand{\gsim}{\mathop{}\!\gtrsim}

\def\const{\mathsf{const}}

\usepackage{pifont}
\newcommand{\sr}{\stackrel}

\makeatletter
\newcommand{\vast}{\bBigg@{3}}
\newcommand{\Vast}{\bBigg@{4}}
\makeatother

\newenvironment{talign*}
 {\let\displaystyle\textstyle\csname align*\endcsname}
 {\endalign}

\def\chrulefill{\leavevmode\leaders\hrule height 0.7ex depth \dimexpr0.4pt-0.7ex\hfill\kern0pt}

\renewcommand{\cite}{\citep}

\usepackage[normalem]{ulem}
\newcommand{\stkout}[1]{\ifmmode\text{\sout{\ensuremath{#1}}}\else\sout{#1}\fi}

\newcommand{\guide}[1]{{\color{Violet}[[#1]]}}

\newenvironment{phaseout}
{\color[rgb]{.6,.6,.6}}
{\color[rgb]{.6,.6,.6}}

\ifx\theorem\undefined
\newtheorem{theorem}{Theorem}
\fi
\ifx\example\undefined
\newtheorem{example}{Example}
\fi
\ifx\property\undefined
\newtheorem{property}{Property}
\fi
\ifx\lemma\undefined
\newtheorem{lemma}[theorem]{Lemma}
\fi
\ifx\proposition\undefined
\newtheorem{proposition}[theorem]{Proposition}
\fi
\ifx\remark\undefined
\newtheorem{remark}[theorem]{Remark}
\fi
\ifx\corollary\undefined
\newtheorem{corollary}[theorem]{Corollary}
\fi
\ifx\definition\undefined
\newtheorem{definition}[theorem]{Definition}
\fi
\ifx\conjecture\undefined
\newtheorem{conjecture}[theorem]{Conjecture}
\fi
\ifx\fact\undefined
\newtheorem{fact}[theorem]{Fact}
\fi
\ifx\claim\undefined
\newtheorem{claim}[theorem]{Claim}
\fi
\ifx\assum\undefined
\newtheorem{assumption}[theorem]{Assumption}
\fi

\usepackage{subcaption}

\newif\ifFINAL
\FINALtrue 

\ifFINAL
    
    \def\guide#1{}
    \def\gray#1{}
    \def\kj#1{}
\else
    \usepackage[inline]{showlabels}
    
\fi

\usepackage[toc,page,header]{appendix}
\usepackage{minitoc}
\usepackage{silence}
\newcommand{\tilcO}{\tilde \cO}

\newcommand{\ep}[1]{\exp\del{{#1}}}

\newcommand{\qgam}{\fr{1}{3}}
\newcommand{\fgaminv}{\fr{2}{3}}
\newcommand{\fgam}{\fr{3}{2}}
\newcommand{\cs}{\fr{1}{6}}
\newcommand{\cstar}{\fr{1}{6}}
\newcommand{\cplus}{\fr{4}{3}}
\newcommand{\cplusq}{\fr{8}{3}}

\newcommand{\muhat}{\hat{\mu}}
\newcommand{\sigmahat}{\hat{\sigma}}

\newcommand{\Kstar}{K^{*}}

\newcommand{\muhatBcal}{\hat{\mu}^{\mathcal{B}}}

\newcommand{\Nmax}{N_{\max}}

\newcommand{\hmuhaver}{\hmu^{\textrm{HAVER}}}
\newcommand{\hmume}{\hmu^{\textrm{LEM}}}

    \newcommand{\hmumlcb}{\hmu^{\textrm{MLCB}}}

\newcommand{\hmuB}{\hmu^{\mathcal{B}}}
\newcommand{\cBs}{\cB^{*}}
\newcommand{\cBp}{\cB^{+}}

\newcommand{\difeps}{2\eps \dif \eps}

\newcommand{\idt}[1]{\mathbbm{1}\left \{#1 \right \}}

\newcommand{\tagcmt}[1]{\tag{\text{#1}}}

\DeclareMathOperator{\MSE}{MSE}
\DeclareMathOperator{\Bias}{Bias}

\newcommand{\Bcals}{\mathcal{B}^*}
\newcommand{\Bcalp}{\mathcal{B}^+}
\newcommand{\Scal}{\mathcal{S}}

\theoremstyle{kjstyle}\newtheorem*{thm*}{Theorem}
\AtBeginEnvironment{thm*}{\begin{tcolorbox}[kjboxstyle]}%
  \AtEndEnvironment{thm*}{\end{tcolorbox}}%
\theoremstyle{kjstyle}\newtheorem*{lem*}{Lemma}
\AtBeginEnvironment{lem*}{\begin{tcolorbox}[kjboxstyle]}%
  \AtEndEnvironment{lem*}{\end{tcolorbox}}%
\theoremstyle{kjstyle}\newtheorem*{cor*}{Corollary}
\AtBeginEnvironment{cor*}{\begin{tcolorbox}[kjboxstyle]}%
  \AtEndEnvironment{cor*}{\end{tcolorbox}}%

\begin{document}
\textfloatsep=.5em
\setlength{\abovedisplayskip}{3pt}%
\setlength{\belowdisplayskip}{4pt}%
\setlength{\abovedisplayshortskip}{3pt}%
\setlength{\belowdisplayshortskip}{4pt}

%

%

\doparttoc 
\faketableofcontents 

\runningtitle{Instance-Dependent Error Bounds for Maximum Mean Estimation and Applications}

\twocolumn[

\aistatstitle{HAVER: Instance-Dependent Error Bounds for Maximum Mean Estimation and Applications to Q-Learning and Monte Carlo Tree Search}

\aistatsauthor{ Tuan Ngo Nguyen \And Jay Barrett
 \And Kwang-Sung Jun }

\aistatsaddress{ University of Arizona \\ \texttt{tnguyen9210@arizona.edu} \And Cornell University \\ \texttt{jab864@cornell.edu} \And University of Arizona \\  \texttt{kjun@cs.arizona.edu} }

]

\begin{abstract}
  We study the problem of estimating the \emph{value} of the largest mean among $K$ distributions via samples from them (rather than estimating \emph{which} distribution has the largest mean), which arises from various machine learning tasks including Q-learning and Monte Carlo Tree Search (MCTS). 
  While there have been a few proposed algorithms, their performance analyses have been limited to their biases rather than a precise error metric. 
  In this paper, we propose a novel algorithm called HAVER (Head AVERaging) and analyze its mean squared error.
  Our analysis reveals that HAVER has a compelling performance in two respects.
  First, HAVER estimates the maximum mean as well as the oracle who knows the identity of the best distribution and reports its sample mean.
  Second, perhaps surprisingly, HAVER exhibits even better rates than this oracle when there are many distributions near the best one.
  Both of these improvements are the first of their kind in the literature, and we also prove that the naive algorithm that reports the largest empirical mean does not achieve these bounds.  
  Finally, we confirm our theoretical findings via numerical experiments where we implement HAVER in bandit, Q-learning, and MCTS algorithms. In these experiments, HAVER consistently outperforms the baseline methods, demonstrating its effectiveness across different applications.
\end{abstract}

\section{Introduction}

We consider the problem of estimating the maximum mean value among $K$ distributions based on samples from them, which we call the \emph{maximum mean estimation} (MME) problem.
Specifically, for each $i\in[K]:=\{1,\ldots,K\}$, we are passively given samples $X_{i,1},\ldots,X_{i,N_i}$ from the $i$-th distribution $\nu_i$.
The learner must estimate the largest mean $\max_{i\in[K]} \EE_{X\sim\nu_i} [X]$ as accurately as possible.
The MME problem arises from various machine learning tasks.
In Q-learning at each time step, an agent updates its state-action value estimates $\hQ(s, a)$ based on the observed reward and the estimated value of the next state $s'$.
The latter requires an accurate estimate of the largest state-action value $\max_{a} Q^*(s', a)$ where $Q^*$ is the true state-action value.
An inaccurate estimator for the expectation can adversely impact the learning process. 
Similarly, in Monte Carlo tree search~\citep{coulom06efficient,kocsis06bandit}, one faces to estimate the value of each node at a non-leaf node, which is the maximum value of its children node.
Accurate estimation of the largest value is paramount to quickly filter out unpromising nodes, which allows budget-efficient identification of the best action.

Arguably, the simplest estimator is to take the largest empirical mean (LEM). 
However, LEM has a positive bias, which can be detrimental to its accuracy when the number of samples is small or the number of distributions $K$ is large.
To overcome such an issue, there have been a number of studies on the problem of estimating the maximum mean value. 
The earliest work we are aware of is \citet{vanhasselt10double,vanhasselt13estimating}, where the authors propose a double estimator (DE) based on sample splitting that is guaranteed to be negatively biased.
In another work of \citet{lan20maxmin}, the authors propose the maxmin estimator that also uses a sample splitting but chooses the largest of the smallest estimator from multiple sampling buckets.
Another idea is to consider a weighted estimator (WE) that computes a weighted average over the empirical means~\cite{tesauro10bayesian,deramo16estimating}.
While these studies report some success in downstream applications, their theoretical justification is limited to characterizing the variance or the direction of the bias rather than precise error metrics such as mean squared error (MSE).
Furthermore, to our knowledge, no studies have ever shown that their proposed methods enjoy orderwise better MSE compared to LEM.
We discuss more on related work in Section~\ref{sec:related}.

\setlist{nolistsep} 
\setlist[itemize]{topsep=.5pt,itemsep=0pt,parsep=2pt}
\setlist[enumerate]{topsep=.5pt,itemsep=0pt,parsep=2pt}

\def\Done{\textbf{D1}\xspace}
\def\Dtwo{\textbf{D2}\xspace}

In this paper, we focus on the MSE as the error metric of interest.
What are the good rates of MSE in the MME problem?
We identify two desiderata centered around a reference estimator that we call \textit{the oracle} who knows the identity of the true best arm and thus reports its sample mean:
\begin{itemize}[leftmargin=3em]
    \item[\Done:] Can we achieve a worst-case MSE bound that is as good as the oracle?
    \item[\Dtwo:] Can we achieve an instance-dependent MSE bound that can be strictly better than the oracle?
\end{itemize}
As a sanity check, we first show that LEM fails to satisfy either of these criteria. 
This means that the two criteria above are not trivial to achieve.

As our main contribution, we propose a novel estimator called HAVER (Head AVERaging).
Our analysis shows that HAVER achieves not only \Done but also, perhaps surprisingly, \Dtwo.  In particular, we derive a generic instance-dependent upper bound on the MSE of HAVER.
This generic bound can further be upperbound by an instance-independent quantity to satisfy \Done.
For \Dtwo, we evaluate the generic upper bound of HAVER under several practical instances including instances whose suboptimality gaps or sample counts vary with specific rates. 
Finally, we conduct empirical studies across various settings, including bandits and Q-learning environments, to show that HAVER consistently outperforms prior methods.

Our theoretical analysis provides a convenient framework for analyzing MSE of maximum mean estimators by leveraging the equality of expectation with the tail bound of nonnegative random variables; i.e., 
\begin{align}\label{eq:expectation-equivalence}
    \EE [(X-\EE[X])^2] = \int_{\eps=0}^\infty \PP((X-\EE[X])^2 > \eps) \dif \eps
\end{align}
which allows us to leverage tight concentration inequalities along with careful event decomposition.
We believe our analysis framework can be of independent interest to the machine learning research community.


\section{Problem Definition and Preliminaries}

\paragraph{Notations.}
Throughout the paper, $c_1$ and $c_2$ are a universal and positive constant, which may have different values for different expressions. 
Both $\tilde\Theta(\cdot)$, $\tilcO(\cdot)$, and ``$\scriptscriptstyle \lsim$'' omits logarithmic factors on $K$ and $N$, but not $1/\dt$. 
We denote $A_{i},A_{i+1}, \ldots, A_{j}$ by $A_{i:j}$.

\paragraph{The maximum mean estimation (MME) problem.}
In the MME problem, we are given $K$ distributions, which we call \textit{arms}, hereafter, following the vocabulary of multi-armed bandits~\cite{thompson33onthelikelihood,lattimore20bandit}.
We denote by $\nu_i$ the distribution of arm $i$ and define $\mu_i = \EE_{X\sim \nu_i}[X]$.
The learner is given $N_i$ observations denoted by $X_{i,1},\ldots,X_{i,N_i}$ that follow $\nu_i$ for every arm $i\in[K]$. 
We define $D := \{X_{1,1:N_1}, \ldots, X_{K,1:N_K}\}$ as the aggregate set of all samples from all the arms.
Note that this problem is a passive sampling setting in that $N_i$ is arbitrarily given from the environment rather than being chosen by the learner.
The goal of the learner is to estimate the largest mean among the $K$ arms, i.e., $\max_{i\in[K]} \mu_i$, as accurately as possible.
Without loss of generality, we assume that the arms are ordered in decreasing order of their mean, i.e., $\mu_1 \ge \mu_2 \ge \ldots \ge \mu_K$.
Thus, the maximum mean is $\mu_1$.
This assumption is for notational convenience only, and the learner is not aware of this ordering.
We define $\Delta_i = \mu_1 - \mu_i$ to be the suboptimality gap of arm $i\in[K]$.

Let $\hmu$ be the estimator computed by a maximum mean estimation algorithm.
As a performance measure, we choose the mean squared error (MSE) metric:
\begin{align*}
    \MSE(\hmu) = \EE\sbr{\del{\hmu - \mu_1}^2}~,
\end{align*}
which is a standard measure of error in statistics.

\begin{definition}[Sub-Gaussian distribution]\label{def:subgaussian} 
  If $X$ is $\sigma$-sub-Gaussian, then for any $\eps \ge 0$,
  \begin{align*}
      \PP\del{X \ge \eps} \le \exp\del{-\fr{\eps^2}{2\sigma^2}}
  \end{align*}
\end{definition}

We assume that the distributions are sub-Gaussian as follows.
\begin{assumption}[Sub-Gaussian distributions]\label{ass:subgaussian} 
  For each arm $i \in [K]$, the distribution $\nu_i$ is 1-sub-Gaussian, and this sub-Gaussian parameter is known to the learner.
\end{assumption}
Note that the choice of 1 for the sub-Gaussian parameter is for brevity only, and all our results can be extended to the generic $\sigma^2$-sub-Gaussian assumption.

We also make the following standard assumption in the maximum mean estimation literature~\citep{vanhasselt10double,vanhasselt13estimating,deramo16estimating,lan20maxmin}.
\begin{assumption}[i.i.d samples]\label{ass:iid} 
  For each arm $i \in [K]$, each sample $X_{i,j}$ is drawn i.i.d from the distribution $\nu_i$, $\forall j \in [N_i]$.
\end{assumption}



\paragraph{Largest Empirical Mean (LEM) estimator.}
The most immediate and intuitive estimator for the MME problem is the largest empirical mean (LEM) estimator that returns
\begin{align*}
  \hmume := \max_{i\in[K]} \hmu_i
\end{align*}
where $\hmu_i := \frac1{N_i}\sum_{j=1}^{N_i} X_{i,j}$ is the empirical mean.

We analyze the MSE of LEM in the following theorem.
\begin{theorem}\label{thm:lem_bound}
  LEM achieves
  \begin{align*}
    \MSE(\hmume) = \cO \del{\fr{\log(2K)}{\min_iN_i}}~.
  \end{align*}
\end{theorem}
Our bound is improved upon the prior work of \citet{vanhasselt13estimating} in the following sense.
\citet{vanhasselt13estimating} provides separate upper bounds for bias and variance as follows:
\begin{align*}
  \Bias^2(\hmume) \le \fr{K-1}{K}\sum_{i = 1}^{K}\fr{1}{N_i}, \hspace{0.5em} \Var(\hmume) \le \sum_{i = 1}^{K}\fr{1}{N_i}.
\end{align*}
One can derive an MSE bound from this since MSE is the sum of the squared bias and the variance, which becomes
\begin{align*}
    \MSE(\hmume) = \cO\del{\sum_{i = 1}^{K}\fr{1}{N_i}}~.
\end{align*}
When all sample sizes are equal, i.e., $N_i = N, \forall i\in[K]$, our bound is a factor of $K/\log(2K)$ tighter. 

\def\AE{{\normalfont{\text{AE}}}}

  It is natural to ask what properties an ideal estimator can possess and how LEM measures up to these standards.
  First, consider an oracle that has prior knowledge of the optimal arm, which has the maximum mean $\mu_1$. The oracle would base its estimation only on samples drawn from this arm. Consequently, the MSE of the oracle would be $\Theta\del{\fr{1}{N_1}}$ which we refer to as the \textbf{oracle rate}. Compared to this oracle rate, LEM does not perform as reliably. We show that, in a two-arms instance where both arms follow Gaussian distributions with the same mean, the MSE of LEM has a lower bound proportional to $\frac{1}{\min_{i \in [2]} N_i}$, which is worse than the oracle rate. This implies that LEM cannot always obtain the oracle rate in general. In certain applications, the number of samples drawn from suboptimal arms can be significantly smaller compared to those from the optimal arm. As a result, using LEM in such scenarios can lead to arbitrarily large errors. For instance, in Q-learning, the agent gradually learns to favor the optimal action, leading to disproportionally few samples from the suboptimal actions. This slows down the internal maximum mean estimation and, consequently, the learning process.

  Second, consider an extreme problem instance where all arms share the same mean, i.e., $\mu_1 = \cdots = \mu_K$, and the number of samples across all arms are identical $\forall i,\, N_i = N$. In this instance, LEM achieves an MSE bound that is proportional to $\frac{1}{N_1}$, which matches the oracle rate. However, can we do even better? Average estimator (AE) is another estimator that is used in the standard Monte Carlo tree search (MCTS) algorithm called UCT~\citep{kocsis06bandit}:
  \begin{align*}
    \hmu^{\AE} := \fr1K \sum_{i=1}^K \sum_{j=1}^{N_i} X_{i,j}~.
  \end{align*}
  In this extreme case, the AE achieves an MSE bound that is proportional to $\fr{1}{KN}$, which is a factor of $K$ improvement over the oracle rate. 
  
  In summary, a good estimator should (i) perform as well as the oracle rate, and (ii) achieve acceleration over the oracle rate in special instances. 
  This raises the following question: does there exist an estimator that satisfies both properties? In the next section, we answer this question in the affirmative by proposing a novel estimator called HAVER.

\section{Warmup: The Oracle Rate via Maximum Lower Confidence Bound}

We begin by investigating whether the maximum lower bound confidence bound estimator (MLCB) can achieve the desired properties, which is known as the \textit{pessimistic principle} and has been useful for estimating \textit{which arm} has the highest mean in various settings including off-policy reinforcement learning~\cite{jin21is} which is different from estimating the value of the best arm.

For every arm $i$, we define its empirical mean as ${\hmu_i := \frac1{N_i}\sum_{j=1}^{N_i} X_{i,j}}$ and  its confidence width as $\gam_i := \sqrt{\fr{16}{N_i}\log\del{\del{\fr{KT}{N_i}}^2}}$, where ${T = \sum_{j \in [K]}N_j}$. 
We propose the MLCB algorithm, which chooses an arm with the highest lower confidence bound, defined as 
$$\hr := \argmax_{i \in [K]} \hmu_i - \gam_i$$
and outputs
$$ \hmumlcb := \hmu_{\hr}~.$$
In the following theorem, we show that MLCB achieves the oracle rate up to logarithmic factors. 
\begin{theorem}\label{thm:maxlcb}
  MLCB achieves
  \begin{align*}
    &\MSE(\hmumlcb) = \tilcO \del{\fr{1}{N_1}}
  \end{align*}
\end{theorem}
We remark that, though the algorithmic principle was inspired by the well-known pessimism, adapting it to our problem of estimating the \textit{value} of the best arm with respect to the MSE metric is not straightforward since the existing work using pessimism cares about a high probability guarantee with a given specified confidence level $\dt$, so setting $\gam_i$ as $\sqrt{\fr2{N_i}\log(K/\delta)}$ is enough.
This is a matter of considering the worst-case behavior of the estimator under high probability events, and invoking concentration bounds with the specified level is enough.
In contrast, analyzing MSE requires studying the tail of the estimator at all the confidence levels (see \eqref{eq:expectation-equivalence}), which requires carefully setting the right confidence width $\gam_i$ to account for the event where the confidence bound fails to hold.

\section{Better than the Oracle: HAVER (Head AVERaging)}

\begin{algorithm}[t]
  \caption{HAVER}\label{algo:haver}
  \begin{algorithmic}
    \STATE{\textbf{Input:} A set of $K$ arms with samples $\cbr{X_{i, 1:N_i}}_{i=1}^{K}$. \\
}
    \STATE{
 \vspace{.4em}
 For each arm $i \in [K]$, compute its empirical mean: \\
\qquad $\hmu_i = \fr{1}{N_i}\sum_{j=1}^{N_i}X_{i,j}$. \\
\vspace{.4em}
 Find the pivot arm $\hr$:  \\
\qquad $\hr = \displaystyle \argmax_{i \in [K]} \muhat_i - \gam_{i}$ \\
\qquad where $\gam_i = \sqrt{\frac{18}{N_{i}}\log\del{\del{\fr{KS}{N_i}}^{4}}}$, \\
\qquad and $S = N_{\max}\sum_{j \in [K]}N_j$ \\
\vspace{.4em}
Form a candidate set $\cB$: \\
\qquad $\cB = \cbr{i \in [K]: \muhat_{i} \ge \muhat_{\hr} - \gam_{\hr},\, \gam_{i} \le \fgam \gam_{\hr}}$ \\
\vspace{.4em}
 Compute the weighted average of the empirical means of candidate arms in set $\cB$: \\
\qquad $\muhatBcal = \frac{1}{Z} \displaystyle \sum_{i \in \cB} N_i\muhat_i$ where $Z = \displaystyle \sum_{i \in \cB} N_i$ \\
}
    \STATE \textbf{Output:} $\muhatBcal$
  \end{algorithmic}
\end{algorithm}


In this section, we propose a novel estimator called HAVER (Head AVERaging). 
The key idea is the following observation: for many problem instances of interest, there exists a set of good arms whose means are \textit{very close} to the maximum mean, so it might be a good idea to average out a few empirical top arms, which may reduce the variance at the price of introducing some bias.

At the same time, we still want to be as good as the oracle, so we should be leveraging some form of pessimism that MLCB employ.

With this intuition in mind, we now describe HAVER in detail whose full pseudocode can be found in Algorithm~\ref{algo:haver}. 


For each arm ${i \in [K]}$, we define $\gam_{i} = \sqrt{\fr{18}{N_{i}}\log\del[2]{\del[1]{\fr{KS}{N_i}}^{4}}}$ as its confidence width where $S = N_{\max}\sum_{j \in [K]}N_j$ and ${N_{\max} = \max_{j \in [K]}N_j}$. HAVER consists of three key steps. First, we select a pivot arm with maximum lower confidence bound $\hr := \argmax_{i \in [K]} \hmu_i - \gam_i$. Next, we form a candidate set of arms, $\cB := \cbr{i \in[K]: \hmu_i \ge \hmu_{\hr} - \gam_{\hr},\, \gam_i \le \fgam \gam_{\hr}}$, whose empirical means exceed the maximum lower confidence bound threshold and whose sample sizes are not significantly smaller than that of the pivot arm. 
Note that the intuition of $\cB$ is to include arms that are statistically not distinguishable from the pivot arm.
However, it is essential to include the constraint $\gam_i \le \fr32 \gam_{\hr}$, which forces us to exclude arms with much lower sample size than the pivot arm to make sure we do not inadvertently include bad arms that are being overrepresented due to low sample sizes (i.e., large variances).
Finally, HAVER computes the average of the empirical means of the arms in the candidate set, weighted by the normalized sample size for each arm. 
\looseness=-1



The following definitions are required for our main theorem presented next. We define $s := \argmax_{i \in [K]}\mu_i - \gam_i$ as the ground truth pivot arm. 
We define
  $$\cR \!:=\!\cbr[2]{r\!\in\![K]\!\!: \mu_s\!-\!\fr{4}{3}\gam_s\!+\!\fr{2}{3}\gam_r \le \mu_r  \le  \mu_s\!-\!\cs\gam_s\!+\!\cs\gam_r}$$
as the statistically-plausible candidate set for the pivot arm $\hr$. For any arm ${r \in \cR}$, we define 
$$\cBs(r) := \cbr[2]{i \in [K]: \mu_{i} \ge \mu_{s} - \cstar\gam_{s},\, \gam_i \le \fgam \gam_{r}}$$
as the set of \textit{good} arms whose means are \textit{very close} to the maximum mean and whose sample sizes are \textit{within a constant factor} from the ground truth pivot arm's sample size. Also, for any arm $r \in \cR$, we define ${\cBp(r) := \cbr{i: \mu_{i} \ge \mu_{s} - \cplus\gam_{s} - \cplus\gam_{i},\, \gam_i \le \fgam \gam_{r}}}$ as the set of \textit{nearly good} arms whose the condition is relaxed from $\cBs(r)$.
On the event of $r$ being the pivot arm, $\cBs(r)$ represents the set of arms that should belong to the set $\cB$ with overwhelming probability, and $[K] \sm \cBp(r)$ represents the set of arms that should not belong to the set $\cB$ with overwhelming probability.
\begin{theorem}\label{thm:haver21count2}
  HAVER achieves
  \begin{align*}
    &\MSE(\hmuhaver) \\
    \stackrel{}{=}\,& \tilcO\!\del{\del{ \max_{r \in \cR} \fr{1}{\sum_{j \in \cBs(r)}N_j}\sum_{i \in \cBp(r)}N_i\Delta_i}^2 \wedge \fr{1}{N_1}} \\
    +& \tilcO\!\del{\max_{r \in \cR} \max_{k=0}^{d(r)}\!\max_{\substack{\cS: \cBs(r) \subseteq \Scal \subseteq \cBp(r) \\ \abs{\Scal}=n_*(r)+k }}\fr{k\sum_{j \in \Scal \sm \cBs(r)}N_j}{\del{\sum_{j \in \Scal}N_j}^2}\!\wedge\!\fr{1}{N_1}}\!\\
    +& \tilcO \del{\fr{1}{\min_{r \in \cR}\sum_{j \in \cBs(r)}N_j} \wedge \fr{1}{N_1}}
    + \tilcO \del{\fr{1}{KN_1}}.
  \end{align*}
  where $d(r) = \abs{\cBp(r)} - \abs{\cBs(r)}$ and $n^*(r) = \abs{\cBs(r)}$.
\end{theorem}
%

  The theorem above provides a fine-grained, instance-dependent upper bound on the MSE of HAVER, though with a rather complicated form.
  While we leave detailed explanation on the bound above to the end of this section, one can still observe that the bound above is of order between $\fr{1}{N_1}$ and $\fr{1}{KN_1}$.
  This demonstrates that HAVER meets the first criterion D1: it performs at least as well as the oracle rate. 
  The primary reason HAVER obtains the oracle rate, $\fr{1}{N_1}$, is due to its use of the pessimism principle when choosing the pivot arm based on the maximum lower confidence bound in a similar way to   MLCB estimator.
  However, due to a careful averaging over the set $\cB$, HAVER is able to achieve instance-dependent accelerated rates in various problems as we show below.

\paragraph{The case of equal number of samples.}
In the next three corollaries, we focus on instances where the sample sizes are equal. This setup allows us to better understand the influence of the arm means to accelerated rates.
\begin{assumption}[Equal number of samples]\label{ass:equal_arm_pulls} 
  The number of samples across all arms is the same, ${\forall i \in [K],\, N_i = N}$ for some $N > 0$.
\end{assumption}

\begin{corollary}\label{cor:haver_equal_arm_pulls_generic}
  Under Assumption~\ref{ass:equal_arm_pulls} (equal sample sizes), let ${\gam := \sqrt{\frac{18}{N}\log\del{\del{K^2N}^{4}}}}$, ${\cBs :=\cbr{i \in [K]: \Delta_{i} \le \cstar\gam}}$, and ${\cBp := \cbr{i \in [K]: \Delta_{i} \le \cplusq\gam}}$. 
  Then,
  \begin{align*}
    \MSE(\hmuhaver) 
    \stackrel{}{=}\,& \tilcO \del[4]{\del[3]{\fr{1}{\abs{\cBs}}\textstyle\sum_{i = 1}^{\abs{\cBp}}\Delta_i}^2 \wedge \fr{1}{N}} \\
    &+ \tilcO \del[4]{\fr{1}{N}\del[3]{\log\del[2]{\fr{\abs{\cBp}}{\abs{\cBs}}}}^2 \wedge \fr{1}{N}} \\
    &+ \tilcO \del{\fr{1}{\abs{\cBs}N}} + \tilcO \del{\fr{1}{KN}} ~.
  \end{align*}
\end{corollary}

The bound above is further simplified from Theorem~\ref{thm:haver21count2}, and exhibits an interesting behavior as a function of $|\cBs|$ and $|\cBp|$.
  Assuming that cardinality of $\cBs$ fixed, HAVER is particularly beneficial for problem instances where the gap between $\cBs$ and $\cBp$ is small, as it leads to a simultaneous decrease in the first two terms. 
  For many cases, the log ratio in the second term is less than 1 and contribute to the acceleration. 
  Indeed, we next present a few such instances where HAVER achieves strictly accelerated rates compared to the oracle.

\begin{corollary}\label{cor:haver_equal_arm_pulls_kstar_best}
  Under Assumption~\ref{ass:equal_arm_pulls} (equal number of samples), consider the $\Kstar$-best instance where $\forall i \in [\Kstar],\, \mu_i = \mu_1$. If $N > \fr{256}{\Delta_{\Kstar+1}^2}\log\del{\fr{256K^2}{\Delta_{\Kstar+1}^2e}}$, HAVER achieves
  \begin{align*}
    \MSE(\hmuhaver) = \tilcO \del{\fr{1}{\Kstar N}}.
  \end{align*}
\end{corollary}
The proof is in Appendix~\ref{cor_proof:haver_equal_arm_pulls_kstar_best}.
When the number of samples grows and exceeds the inverse gap of the $\Kstar+1$ arm, HAVER becomes more effective at rejecting the bad arms. This enables HAVER to achieve an acceleration by a factor of $\Kstar$.

\begin{corollary}\label{cor:haver_equal_arm_pulls_alpha_poly}
  Under Assumption~\ref{ass:equal_arm_pulls} (equal number of samples), consider the Poly($\alpha$) instance where $\forall i \ge 2,\, \Delta_i = \del[1]{\fr{i}{K}}^{\alpha}$ where $\alpha \ge 1$. If $N \le \fr{1}{2}\log(K^2N)\del[1]{\fr{K}{\alpha}\log(2)}^{2\alpha}$, HAVER achieves
  \begin{align*}
    \MSE(\hmuhaver) = \tilcO \del{\fr{1}{(\alpha \wedge K) N}}.
  \end{align*}
\end{corollary}
The proof is in Appendix~\ref{cor_proof:haver_equal_arm_pulls_alpha_poly}.
  In this problem instance, HAVER achieves a factor $\alpha$ acceleration. This result is intuitive: as $\alpha$ increases, there are more good arms, allowing HAVER to capture many more good arms and thus achieve greater acceleration. However, when $N$ becomes too large, acceleration diminishes. This is because the empirical means of all arms concentrate more around their means, causing HAVER to only include the best arm in $\cB$, thereby loosing the potential for acceleration.
  In the next corollary, we assume all arm means are identical and shift our focus to the impact of varying sample sizes.

\paragraph{The case of all arms having equal means.}
We now turn to a case where all arms have equal means to focus on the effect of the sample counts on the accelerated rate.
\begin{corollary}\label{cor:haver_non_equal_arm_pulls_allbest}
  Consider the all-best instance where $\forall i \in [K],\, \mu_i = \mu_{1}$ and the number of samples are characterized by $N_i = \del{K - i + 1}^{\beta}$. In the regime  of $\beta \in (0,1)$, HAVER achieves
  \begin{align*}
    \MSE(\hmuhaver) = \tilcO\del{\fr{1}{K N_1}}.
  \end{align*}
\end{corollary}
The proof is in Appendix~\ref{cor_proof:haver_non_equal_arm_pulls_allbest}.
  In this instance, the number of samples are characterized by $N_i = \del{K - i + 1}^{\beta}$. For $\beta \in (0, 1)$, the number of samples forms a smoothly decreasing polynomial curve. 
  In this regime, HAVER achieves a factor of $K$-fold acceleration.
  This corollary provides a different perspective on the problem, showing that HAVER can achieve acceleration in both changing arm means and their sample sizes. 

\paragraph{Detailed interpretation of the bound presented in Theorem~\ref{thm:haver21count2}.}

  Through these corollaries, we have demonstrated that HAVER can indeed achieve acceleration over the oracle in certain instances. Therefore, we can conclude that HAVER satisfies both of the desired criteria. 
  
  Next, we provide a more detailed into HAVER's theorem and offer some intuition on where this acceleration comes from. HAVER can achieve acceleration by forming a candidate set of good arms, $\cB$, and taking weighted average of their empirical means. To delve deeper, we need to understand the intuition of $\cR$, $\cBs(r)$, and $\cBp(r)$. Since the algorithm revolves around the pivot arm, it is natural to define a set $\cR$ that includes all the arms that can be pivot with nontrivial probability. For any pivot arm $r$, we define the set $\cBs(r)$, referred to as the set of \textit{good} arms, based on conditions that ensure it closely aligns with the candidate $\cB$, by matching both the mean deviation and sample size constraints. However, noticing that $\cB$ is highly likely to include nearly-good arms that are marginally close to the good arms, we introduce the set of \textit{nearly-good} arms $\cBp(\hr)$, to further refine our analysis.

  Since the set $\cB$ is a random set, its composition varies depending on the observed samples. 
  To analyze, we first identify a set of statistically-plausible set of pivot arms, which is $\cR$ we define above.
  Assuming $\hr \in \cR$, which is a likely event, we identify four events based on 
  the set $\cB$. The first event, $G_0 = \cbr{\cB = \cBs(\hr)}$, represents the ideal 
  scenario where the candidate set $\cB$ exactly aligns with $\cBs(\hr)$. 
  This event contributes to the third term in the theorem. 
  In this ideal event, HAVER's MSE is in order of $\fr{1}{\sum_{j \in \cBs(r)}N_j}$. Depending on the instances, HAVER will achieve acceleration as shown in Corollaries \ref{cor:haver_equal_arm_pulls_kstar_best}, \ref{cor:haver_equal_arm_pulls_alpha_poly}, and \ref{cor:haver_non_equal_arm_pulls_allbest}. The second event, $G_1 = \cbr{\cBs(\hr) \subset \cB \subseteq \cBp(\hr)}$, is also highly likely and captures scenarios where $\cB$ contains the good arms but may include additional nearly-good arms. This event contributes to the first and second terms in the theorem.
  The next two events, ${G_2 = \cbr{\exists i \in \cBs(\hr),\, i \not\in \cB}}$ and ${G_3 = \cbr{\cBs(\hr) \subseteq \cB,\, \exists i \not\in \cBp(\hr),\, i \in \cB}}$, are highly unlikely to occur. The event $G_2$ captures the scenario where $\cB$ misses an arm from $\cBs(\hr)$, while the event $G_3$ corresponds to $\cB$ including supoptimal arm that is not part of $\cBp(\hr)$. 
  These unlikely events contribute to the potential $K$-fold acceleration.


\section{Experiment}


In this section, we present a comprehensive empirical analysis to evaluate and compare the performance of HAVER, LEM, DE, and WE across three distinct scenarios: (1) the multi-armed bandit setting, (2) Q-learning applied to the Grid World problem, and (3) Monte Carlo Tree Search (MCTS) applied to the FrozenLake environment in OpenAI Gym. Our first experiment setting, in the multi-armed bandits (MAB) domain, is a controlled environment designed to examine the scaling effect as we increase the number of samples and number of distributions. Within this setting, we consider three problem instances $K^*$-best and Poly($\alpha$) and uniform distributed means. The first two instances align with our Corollary 8 and 9, and the last instance can be considered an almost general case where we do not assume any structure based on the distributions' means. Our second and third experimental settings, focused on Q-learning and Monte Carlo Tree Search (MCTS), are designed as a fully general environment. In these settings, we do not impose any implicit structural assumptions on the problem. In particular, when the agent is at a particular state, the reward distributions for the available actions are not required to follow any specific structure.

We notice several potential challenges that arise in practice and suggest solutions to address them.
First, the variance of an arm's distribution is often unknown to the user. To mitigate this issue, we use an unbiased estimate of the variance, i.e., $\hat{\sigma}_i^2 = \fr{\sum_{j=1}^{N_i}\del{X_{i,j} - \hmu_i}^2}{N_i-1}$. Second, since we are working with estimates of variances, we should incorporate these estimates into the average weights in the averaging step. For each arm $i \in [K]$, its average weight becomes $\fr{N_i}{\hat{\sigma}_i^2}$. However, $\hat{\sigma}_i^2$ can be arbitrarily small, particularly in early episodes in Q-learning, which causes the denominator to explode. To prevent this, we add a small tunable hyperparameter $\eps$ to $\hat{\sigma}_i^2$. This parameter is set to $0.01$ in our experiments. 
The detailed pseudocode is available in Appendix~\ref{sec:haver_in_practice}.

\subsection{Multi-Armed Bandits}

The multi-armed bandit environment can serve as a controlled environment for our experiments. It is important to note that the standard multi-armed bandit problem, particularly best-arm identification with a fixed budget, differs from the maximum mean estimation problem.
In the former, a learner adaptively selects arms to sample and, after exhausting the budget, identifies the optimal arm. In contrast, in the maximum mean estimation problem, the samples are fixed and provided by the environment. Our focus is solely on estimating the value of the largest mean among $K$ distributions. 

We follow a similar experiment setup as formulated in \citet{vanhasselt13estimating}. Consider $K$ ads, each with a click rate $\mu_i$. For simplicity, we assume the return for each ad follows the Bernoulli distribution (i.e., each click has a reward of one), so the ad's mean return is equal to its click rate $\mu_i$. Each ad is equally shown to customers $N$ times, and the return is recorded each time.
We compare the results of HAVER, LEM, DE, and WE in two different settings for three problem instances. In the default configuration, we set the number of ads $K = 50$ and the number of samples per ad is $N=500$, and mean click rates are from the interval $[0.002, 0.005]$. In the first setting, we vary the number of samples per ad $N = \cbr{100, 200, \ldots, 1000}$, which is equivalent to the total number of samples ranging from $5000$ to $50000$. In the second setting, we vary the number of ads $K = \cbr{30,40,\ldots,100}$.

We consider two problem instances ($\Kstar$-best and Poly($\alpha$)) mentioned in our previous corollaries, as well as a generic uniformly-sampled instance. For the $\Kstar$-best instance, the arms are divided into two halves: the arms in the first half have mean click rates of $0.005$, while the remaining arms have mean click rates of $0.002$. For the Poly$(\alpha)$, we set $\alpha = 2$ and normalize the mean click rates to the interval $[0.002, 0.005]$. Lastly, for the generic instance, the mean click rates are uniformly sampled from the interval $[0.002, 0.005]$. For each problem instance, we repeat the two experiment settings accordingly. We present the results in terms of $\MSE = \Bias^2 + \Var$ for all these experiments, averaging the results over 1000 trials.

We observe similar patterns over the three problem instances. We present the results for the generic instance in Figure~\ref{fig:uniform_mse_vs_num_actions} while the results for the other two instances are included in the appendix due to space constraints.
The results of the first and second experimental settings are shown at the top and bottom, respectively. Overall, HAVER's MSE effectively decays with both the number of samples and the number of ads, which aligns with our theorem and corollaries. In the first setting, unsurprisingly, the MSE decreases for all estimators as the number of samples per ad increases, with HAVER consistently achieving the lowest MSE in all cases. Additionally, HAVER has the lowest squared bias among the estimators, suggesting its effectiveness in capturing the good arms. In the second setting, the MSE of both HAVER and DE decreases as the number of ads increases, whereas LEM's performance deteriorates with a larger number of ads. Interestingly, the MSE of WE does not decay with an increasing number of ads.

\begin{figure}[ht!]
  \begin{subfigure}{0.9\linewidth}
    \begin{center}
      \includegraphics[width=1.0\linewidth]{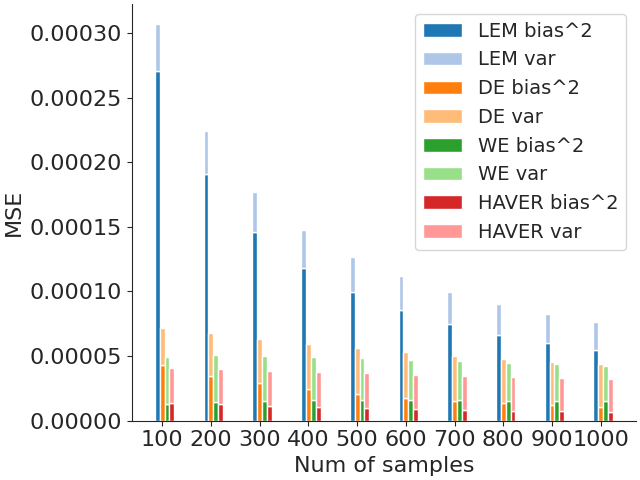}
    \end{center}
  \end{subfigure}
  \begin{subfigure}{0.9\linewidth}
    \begin{center}
      \includegraphics[width=1.0\linewidth]{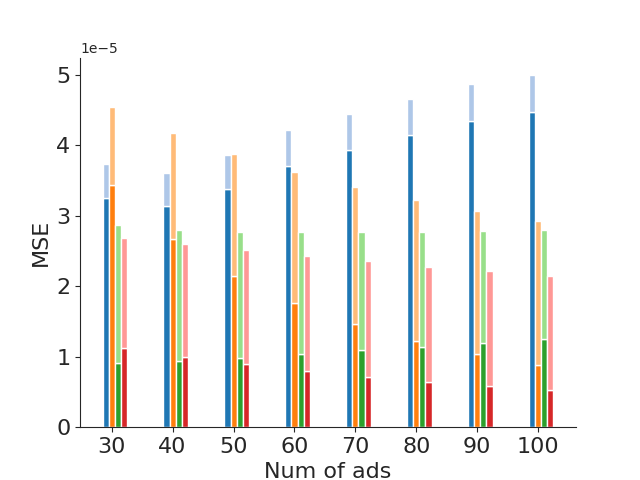}
    \end{center}
  \end{subfigure}
  \caption{Uniform sampling instance. The results are averaged over 1000 trials.}
  \label{fig:uniform_mse_vs_num_actions}    
\end{figure}

\subsection{Q-Learning Applied to the Simple Grid World}

The maximum mean estimation problem is often discussed in the context of Q-learning \citep{watkins89learning}. The update of Q-learning is
\begin{align*}
  Q\del{s_t,a_t} \leftarrow &~Q\del{s_t,a_t} \\
  &+ \alpha_t\del[1]{r_t + \eta \max_{a}Q\del{s_{t+1},a} - Q\del{s_t,a_t} }.
\end{align*}
where $Q\del{s,a}$ represents the value of action $a$ in state $s$ at time $t$, the reward $r_t$ is drawn from a fixed distribution as a function of $s_t$ and $a_t$, and $\eta$ is a discount factor.
Here, given the history of visits and action selections in the state $s_{t+1}$, we want to estimate the maximum value of $Q(s_{t+1},a)$ among the actions. Errors propagation problems resulting from inaccurate estimation can have a negative effect on the learning process.
\looseness=-1

In this experiment, we use a 3x3 grid world environment setup similar to that in \citet{vanhasselt10double}. The agent starts in the lower-left cell, and the terminal cell is located in upper-right corner. The agent's goal to maximize its cumulative reward before reaching the terminal cell. There are two kinds of rewards. At each step, the agent performs an action and gains a random reward drawn from the standard Gaussian distribution $\cN(\mu = 0, \sigma^2 = 1)$. 
Upon reaching the terminal cell, the agent gains a fixed reward of 5. The discount factor is set to $\eta = 0.95$. We run Q-learning for 10000 steps, and each time the agent reaches the terminal state, it is reset to the initial position.  
\looseness=-1

We focus on the estimation at the initial state, as it serves as a natural indicator of the agent's performance. If the agent performs well at the initial state, it is likely to perform well in subsequent states. For the initial state, the maximum action value is $5\eta^{4} \approx 4.073$. In addition, we consider the optimal average reward per step, which is 1, as a direct measure of the agent's performance. 
To investigate the effect of increasing the number of actions, we design a unique grid world setting where each action-up, down, left, right- is duplicated $M$ times, resulting in a total of $4\cdot M$ actions. We refer to this setting as the inflated grid world.
\looseness=-1

\begin{figure}[t]
  \begin{subfigure}{0.9\linewidth}
    \begin{center}
      \includegraphics[width=0.8\linewidth]{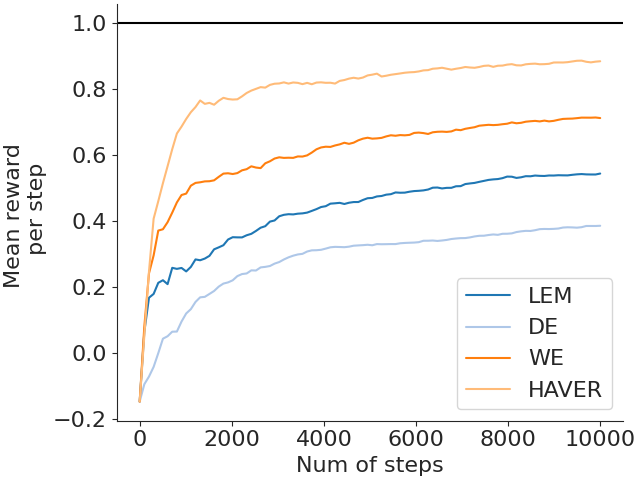}
    \end{center}
  \end{subfigure}
  \begin{subfigure}{0.9\linewidth}
    \begin{center}
      \includegraphics[width=0.8\linewidth]{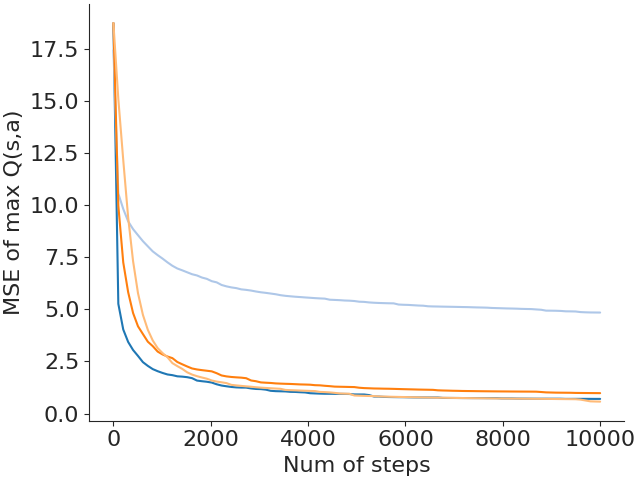}
    \end{center}
  \end{subfigure}
  \caption{Q-learning in the regular grid world environment. The results are averaged over 1000 trials. The optimal mean reward per step is the black line.}
  \label{fig:qlearning_k4}   
\end{figure}

\begin{figure}[t]
  \begin{subfigure}{0.9\linewidth}
    \begin{center}
      \includegraphics[width=0.8\linewidth]{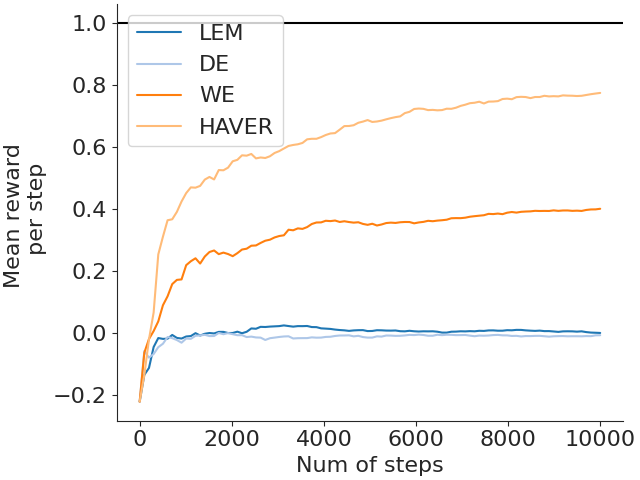}
    \end{center}
  \end{subfigure}
  \begin{subfigure}{0.9\linewidth}
    \begin{center}
      \includegraphics[width=0.8\linewidth]{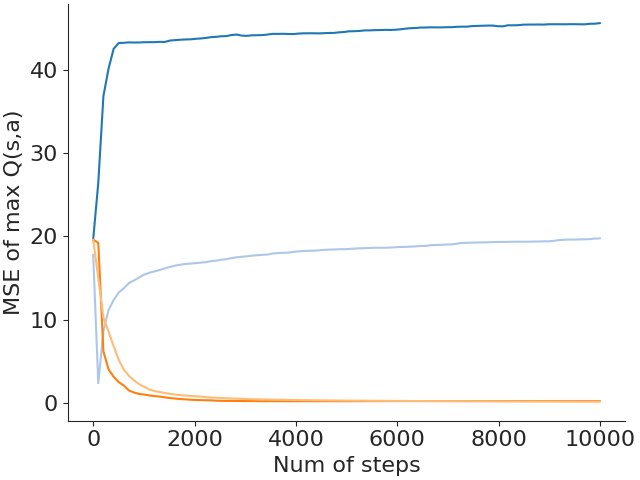}
    \end{center}
  \end{subfigure}
  \caption{Q-learning in the inflated grid world with the number of actions at each state is duplicated to 4. The results are averaged over 1000 trials. The optimal mean reward per step is the black line.}
  \label{fig:qlearning_k16}   
\end{figure}

We compare the performance of HAVER, LEM, DE, and WE estimators within Q-learning framework in two settings: the regular grid world and the inflated grid world where $M = 4$. The results for both settings are shown in Figure~\ref{fig:qlearning_k4} and ~\ref{fig:qlearning_k16} respectively. In both settings, HAVER consistently outperforms the other estimators in achieving higher mean rewards and in lower MSE in $Q(s,a)$ estimation. HAVER's performance is more pronounced in the inflated grid world setting, where it converges to the mean reward much faster than the other estimators. We observe that the HAVER's MSE in $Q(s,a)$ estimation tends to lag behind in earlier time steps, likely due to the high bias introduced during the averaging computation. 
\looseness=-1

\subsection{Monte Carlo Tree Search (MCTS) Applied to the FrozenLake Environment}

Maximum mean estimation is an integral part of the Monte Carlo Tree Search (MCTS) algorithm. During the backup step in MCTS, the agent aims to estimate the maximum value of a node based on the highest values of its child nodes. In this study, we investigate the effectiveness of maximum mean estimators in MCTS within the FrozenLake environment as implemented in OpenAI Gym \citep{brockman2016openai}. In this environment, the agent's objective is to navigate an icy grid starting from the top-left cell to the bottom-right goal position. A key challenge is the presence of holes scattered across the grid; if the agent falls into a hole, the episode ends immediately with a reward of 0. Successfully reaching the goal position, however, grants a reward of 10, the maximum possible reward. If the agent fails to reach the goal position within a time limit of 40 steps, it receives a reward of 0.

\begin{figure}[ht!]
  \begin{subfigure}{0.9\linewidth}
    \begin{center}
      \includegraphics[width=0.7\linewidth]{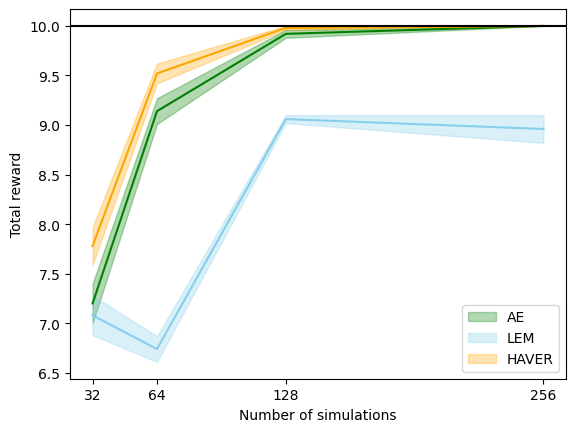}
    \end{center}
  \end{subfigure}
  \begin{subfigure}{0.9\linewidth}
    \begin{center}
      \includegraphics[width=0.7\linewidth]{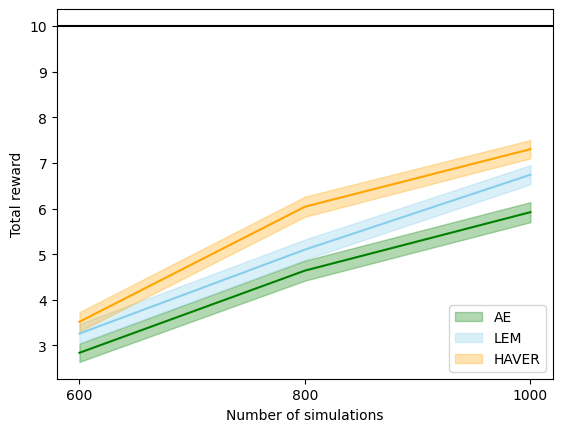}
    \end{center}
  \end{subfigure}
  \caption{MCTS applied to the FrozenLake environments (top: 4x4 environment, bottom: 8x8 environment). The results are averaged over 500 trials. The optimal total reward is the black line.}
  \label{fig:mcts_frozenlake_deterministic}    
\end{figure}

We evaluate the performance of HAVER, LEM, and AE (the average estimator utilized in the original MCTS implementation by \citet{kocsis06bandit}) in grid environments: a smaller 4x4 grid and a larger 8x8 grid. Both environments are deterministic, meaning that when the agent selects a direction, it moves exactly in that direction without any stochasticity. Our experiments vary the number of MCTS simulations to assess the impact of simulation count on the agent's performance. For the smaller 4x4 grid, we test a range of 32, 64, 128, and 256 MCTS simulations. For the larger 8x8 grid, which presents a larger state space and higher complexity, we extend the range of MCTS simulations to 600, 800, and 1000 to provide the agent with sufficient exploration capabilities. Figure \ref{fig:mcts_frozenlake_deterministic} shows that HAVER outperforms the other estimators, achieving higher rewards across all tested simulation counts.

\section{Related Work}
\label{sec:related}
\vspace{-.5em}

\textbf{Maximum mean estimation (MME).}
MME can be taken as a rather naive task since the naive estimator of LEM seems to be very reasonable.
However, as reported by~\citet{smith06optimizer}, LEM suffers from an overestimation bias, which can be large when the sample size is small or the number of distributions $K$ is large.
In the machine learning community, \citet{vanhasselt10double} is the first one to formally study the MME problem, to our knowledge.
In particular, the author reports that the overestimation problem of LEM is harmful to the performance of Q-learning~\cite{watkins89learning}, which is a popular algorithm for reinforcement learning, and instead proposes a negatively biased estimator called double estimator (DE) that works similarly to the 2-fold cross-validation.
There are numerous follow-up studies that propose various MME algorithms including weighted estimator (WE)~\citep{deramo16estimating,deramo21gaussian} and maxmin estimator~\citep{lan20maxmin}.
The maximum mean estimation problem also arises from Monte Carlo tree search (MCTS)~\citep{coulom06efficient,kocsis06bandit}, which is an online planning methodology that forms a search tree from the current state $s$ and aims to identify the best action by repeatedly running randomized rollouts in a simulator.
Interestingly, the de facto standard algorithm called UCT~\citep{kocsis06bandit} uses the average estimator (AE) $\fr{1}{\sum_{i'=1}^K N_{i'}} \sum_{i=1}^K \sum_{j=1}^{N_i} X_{i,j}$ as an estimator for the value of each state.
While this is not a consistent estimator in general, it is consistent in the MCTS setting since UCT is designed to take the best action most of the time asymptotically, i.e., $\fr{N_1}{\sum_i N_i} \rarrow 1$.
To address the issue of the slow convergence of AE in the nonasymptotic regime, \citet{tesauro10bayesian} proposed a Bayesian estimator that is essentially WE and proved its effectiveness in MCTS.
In \citet{dam2024unified}, The estimator WE was further generalized to the weighted power mean estimator and applied to MCTS.
However, to our knowledge, all of these studies lack precise error analysis and instead provide bounds on bias and variance separately. 
\textbf{Best-arm identification.}
A closely-related problem to MME is the best-arm identification problem~\citep{chernoff59sequential,even-dar06action,bubeck09pure}, especially the fixed budget setting.
In this setting, at time step $t\in \{1,\ldots,T\}$, the learner chooses an arm $I_t$ from $K$ arms and observes a reward $r_t$ from the chosen arm.
After $T$ time steps, the learner must output the estimated best arm, i.e., the arm with the largest mean reward, and the learner's performance is measured by the accuracy of the estimation.
BAI has been studied extensively~\cite{kaly12pac,karnin13almost,jamieson14lil,zhao23revisiting}, resulting in applications in adaptive crowdsourcing~\cite{tanczos17akllucb}, hyperparameter optimization~\cite{li18hyperband}, and accelerating the $k$-medoids algorithm~\cite{bagaria21bandit}.
However, BAI is different from MME in two respects: MME (i) requires estimation of the largest mean \textit{value} rather than the \textit{identity} of the arm with the largest mean and (ii) does not perform active sampling; i.e., the samples are chosen from an external entity rather than being chosen by the learner.

\section{Conclusion and Future Work}

We have identified two key desiderata of an ideal estimator: (i) it should achieve an MSE rate as good as the oracle rate and (ii) it can strictly outperform the oracle rate in an instance-dependent manner. Then, we have proposed a novel estimator HAVER and proved that it satisfies the two desiderata.
Our result is first of its kind and opens up numerous interesting future research directions. 
Finally, it would be interesting and relevant to practice to remove the i.i.d. assumption and instead consider nonstationary distributions, which we believe are better models of sample generation in Q-learning.

\section*{Acknowledgements}
Tuan Ngo Nguyen and Kwang-Sung Jun were supported in part by the National Science Foundation under grant CCF-2327013.






\bibliographystyle{abbrvnat_lastname_first_overleaf}
\bibliography{library-overleaf}

\section*{Checklist}



 \begin{enumerate}

 \item For all models and algorithms presented, check if you include:
 \begin{enumerate}
   \item A clear description of the mathematical setting, assumptions, algorithm, and/or model. [Yes]
   \item An analysis of the properties and complexity (time, space, sample size) of any algorithm. [Yes]
   \item (Optional) Anonymized source code, with specification of all dependencies, including external libraries. [No]
 \end{enumerate}

 \item For any theoretical claim, check if you include:
 \begin{enumerate}
   \item Statements of the full set of assumptions of all theoretical results. [Yes]
   \item Complete proofs of all theoretical results. [Yes] (most proofs are in the appendix)
   \item Clear explanations of any assumptions. [Yes]  
 \end{enumerate}

 \item For all figures and tables that present empirical results, check if you include:
 \begin{enumerate}
   \item The code, data, and instructions needed to reproduce the main experimental results (either in the supplemental material or as a URL). [No]
   \item All the training details (e.g., data splits, hyperparameters, how they were chosen). [Not Applicable]
         \item A clear definition of the specific measure or statistics and error bars (e.g., with respect to the random seed after running experiments multiple times). [No Applicable]
         \item A description of the computing infrastructure used. (e.g., type of GPUs, internal cluster, or cloud provider). [No Applicable]
 \end{enumerate}

 \item If you are using existing assets (e.g., code, data, models) or curating/releasing new assets, check if you include:
 \begin{enumerate}
   \item Citations of the creator If your work uses existing assets. [Not Applicable]
   \item The license information of the assets, if applicable. [Not Applicable]
   \item New assets either in the supplemental material or as a URL, if applicable. [Not Applicable]
   \item Information about consent from data providers/curators. [Not Applicable]
   \item Discussion of sensible content if applicable, e.g., personally identifiable information or offensive content. [Not Applicable]
 \end{enumerate}

 \item If you used crowdsourcing or conducted research with human subjects, check if you include:
 \begin{enumerate}
   \item The full text of instructions given to participants and screenshots. [Not Applicable]
   \item Descriptions of potential participant risks, with links to Institutional Review Board (IRB) approvals if applicable. [Not Applicable]
   \item The estimated hourly wage paid to participants and the total amount spent on participant compensation. [Not Applicable]
 \end{enumerate}

 \end{enumerate}

\onecolumn
\appendix
\aistatstitle{Supplementary Materials}

\fancypagestyle{plain}{
\fancyfoot[C]{\thepage}}
\pagestyle{plain}
\setlength{\footskip}{20pt}	

\part{Appendix} 

\parttoc 



\clearpage
\section{More on  Experiments}

\subsection{HAVER Estimator in Practice}
\label{sec:haver_in_practice}

The following Algorithm~\ref{algo:haver_in_practice} is a practical version of HAVER. There are two key parts. First, we compute the unbiased estimate of the variance, i.e., $\hsigma_i^2 = \fr{\sum_{j=1}^{N_i}\del{X_{i,j} - \hmu_i}^2}{N_i-1}$. Second, we incorporate these variance estimates into the the average weights in the averaging step. For each arm $i \in [K]$, its average weight becomes $\fr{N_i}{\hat{\sigma}_i^2}$. However, $\hat{\sigma}_i^2$ can be arbitrarily small, so it can causes the denominator to explode. To prevent this, we add a small tunable hyperparameter $\eps$ to $\hat{\sigma}_i^2$. This hyperparameter is set to $0.01$ in our experiments.

\begin{algorithm}[H]
  \caption{HAVER Estimator in Practice}\label{algo:haver_in_practice}
  \begin{algorithmic}
    \STATE{\textbf{Input:} A set of $K$ arms with samples $\cbr{X_{i, 1:N_i}}_{i=1}^{K}$, hyperparameter $\eps$.
}
    \STATE{
 For each arm $i$, compute its empirical mean $\hmu_i$ and variance $\hsigma_i^2$ \\
\qquad $\hmu_i = \fr{1}{N_i}\sum_{j=1}^{N_i}X_{i,j}$, \qquad $\hsigma_i^2 = \fr{\sum_{j=1}^{N_i}\del{X_{i,j} - \hmu_i}^2}{N_i-1}$. \\
 Find pivot arm $\hr$  \\
\qquad $\hr = \displaystyle \argmax_{i \in [K]} \muhat_i - \gam_{i}$ \\
\qquad where $\sqrt{\frac{18}{N_{i}}\log\del{\del{\fr{KS}{N_i}}^{4}}}$, \\
\qquad $S = N_{\max}\sum_{j \in [K]}N_j$, and $N_{\max} = \max_{i \in [K]}N_i$. \\
 Form a set $\cB$ of desirable arms \\
\qquad $\cB = \cbr{i \in [K]: \muhat_{i} \ge \muhat_{\hr} - \gam_{\hr},\, \gam_{i} \le \fgam \gam_{\hr}}$. \\
 Compute the weighted average of empirical means of desirable arms in set $\cB$ \\
\qquad $\hmuB = \fr{1}{Z} \displaystyle \sum_{i \in \cB} \fr{N_i}{\sigmahat_i^2+\eps}\muhat_i$, \\
\qquad where $Z = \sum_{i \in \cB} \fr{N_i}{\sigmahat_i^2+\eps}$. \\
}
    \RETURN $\hmuB$.
  \end{algorithmic}
\end{algorithm}

\clearpage
\subsection{Multi-Armed Bandits for Internet Ads}

\begin{figure}[ht!]
  \begin{subfigure}{0.45\linewidth}
    \begin{center}
      \includegraphics[width=1.0\linewidth]{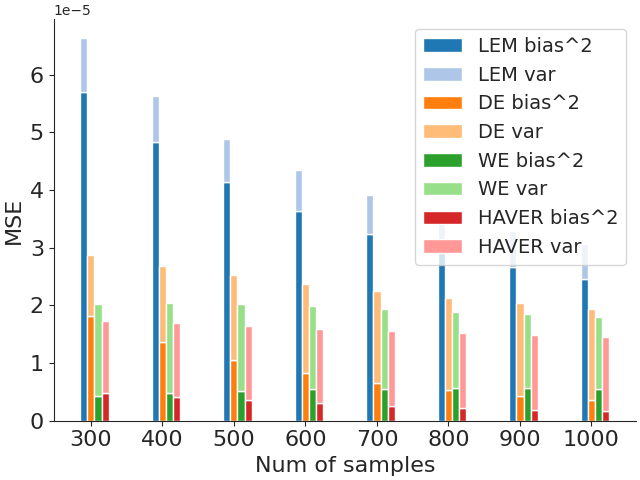}
    \end{center}
  \end{subfigure}
  \begin{subfigure}{0.45\linewidth}
    \begin{center}
      \includegraphics[width=1.0\linewidth]{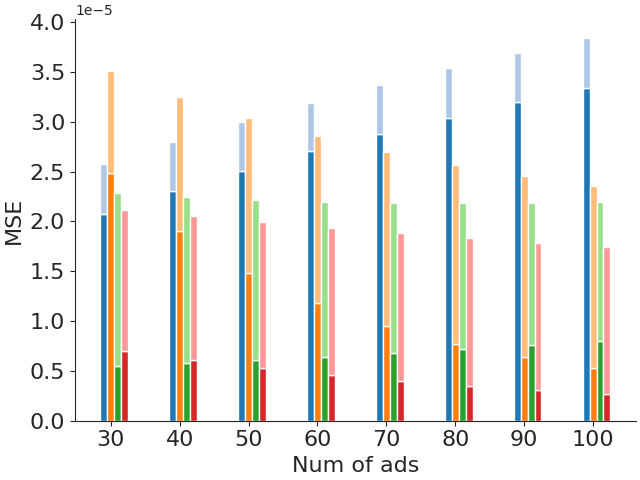}
    \end{center}
  \end{subfigure}
  \caption{Polynomial($\alpha$) instance with $\alpha = 2$. The results are averaged over 1000 trials.}
  \label{fig:polynomial_mse_vs_num_actions}    
\end{figure}

The following Figures~\ref{fig:kstarbest_mse_vs_num_actions} and~\ref{fig:polynomial_mse_vs_num_actions} respectively shows the results for the $K^*$-best and Polynomial($\alpha$) instances mentioned in the main paper.

\begin{figure}[ht!]
  \begin{subfigure}{0.45\linewidth}
    \begin{center}
      \includegraphics[width=1.0\linewidth]{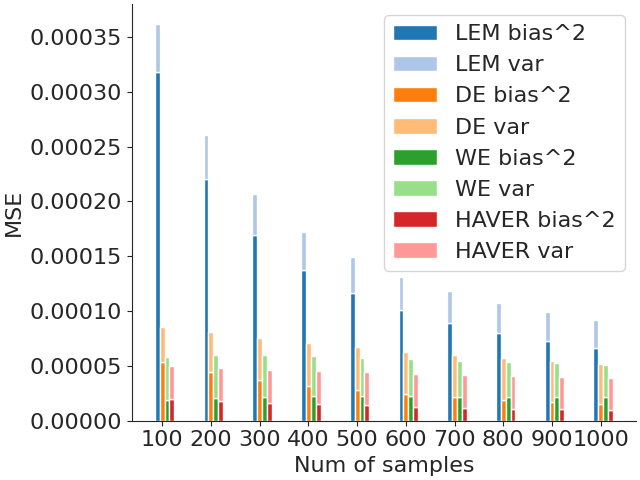}
    \end{center}
  \end{subfigure}
  \begin{subfigure}{0.45\linewidth}
    \begin{center}
      \includegraphics[width=1.0\linewidth]{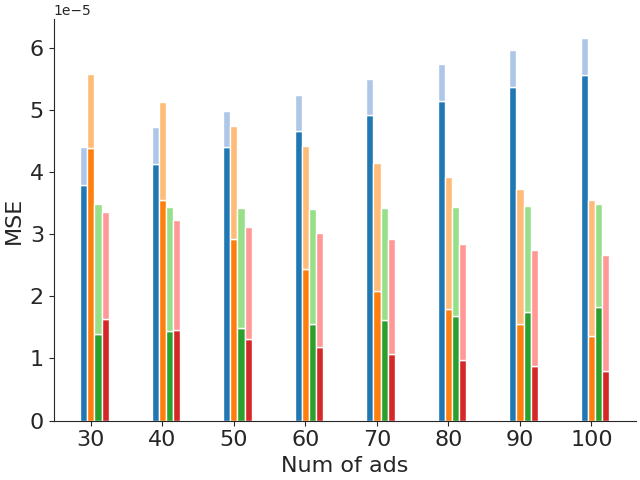}
    \end{center}
  \end{subfigure}
  \caption{$\Kstar$-best instance. The results are averaged over 1000 trials.}
  \label{fig:kstarbest_mse_vs_num_actions}    
\end{figure}

\clearpage
\section{HAVER's Corollaries}

\newtheorem*{cor-0}{Corollary ~\ref{cor:haver_equal_arm_pulls_generic}}
\begin{cor-0}
  Under Assumption~\ref{ass:equal_arm_pulls} (equal number of samples), let $\gam := \sqrt{\fr{18}{N}\log\del{\del{K^2N}^{4}}}$, $\cBs :=\cbr{i \in [K]: \Delta_{i} \le \cstar\gam}$, and $\cBp := \cbr{i \in [K]: \Delta_{i} \le \cplusq\gam}$. HAVER achieves 
  \begin{align*}
    &\MSE(\hmuhaver) \\
    \stackrel{}{=}\,& \tilcO \del{\del{\fr{1}{\abs{\cBs}}\sum_{i = 1}^{\abs{\cBp}}\Delta_i}^2 \wedge {\fr{1}{N}}} \\
    &+ \tilcO \del{\del{\fr{1}{N}\del{\log\del{\fr{\abs{\cBp}}{\abs{\cBs}}}}^2} \wedge {\fr{1}{N}}} \\
    &+ \tilcO \del{\fr{1}{\abs{\cBs}N}} + \tilcO \del{\fr{1}{KN}}. 
  \end{align*}
\end{cor-0}
\begin{proof}
  Under Assumption~\ref{ass:equal_arm_pulls} (equal number of samples), the definitions used in Theorem~\ref{thm:haver21count2} are reduced to the following definitions. 
  
  First, $\forall i \in [K]$, the confidence width $\gam_i$ becomes $\gam$
  \begin{align*}
    \gam_i = \sqrt{\fr{18}{N_i}\log\del{\del{\fr{KN_{\max}\sum_{j \in [K]}N_j}{N_i}}^{4}}} = \sqrt{\fr{18}{N}\log\del{\del{K^2N}^{4}}} =: \gam. 
  \end{align*}
  Second, the ground truth of the pivot arm $s$ becomes 1: $s = \argmax_{i} \hmu_{i} - \cs \gam_{i} = \argmax_{i} \hmu_{i} - \cs \gam = 1$
  Third, $\cBs(r)$ and $\cBp(r)$ become $\cBs$ and $\cBp$ respectively as follow:
  \begin{align*}
    \cBs(r)
    \stackrel{}{=}\,& \cbr{i \in [K]: \mu_{i} \ge \mu_{s} - \cstar\gam_{s},\, \gam_i \le \fgam \gam_{r}} \\
    \stackrel{}{=}\,& \cbr{i \in [K]: \mu_{i} \ge \mu_{1} - \cstar\gam} \\
    \stackrel{}{=}\,& \cbr{i \in [K]: \Delta_{i} \le \cstar\gam} =: \cBs
  \end{align*}
  \begin{align*}
    \cBp(r)
    \stackrel{}{=}\,& \cbr{i \in [K]: \mu_{i} \ge \mu_{s} - \cplus\gam_{s} - \cplus\gam_{i},\, \gam_i \le \fgam \gam_{r}} \\
    \stackrel{}{=}\,& \cbr{i \in [K]: \mu_{i} \ge \mu_{1} - \cplus\gam - \cplus\gam} \\
    \stackrel{}{=}\,& \cbr{i \in [K]: \Delta_{i} \le \cplusq\gam} =: \cBp.
  \end{align*}
  Additionally, $d(r)$ and $n_*(r)$ becomes $d$ and $n_*$: $d(r) = \abs{\cBp(r)} - \abs{\cBs(r)} = \abs{\cBp} - \abs{\cBs} =: d$ and $n_*(r) = \abs{\cBs(r)} = \abs{\cBs} =: n_*$. We adopt a slight abuse of notation for $\cBs$, $\cBp$, $d$ and $n_*$.
  From Theorem~\ref{thm:haver21count2}, we have 
  \begin{align*}
    &\MSE(\hmuhaver) \\
    \stackrel{}{=}\,& \tilcO \del{\del{\displaystyle \max_{r \in \cR} \fr{1}{\sum_{j \in \cBs(r)}N_j}\sum_{i \in \cBp(r)}N_i\Delta_i}^2 \wedge {\fr{1}{N_1}}} \\
    &+ \tilcO \del{\del{\max_{r \in \cR} \max_{k=0}^{d(r)} \max_{\substack{S: \cBs(r) \subseteq \Scal \subseteq \cBp(r) \\ \abs{\Scal}=n_*(r)+k }}\fr{k\sum_{j \in \Scal \sm \cBs(r)}N_j}{\del{\sum_{j \in \Scal}N_j}^2}} \wedge {\fr{1}{N_1}}} \\
    &+ \tilcO\del{\del{\fr{1}{\min_{r \in \cR}\sum_{j \in \cBs(r)}N_j}} \wedge {\fr{1}{N_1}}} \\
    &+ \tilcO \del{\fr{1}{KN_1}}. 
  \end{align*}
  The first component becomes
  \begin{align*}
    \max_{r \in \cR} \fr{1}{\sum_{j \in \cBs(r)}N_j}\sum_{i \in \cBp(r)}N_i\Delta_i^2 
    \stackrel{}{=}\,& \del{\fr{1}{\abs{\cBs}}\sum_{i = 1}^{\abs{\cBp}}\Delta_i}^2.
  \end{align*}
  The second component becomes
  \begin{align*}
    &\max_{r \in \cR} \max_{k=0}^{d(r)} \max_{\substack{S: \cBs(r) \subseteq \Scal \subseteq \cBp(r) \\ \abs{\Scal}=n_*(r)+k }}\fr{k\sum_{j \in \Scal \sm \cBs(r)}N_j}{\del{\sum_{j \in \Scal}N_j}^2} \\
    \stackrel{}{=}\,& \max_{k=0}^{d} \fr{k^2N}{\del{(n_*(r)+k)N}^2} \\
    \stackrel{}{=}\,& \max_{k=0}^{d} \fr{k^2}{\del{n_*(r)+k}^2N} \\
    \stackrel{}{\le}\,& \fr{1}{N}\del{\log\del{\fr{n_*+d}{n_*}}}^2  \tagcmt{use Lemma~\ref{xlemma:max_bound}} \\
    \stackrel{}{=}\,& \fr{1}{N}\del{\log\del{\fr{\abs{\cBp}}{\abs{\cBs}}}}^2. 
  \end{align*}
  The third component becomes
  \begin{align*}
    \fr{1}{\min_{r \in \cR}\sum_{j \in \cBs(r)}N_j} = \fr{1}{\abs{\cBs}N}. 
  \end{align*}
  Thus, we have
  \begin{align*}
    &\MSE(\hmuhaver) \\
    \stackrel{}{=}\,& \tilcO \del{\del{\fr{1}{\abs{\cBs}}\sum_{i = 1}^{\abs{\cBp}}\Delta_i}^2 \wedge {\fr{1}{N}}} \\
    &+ \tilcO \del{\del{\fr{1}{N}\del{\log\del{\fr{\abs{\cBp}}{\abs{\cBs}}}}^2} \wedge {\fr{1}{N}}} \\
    &+ \tilcO\del{\del{\fr{1}{\abs{\cBs}N} } \wedge {\fr{1}{N}}} \\
    &+ \tilcO \del{\fr{1}{KN}} \\
    \stackrel{}{=}\,& \tilcO \del{\del{\fr{1}{\abs{\cBs}}\sum_{i = 1}^{\abs{\cBp}}\Delta_i}^2 \wedge {\fr{1}{N}}} \\
    &+ \tilcO \del{\del{\fr{1}{N}\del{\log\del{\fr{\abs{\cBp}}{\abs{\cBs}}}}^2} \wedge {\fr{1}{N}}} \\
    &+ \tilcO \del{\fr{1}{\abs{\cBs}N} } + \tilcO \del{\fr{1}{KN}}.
  \end{align*}
\end{proof}

\begin{corollary}\label{cor:haver_equal_arm_pulls_all_best}
  Under Assumption~\ref{ass:equal_arm_pulls} (equal number of samples). Consider the all-best instance where $\forall i \in [K], \mu_i = \mu_1$, HAVER achieves
  \begin{align*}
    \MSE(\hmuhaver) = \tilcO \del{\fr{1}{KN}}.
  \end{align*}
\end{corollary}


\begin{proof}
  From Corollary \ref{cor:haver_equal_arm_pulls_generic}, we have $\Bcals = \cbr{i \in [K]: \Delta_i \le \cstar \gam}$ and $\Bcalp = \cbr{i \in [K]: \Delta_i \le \cplusq \gam}$, and
  \begin{align*}
    &\MSE(\hmuhaver) \\
    \stackrel{}{=}\,& \tilcO \del{\del{\fr{1}{\abs{\cBs}}\sum_{i = 1}^{\abs{\cBp}}\Delta_i}^2 \wedge {\fr{1}{N}}} \\
    &+ \tilcO \del{\del{\fr{1}{N}\del{\log\del{\fr{\abs{\cBp}}{\abs{\cBs}}}}^2} \wedge {\fr{1}{N}}} \\
    &+ \tilcO \del{\fr{1}{\abs{\cBs}N}} + \tilcO \del{\fr{1}{KN}}.
  \end{align*}
  For the all-best instance, we can see that $\Bcals = [K]$, $\Bcalp = [K]$. Therefore, from Corollary \ref{cor:haver_equal_arm_pulls_generic}, the first term $\displaystyle \del{\fr{1}{\abs{\Bcals}}\sum_{i = 1}^{\abs{\Bcalp}}\Delta_i}^2$ becomes 0 since $\forall i \in \Bcalp = [K],\, \Delta_i = 0$. Also, the second term $\fr{1}{N}\del{\log\del{\fr{\abs{\Bcalp}}{\abs{\Bcals}}}}^2 = 0$ since $\log\del{\fr{\abs{\Bcalp}}{\abs{\Bcals}}} = 0$. Also, we have $\fr{1}{\abs{\Bcals}N} = \fr{1}{KN}$.

  Therefore, from from Corollary \ref{cor:haver_equal_arm_pulls_generic}, HAVER achieves
  \begin{align*}
    \MSE(\hmuhaver) = \tilcO \del{\fr{1}{KN}}.
  \end{align*}
  
\end{proof}

\subsection{Proof of Corollary ~\ref{cor:haver_equal_arm_pulls_kstar_best}}
\label{cor_proof:haver_equal_arm_pulls_kstar_best}

\newtheorem*{cor-2}{Corollary ~\ref{cor:haver_equal_arm_pulls_kstar_best}}
\begin{cor-2}
  Under Assumption~\ref{ass:equal_arm_pulls} (equal number of samples), consider the $\Kstar$-best instance where $K^* \le K$ $\forall i \in [\Kstar],\, \mu_i = \mu_1$. If $N > \fr{256}{\Delta_{\Kstar+1}^2}\log\del{\fr{256K^2}{\Delta_{\Kstar+1}^2e}}$, HAVER achieves
  \begin{align*}
    \MSE(\hmuhaver) = \tilcO \del{\fr{1}{\Kstar N}}.
  \end{align*}
\end{cor-2}

\begin{proof}
  From Corollary \ref{cor:haver_equal_arm_pulls_generic}, we have $\Bcals = \cbr{i \in [K]: \Delta_i \le \cstar \gam}$, $\Bcalp = \cbr{i \in [K]: \Delta_i \le \cplusq \gam}$, and
  \begin{align*}
    &\MSE(\hmuhaver) \\
    \stackrel{}{=}\,& \tilcO \del{\del{\fr{1}{\abs{\cBs}}\sum_{i = 1}^{\abs{\cBp}}\Delta_i}^2 \wedge {\fr{1}{N}}} \\
    &+ \tilcO \del{\del{\fr{1}{N}\del{\log\del{\fr{\abs{\cBp}}{\abs{\cBs}}}}^2} \wedge {\fr{1}{N}}} \\
    &+ \tilcO \del{\fr{1}{\abs{\cBs}N}} + \tilcO \del{\fr{1}{KN}}. 
  \end{align*}
  Lemma~\ref{xlemma:delta_sufficient} states that $N > \fr{256}{\Delta_{\Kstar+1}^2}\log\del{\fr{256K^2}{\Delta_{\Kstar+1}^2e}}$ implies $\Delta_{\Kstar+1} > \fr{8}{3}\sqrt{\fr{18}{N}\log\del{K^{2}N}}$, Thus, we have $\Delta_{\Kstar+1} > \fr{8}{3}\gam$. Therefore, we have $\Bcals = [\Kstar]$, $\Bcalp = [\Kstar]$.

  Therefore, from Corollary \ref{cor:haver_equal_arm_pulls_generic}, the first term $\displaystyle \fr{1}{\abs{\Bcals}}\sum_{i = 1}^{\abs{\Bcalp}}\Delta_i$ becomes 0 since $\forall i \in \Bcalp = [\Kstar],\, \Delta_i = 0$. Also, the second term $\fr{1}{N}\del{\log\del{\fr{\abs{\Bcalp}}{\abs{\Bcals}}}}^2 = 0$ since $\log\del{\fr{\abs{\Bcalp}}{\abs{\Bcals}}} = 0$. Also, we have $\fr{1}{\abs{\Bcals}N} = \fr{1}{\Kstar N}$.

  Therefore, from from Corollary \ref{cor:haver_equal_arm_pulls_generic}, HAVER achieves
  \begin{align*}
    \MSE(\hmuhaver) = \tilcO \del{\fr{1}{\Kstar N}} + \tilcO \del{\fr{1}{KN}} = \tilcO \del{\fr{1}{\Kstar N}}.
  \end{align*}
  
\end{proof}

\subsection{Proof of Corollary ~\ref{cor:haver_equal_arm_pulls_alpha_poly}}
\label{cor_proof:haver_equal_arm_pulls_alpha_poly}

\newtheorem*{cor-3}{Corollary ~\ref{cor:haver_equal_arm_pulls_alpha_poly}}
\begin{cor-3}
  Under Assumption~\ref{ass:equal_arm_pulls} (equal number of samples), consider the Poly($\alpha$) instance where $\forall i \ge 2,\, \Delta_i = \del{\fr{i}{K}}^{\alpha}$ where $\alpha \ge 0$. If $N \le \fr{1}{2}\log(K^2N)\del{\fr{K}{\alpha}\log(2)}^{2\alpha}$, HAVER achieves
  \begin{align*}
    \MSE(\hmuhaver) = \tilcO \del{\fr{1}{(\alpha \wedge K) N}}.
  \end{align*}
\end{cor-3}

\begin{proof}
  From Corollary \ref{cor:haver_equal_arm_pulls_generic}, we have $\Bcals = \cbr{i \in [K]: \Delta_i \le \cstar \gam}$ and $\Bcalp = \cbr{i \in [K]: \Delta_i \le \cplusq \gam}$, and
  \begin{align*}
    &\MSE(\hmuhaver) \\
    \stackrel{}{=}\,& \tilcO \del{\del{\fr{1}{\abs{\cBs}}\sum_{i = 1}^{\abs{\cBp}}\Delta_i}^2 \wedge {\fr{1}{N}}} \\
    &+ \tilcO \del{\del{\fr{1}{N}\del{\log\del{\fr{\abs{\cBp}}{\abs{\cBs}}}}^2} \wedge {\fr{1}{N}}} \\
    &+ \tilcO \del{\fr{1}{\abs{\cBs}N}} + \tilcO \del{\fr{1}{KN}}.
  \end{align*}

  We observe the condition in set $\cBs$, $\forall i \in [K]$ 
  \begin{align*}
    \Delta_i
    \stackrel{}{\le}\, \fr{1}{6}\gam 
    \stackrel{}{=}\, \fr{1}{6}\sqrt{\fr{18}{N}\log\del{\del{K^2N}^4}} 
    \stackrel{}{=}\, \sqrt{2\log\del{K^2N}}\sqrt{\fr{1}{N}} 
  \end{align*}
  We denote $a_1 = \sqrt{2\log\del{K^2N}}$. We rewrite $\cBs = \cbr{\forall i \in [K]: a_1\sqrt{\fr{1}{N}}}$
  We observe the condition in set $\cBp$, $\forall i \in [K]$ 
  \begin{align*}
    \Delta_i 
    \stackrel{}{\le}\, \fr{8}{3}\gam 
    \stackrel{}{=}\, \fr{8}{3}\sqrt{\fr{18}{N}\log\del{\del{K^2N}^4}} 
    \stackrel{}{=}\, 16\sqrt{2\log\del{K^2N}}\sqrt{\fr{1}{N}}. 
  \end{align*}
  We denote $a_2 = 16\sqrt{2\log\del{K^2N}}$. We rewrite $\cBp = \cbr{\forall i \in [K]: a_2\sqrt{\fr{1}{N}}}$. We have the ratio $\fr{2a_2}{a_1} = 32$.
  The assumption ${N \le \fr{1}{2}\log(K^2N)\del{\fr{K}{\alpha}\log(2)}^{2\alpha}}$ satisfies ${N \le \del{\fr{a_1^2}{4} \wedge a_2} \del{\fr{K}{\alpha}\log(2)}^{2\alpha}}$.
  Within this context, we can use Lemma~\ref{lemma302:bcals}.
  For the first term, we use Lemma~\ref{lemma302:bcals} to bound
  \begin{align*}
    &\del{\displaystyle \fr{1}{\abs{\Bcals}}\sum_{i = 1}^{\abs{\Bcalp}}\Delta_i}^2 \\
    \stackrel{}{\le}\,& \del{\fr{2a_2\del{\fr{4a_2}{a_1}}^{\fr{1}{\alpha}}}{\del{\alpha+1}\sqrt{N}}}^2 \\
    \stackrel{}{\le}\,& \del{\fr{32\sqrt{2\log\del{K^2N}}\del{64}^{\fr{1}{\alpha}}}{\del{\alpha+1}\sqrt{N}}}^2 \\
    \stackrel{}{=}\,& \fr{2048\log\del{K^2N}\del{64}^{\fr{2}{\alpha}}}{\del{\alpha+1}^2N}.
  \end{align*}

  For the second term, we use Lemma~\ref{lemma302:bcals} to bound 
  \begin{align*}
    \fr{\abs{\Bcalp}}{\abs{\Bcals}} \le \del{\fr{2a_2}{a_1}}^{\fr{1}{\alpha}} = 32^{\fr{1}{\alpha}}.
  \end{align*}
  Hence,
  \begin{align*}
    \fr{1}{N}\del{\log\del{\fr{\abs{\Bcalp}}{\abs{\Bcals}}}}^2 \le \fr{1}{N}\del{\log\del{32^{\fr{1}{\alpha}}}}^2 = \fr{1}{\alpha^2N}\log^2(32)
  \end{align*}
  For the third term, we use Lemma~\ref{lemma302:bcals} to bound
  \begin{align*}
    \fr{1}{\abs{\cBs}} \le \fr{1}{K\del{\fr{\sqrt{2\log\del{K^2N}}}{2}\sqrt{\fr{1}{N}}}^{\fr{1}{\alpha}}}
  \end{align*}
  From the assumption ${N \le \fr{1}{2}\log(K^2N)\del{\fr{K}{\alpha}\log(2)}^{2\alpha}}$, we have
  \begin{align*}
    N &\le \fr{1}{2}\log(K^2N)\del{\fr{K}{\alpha}\log(2)}^{2\alpha} \\
    \stackrel{}{\Leftrightarrow}\, \sqrt{\fr{1}{N}} &\ge \fr{1}{\sqrt{\fr{1}{2}\log(K^2N)\del{\fr{K}{\alpha}\log(2)}^{2\alpha}}} \\
    \stackrel{}{\Leftrightarrow}\, \fr{\sqrt{2\log\del{K^2N}}}{2}\sqrt{\fr{1}{N}} &\ge \fr{\fr{\sqrt{2\log\del{K^2N}}}{2}}{\sqrt{\fr{1}{2}\log(K^2N)\del{\fr{K}{\alpha}\log(2)}^{2\alpha}}} \\
    \stackrel{}{\Leftrightarrow}\, \fr{\sqrt{2\log\del{K^2N}}}{2}\sqrt{\fr{1}{N}} &\ge \fr{1}{\del{\fr{K}{\alpha}\log(2)}^{\alpha}} \\
    \stackrel{}{\Leftrightarrow}\, \del{\fr{\sqrt{2\log\del{K^2N}}}{2}\sqrt{\fr{1}{N}}}^{\fr{1}{\alpha}} &\ge \fr{1}{\fr{K}{\alpha}\log(2)} \\
    \stackrel{}{\Leftrightarrow}\, K\del{\fr{\sqrt{2\log\del{K^2N}}}{2}\sqrt{\fr{1}{N}}}^{\fr{1}{\alpha}} &\ge \fr{\alpha}{\log(2)}. 
  \end{align*}
  Therefore,
  \begin{align*}
    \fr{1}{\abs{\cBs}N} \le \fr{1}{K\del{\fr{\sqrt{2\log\del{K^2N}}}{2}\sqrt{\fr{1}{N}}}^{\fr{1}{\alpha}}N} \le \fr{\log(2)}{\alpha N}.
  \end{align*}

  Combining all the terms, we have
  \begin{align*}
    &\MSE(\hmuhaver) \\
    \stackrel{}{=}\,& \tilcO \del{\del{\fr{1}{\abs{\cBs}}\sum_{i = 1}^{\abs{\cBp}}\Delta_i}^2 \wedge {\fr{1}{N}}} \\
    &+ \tilcO \del{\del{\fr{1}{N}\del{\log\del{\fr{\abs{\cBp}}{\abs{\cBs}}}}^2} \wedge {\fr{1}{N}}} \\
    &+ \tilcO \del{\fr{1}{\abs{\cBs}N}} + \tilcO \del{\fr{1}{KN}} \\
    \stackrel{}{=}\,& \tilcO \del{\fr{2048\log\del{K^2N}\del{64}^{\fr{2}{\alpha}}}{\del{\alpha+1}^2N} \wedge \del{\fr{1}{N}}} \\
    &+ \tilcO \del{\fr{1}{\alpha^2N}\log^2(32) \wedge \del{\fr{1}{N}}} \\
    &+ \tilcO \del{\fr{\log(2)}{\alpha N}} + \tilcO \del{\fr{1}{KN}} \\
    \stackrel{}{=}\,& \tilcO \del{\fr{1}{(\alpha \wedge K) N}}.
  \end{align*}
  
\end{proof}

\subsection{Proof of Corollary ~\ref{cor:haver_non_equal_arm_pulls_allbest}}
\label{cor_proof:haver_non_equal_arm_pulls_allbest}

\newtheorem*{cor-4}{Corollary ~\ref{cor:haver_non_equal_arm_pulls_allbest}}
\begin{cor-4}
  Consider the all-best instance where $\forall i \in [K],\, \mu_i = \mu_{1}$ and the number of samples are characterized by $N_i = \del{K - i + 1}^{\beta}$ where $\beta \in (0,1)$,  HAVER achieves
  \begin{align*}
    \MSE(\hmuhaver) = \tilcO\del{\fr{1}{K N_1}}.
  \end{align*}
\end{cor-4}

\begin{proof}
  In this instance, we have $s = \argmax_{i \in [K]} \mu_i - \cs \gam_i = 1$. The reasons are (1) $\forall i \in [K],\, \mu_i = \mu_1$ in the all-best instance and (2) the number of samples $N_i = \del{K - i + 1}^{\beta}$ decreases and $\gam_i$ increases from arm 1 to $K$, thus $\gam_1$ is the smallest. 
  From Theorem~\ref{thm:haver21count2}, we have
  \begin{align*}
    \cBs(r) = \cbr{i \in [K]: \mu_{i} \ge \mu_{1} - \cstar\gam_{1},\, \gam_i \le \fgam \gam_{r}} = \cbr{i \in [K]: \Delta_{i} \le \cstar\gam_{1},\, \gam_i \le \fgam \gam_{r}}
  \end{align*}
  and
  \begin{align*}
    \cBp(r) := \cbr{i \in [K]: \mu_{i} \ge \mu_{1} - \cplus\gam_{1} - \cplus\gam_{i},\, \gam_i \le \fgam \gam_{r}} = \cbr{i \in [K]: \Delta_{i} \le \cplus\gam_{i} + \cplus\gam_{1},\, \gam_i \le \fgam \gam_{r}}.
  \end{align*}
  In the all-best instance, we have $\forall i \in [K],\, \Delta_i = 0$. Therefore, in this instance, for every $r \in \cR$, the first condition of $\cBs(r)$, $\forall i \in [K],\, \Delta_{i} \le \cstar\gam_{1}$ is always satisfied and the first condition of $\cBp(r)$, $\forall i \in [K],\, \Delta_{i} \le \cplus\gam_{i} + \cplus\gam_{1}$, is always satisfied. 
  Also, the second conditions of both $\cBs(r)$ and $\cBp(r)$ are the same. Thus, we have $\forall r \in \cR,\, \cBp(r) = \cBs(r)$.

  From Theorem~\ref{thm:haver21count2}, we have
  \begin{align*}
    &\MSE(\hmuhaver) \\
    \stackrel{}{=}\,& \tilcO \del{\del{\displaystyle \max_{r \in \cR} \fr{1}{\sum_{j \in \cBs(r)}N_j}\sum_{i \in \cBp(r)}N_i\Delta_i}^2 \wedge {\fr{1}{N_1}}} \\
    &+ \tilcO \del{\del{\max_{r \in \cR} \max_{k=0}^d \max_{\substack{S: \cBs(r) \subseteq \Scal \subseteq \cBp(r) \\ \abs{\Scal}=n_*(r)+k }}\fr{k\sum_{j \in \Scal \sm \cBs(r)}N_j}{\del{\sum_{j \in \Scal}N_j}^2}} \wedge {\fr{1}{N_1}}} \\
    &+ \tilcO\del{\del{\fr{1}{\min_{r \in \cR}\sum_{j \in \cBs(r)}N_j}} \wedge {\fr{1}{N_1}}} \\
    &+ \tilcO \del{\fr{1}{KN_1}}.
  \end{align*}

  For the first term, $\displaystyle \del{ \max_{r \in \cR} \fr{1}{\sum_{j \in \cBs(r)}N_j}\sum_{i \in \cBp(r)}N_i\Delta_i}^2$ becomes $0$ since $\forall i \in [K],\, \Delta_i = 0$.
  
  For the second term, $\displaystyle \del{\max_{r \in \cR} \max_{k=0}^{d(r)} \max_{\substack{S: \cBs(r) \subseteq \Scal \subseteq \cBp(r) \\ \abs{\Scal}=n_*(r)+k }}\fr{k\sum_{j \in \Scal \sm \cBs(r)}N_j}{\del{\sum_{j \in \Scal}N_j}^2}}$ becomes $0$ since $\forall r \in \cR,\, d(r) = \abs{\cBp(r)} - \abs{\cBs(r)} = 0$ as $\cBp(r) = \cBs(r)$.

  For the third term, we have 
  \begin{align*}
    \fr{1}{\min_{r \in \cR}\sum_{j \in \cBs(r)}N_j} = \fr{1}{\sum_{j \in \cBs(1)}N_j}. 
  \end{align*}
  Because the first condition of set $\cBs$ is always satisfied, set $\cBs$ becomes just a matter of arm counts condition. Therefore, $r = 1$ results in minimal number of arms in $\cBs(r)$ because (1) the optimal arm $1$ is always in $\cR$ and (2) among arm $r \in \cR$, arm $r = 1$ has the highest number of samples making the condition $\gam_i \le \fgam \gam_r$ most restrictive. 
  An arm $i$ is included in $\cBs(1)$ if it satisfies
  \begin{align*}
    &\gam_i \le \fgam \gam_1 \\
    \stackrel{}{\Leftrightarrow}\,& \fr{18}{N_i}\log\del{\del{\fr{KS}{N_i}}^{2\lam}} \le \fr{9}{4}\fr{18}{N_1}\log\del{\del{\fr{KS}{N_1}}^{2\lam}} \\
    \stackrel{}{\Leftrightarrow}\,& \fr{1}{N_i}\log\del{\fr{KS}{N_i}} \le \fr{9}{4}\fr{1}{N_1}\log\del{\fr{KS}{N_1}} \\
    \stackrel{}{\Leftrightarrow}\,& N_i \ge \fr{4}{9}N_1 \fr{\log\del{\fr{KS}{N_i}}}{\log\del{\fr{KS}{N_1}}} \\
    \stackrel{}{\Rightarrow}\,& N_i \ge \fr{4}{9}N_1 \tagcmt{use $N_i \le N_1$} \\
    \stackrel{}{\Leftrightarrow}\,& \del{K-i+1}^{\beta} \ge \fr{4}{9}K^{\beta} \\
    \stackrel{}{\Leftrightarrow}\,& K-i+1 \ge \del{\fr{4}{9}}^{\fr{1}{\beta}}K \\
    \stackrel{}{\Leftrightarrow}\,& i \le K\del{1 - \del{\fr{4}{9}}^{\fr{1}{\beta}}}.
  \end{align*}
  Thus, we have
  \begin{align*}
    \sum_{j \in \cBs(1)}N_j \ge \sum_{j = 1}^{K\del{1 - \del{\fr{4}{9}}^{\fr{1}{\beta}}}}N_j.
  \end{align*}
  Let $\omega = 1 - \del{\fr{4}{9}}^{\fr{1}{\beta}}$. We have 
  \begin{align*}
    \sum_{j = 1}^{K\del{1 - \del{\fr{4}{9}}^{\fr{1}{\beta}}}}N_j 
    \stackrel{}{=}\, \sum_{j = 1}^{\omega K}N_j 
    \stackrel{}{=}\, \sum_{j = 1}^{\omega K}(K-j+1)^{\beta} 
    \stackrel{}{\ge}\, \int_{j = 1}^{\omega K}(K-j+1)^{\beta} \dif j. 
  \end{align*}
  We do change of variables of the integral. Let $u = K-j+1$, then $\dif u = -\dif j$. When $j = 1$, $u = K$. When $j = \omega K$, $u = K - \omega K$. The integral becomes
  \begin{align*}
    &\int_{j = 1}^{\omega K}(K-j+1)^{\beta} \dif j \\
    \stackrel{}{=}\,& \int_{j = K}^{K - \omega K} -u^{\beta} \dif u \\
    \stackrel{}{=}\,& \int_{j = K - \omega K}^{K} u^{\beta} \dif u \\
    \stackrel{}{=}\,& \del{\fr{u^{\beta+1}}{\beta+1}}\Big|_{K - \omega K}^{K} \\
    \stackrel{}{=}\,& \fr{K^{\beta+1} - \del{K - \omega K}^{\beta+1}}{\beta+1} \\
    \stackrel{}{=}\,& \fr{K^{\beta+1} - \del{(1-\omega)K}^{\beta+1}}{\beta+1}. 
  \end{align*}
  Consider function $f(x) = x^{\beta + 1}$ for any $x \in \RR$. We have $f'(x) = (\beta + 1)x^{\beta}$ It is trivial to see that $f(x)$ is convex. Thus, we have $f(x) - f(y) \ge f'(y)(x - y)$. We plug in $x = K$ and $y = (1-\omega)K$. We have
  \begin{align*}
    K^{\beta+1} - \del{(1-\omega)K}^{\beta+1} \ge  (\beta + 1)\del{(1-\omega)K}^{\beta} \omega K.
  \end{align*}
  Therefore,
  \begin{align*}
    &\int_{j = 1}^{\omega K}(K-j+1)^{\beta} \dif j \\
    \stackrel{}{=}\,& \fr{K^{\beta+1} - \del{(1-\omega)K}^{\beta+1}}{\beta+1} \\
    \stackrel{}{\ge}\,& \fr{(\beta + 1)\del{(1-\omega)K}^{\beta} \omega K}{\beta+1} \\
    \stackrel{}{=}\,& \del{(1-\omega)K}^{\beta} \omega K. 
  \end{align*}
  Plugging $\omega = 1 - \del{\fr{4}{9}}^{\fr{1}{\beta}}$, we have
  \begin{align*}
    &\del{(1-\omega)K}^{\beta} \omega K \\
    \stackrel{}{=}\,& \del{\del{1 - 1 + \del{\fr{4}{9}}^{\fr{1}{\beta}}}K}^{\beta} \del{1 - \del{\fr{4}{9}}^{\fr{1}{\beta}}} K \\
    \stackrel{}{=}\,& \fr{4}{9} \del{1 - \del{\fr{4}{9}}^{\fr{1}{\beta}}} K^{\beta + 1} \\
    \stackrel{}{\ge}\,& \fr{4}{9}\fr{5}{9}K^{\beta+1} \tagcmt{use $(1 - \del{\fr{4}{9}}^{\fr{1}{\beta}}) > \fr{5}{9}$ with $\beta \in (0,1)$ } \\
    \stackrel{}{=}\,& \fr{20}{81}K^{\beta+1}.
  \end{align*}
  Therefore,
  \begin{align*}
    \sum_{j \in \cBs(r)}N_j \ge \sum_{j = 1}^{K\del{1 - \del{\fr{4}{9}}^{\fr{1}{\beta}}}}N_j \ge \int_{j = 1}^{\omega K}(K-j+1)^{\beta} \dif j \ge \fr{4}{27}K^{\beta+1}.
  \end{align*}
  Therefore, for the third term, we have
  \begin{align*}
    &\fr{1}{\min_{r \in \cR}\sum_{j \in \cBs(r)}N_j} \\
    \stackrel{}{=}\,& \fr{1}{\sum_{j \in \cBs(1)}N_j} \\
    \stackrel{}{\le}\,& \fr{1}{\fr{20}{81}K^{\beta+1}} \\
    \stackrel{}{=}\,& \fr{81}{20}\del{\fr{1}{KN_1}} \tagcmt{use $N_1 = K^{\beta}$}.
  \end{align*}

  Combining all the terms, we have
  \begin{align*}
    &\MSE(\hmuhaver) \\
    \stackrel{}{=}\,& \tilcO \del{\del{\displaystyle \max_{r \in \cR} \fr{1}{\sum_{j \in \cBs(r)}N_j}\sum_{i \in \cBp(r)}N_i\Delta_i}^2 \wedge {\fr{1}{N_1}}} \\
    &+ \tilcO \del{\del{\max_{r \in \cR} \max_{k=0}^d \max_{\substack{S: \cBs(r) \subseteq \Scal \subseteq \cBp(r) \\ \abs{\Scal}=n_*(r)+k }}\fr{k\sum_{j \in \Scal \sm \cBs(r)}N_j}{\del{\sum_{j \in \Scal}N_j}^2}} \wedge {\fr{1}{N_1}}} \\
    &+ \tilcO\del{\del{\fr{1}{\min_{r \in \cR}\sum_{j \in \cBs(r)}N_j}} \wedge {\fr{1}{N_1}}} \\
    &+ \tilcO \del{\fr{1}{KN_1}} \\
    \stackrel{}{=}\,& \tilcO\del{\del{\fr{1}{K N_1}} \wedge {\fr{1}{N_1}}} \\
    &+ \tilcO \del{\fr{1}{KN_1}} \\
    \stackrel{}{=}\,& \tilcO\del{\fr{1}{K N_1}}.
  \end{align*}

\end{proof}

\clearpage
\section{HAVER's Theorem ~\ref{thm:haver21count2}}
\label{thm_proof:haver21count2}

Recall the following definitions that would be used in HAVER.

For each arm $i \in [K]$, we define 
\begin{align*}
  \gam_{i}: = \sqrt{\fr{18}{N_{i}}\log\del{\del{\fr{KS}{N_i}}^{2\lam}}}, 
\end{align*}
as its confidence width where $S = N_{\max}\sum_{j \in [K]}N_j$ and $\lam = 2$. We define 
\begin{align*}
  \hr := \argmax_{i \in [K]} \hmu_i - \gam_i
\end{align*}
as the pivot arm with maximum lower confidence bound, and define
\begin{align*}
  s := \argmax_{i \in [K]} \mu_i - \cs\gam_i
\end{align*}
as the ground truth version of the pivot arm. We define 
\begin{align*}
  \cB := \cbr{i \in [K]: \hmu_{i} \ge \hmu_{\hr} - \gam_{\hr},\, \gam_{i} \le \fgam \gam_{\hr}}
\end{align*}
as the candidate set of arms. We define 
\begin{align*}
  \cR :=  \cbr{r \in [K]: \mu_s - \fr{4}{3}\gam_s + \fr{2}{3}\gam_r \le \mu_r \le \mu_s - \cs\gam_s + \cs\gam_r}
\end{align*}
as the statistically-plausible set of pivot arms $\hr$. For any $r \in \cR$, we define 
\begin{align*}
  \cBs(r) := \cbr{i \in [K]: \mu_{i} \ge \mu_{s} - \cstar\gam_{s},\, \gam_i \le \fgam \gam_{r}},
\end{align*}
\begin{align*}
  \cBp(r) := \cbr{i \in [K]: \mu_{i} \ge \mu_{s} - \cplus\gam_{s} - \cplus\gam_{i},\, \gam_i \le \fgam \gam_{r}},
\end{align*}
$d(r) = \abs{\cBp(r)} - \abs{\cBs(r)}$, and $n_*(r) = \abs{\cBs(r)}$.

In addition, throughout the proofs, we define $(x)_+ = \max\del{0,x}$.

\newtheorem*{thm-1}{Theorem ~\ref{thm:haver21count2}}
\begin{thm-1}
  HAVER achieves
  \begin{align*}
    &\MSE(\hmuhaver) \\
    \stackrel{}{=}\,& \tilcO \del{\del{\displaystyle \max_{r \in \cR} \fr{1}{\sum_{j \in \cBs(r)}N_j}\sum_{i \in \cBp(r)}N_i\Delta_i}^2 \wedge {\fr{1}{N_1}}} \\
    &+ \tilcO \del{\del{\max_{r \in \cR} \max_{k=0}^{d(r)} \max_{\substack{S: \cBs(r) \subseteq \Scal \subseteq \cBp(r) \\ \abs{\Scal}=n_*(r)+k }}\fr{k\sum_{j \in \Scal \sm \cBs(r)}N_j}{\del{\sum_{j \in \Scal}N_j}^2}} \wedge {\fr{1}{N_1}}} \\
    &+ \tilcO \del{\del{\fr{1}{\min_{r \in \cR}\sum_{j \in \cBs(r)}N_j}} \wedge {\fr{1}{N_1}}} \\
    &+ \tilcO \del{\fr{1}{KN_1}},
  \end{align*}
  where $d(r) = \abs{\cBp(r)} - \abs{\cBs(r)}$ and $n_*(r) = \abs{\cBs(r)}$.
\end{thm-1}

\begin{proof}
  We have
  \begin{align*}
    &\MSE(\hmuhaver) \\
    \stackrel{}{=}\,& \EE\sbr{\del{\hmuB - \mu_1}^2} \\
    \stackrel{}{=}\,& \int_{0}^{\infty} \PP\del{ \del{\hmuB - \mu_1}^2 > \eps }  \dif \eps  \\
    \stackrel{}{=}\,& \int_{0}^{\infty} \PP\del{ \abs{\hmuB - \mu_1} > \eps } \difeps \tagcmt{change of variable} \\
    \stackrel{}{=}\,& \int_{0}^{\infty} \PP\del{ \abs{\hmuB - \mu_1} > \eps,\, \exists i,\, \abs{\hmu_i - \mu_i} \ge \qgam \gam_i} \difeps \\
    &+ \int_{0}^{\infty} \PP\del{ \abs{\hmuB - \mu_1} > \eps,\, \forall i,\, \abs{\hmu_i - \mu_i} < \qgam \gam_i} \difeps.
  \end{align*}
  For the first term, we use Lemma~\ref{lemma200.a:haver21count2} to obtain
  \begin{align*}
    \int_{0}^{\infty} \PP\del{ \abs{\hmuB - \mu_1} > \eps,\, \exists i,\, \abs{\hmu_i - \mu_i} \ge \qgam \gam_i} \difeps = \tilcO \del{\fr{1}{KN_1}}.
  \end{align*}

  For the second term, we claim that the condition
  \begin{align}\label{eq:cr_cond}
    \forall i,\, \abs{\hmu_i - \mu_i} < \qgam \gam_i 
  \end{align}
  implies that $\hr \in \cR$. To see this, by the definition of $\hr$, we have
  \begin{align*}
    &\hmu_{\hr} - \hmu_s \ge \gam_{\hr} - \gam_{s} \\
    \stackrel{}{\Rightarrow}\,& \mu_{\hr} + \qgam \gam_{\hr} - \mu_{s} + \qgam \gam_{s} \ge \gam_{\hr} - \gam_{s} \tagcmt{use the condition \eqref{eq:cr_cond}} \\
    \stackrel{}{\Leftrightarrow}\,& \mu_{\hr} \ge \mu_s + \fr{2}{3}\gam_{\hr} - \fr{4}{3}\gam_{s}. 
  \end{align*}
  In addition, by definition of $s$, we have
  \begin{align*}
    \mu_{\hr} \le \mu_{s} - \cs\gam_{s} + \cs\gam_{\hr}.
  \end{align*}
  These conditions imply that $\hr \in \cR = \cbr{r \in [K]: \mu_s - \fr{4}{3}\gam_s + \fr{2}{3}\gam_r \le \mu_r \le \mu_s - \cs\gam_s + \cs\gam_r}$.

  With $\hr \in \cR$, we decompose the second term 
  \begin{align*}
    &\int_{0}^{\infty} \PP\del{ \abs{\hmuB - \mu_1} < \eps, \hr \in \cR} \difeps \\
    \stackrel{}{=}\,& \int_{0}^{\infty} \PP\del{ \hmuB - \mu_1 < -\eps,\, \hr \in \cR} \difeps \\
    &+ \int_{0}^{\infty} \PP\del{ \hmuB - \mu_1 > \eps,\, \hr \in \cR} \difeps.
  \end{align*}
  For the first subterm, we use Lemma~\ref{lemma_ga:haver21count2} to obtain
  \begin{align*}
    &\int_{0}^{\infty} \PP\del{ \hmuB - \mu_1 < -\eps,\, \hr \in \cR} \difeps \\
    \stackrel{}{=}\,& \tilcO \del{\del{\displaystyle \max_{r \in \cR} \fr{1}{\sum_{j \in \cBs(r)}N_j}\sum_{i \in \cBp(r)}N_i\Delta_i}^2 \wedge {\fr{1}{N_1}}} \\
    &+ \tilcO \del{\del{\max_{r \in \cR} \max_{k=0}^{d(r)} \max_{\substack{S: \cBs(r) \subseteq \Scal \subseteq \cBp(r) \\ \abs{\Scal}=n_*(r)+k }}\fr{k\sum_{j \in \Scal \sm \cBs(r)}N_j}{\del{\sum_{j \in \Scal}N_j}^2}} \wedge {\fr{1}{N_1}}} \\
    &+ \tilcO\del{\del{\fr{1}{\min_{r \in \cR}\sum_{j \in \cBs(r)}N_j}} \wedge {\fr{1}{N_1}}} \\
    &+ \tilcO \del{\fr{1}{KN_1}}.
  \end{align*}
  For the second subterm, we use Lemma~\ref{lemma_gb:haver21count2} to obtain
  \begin{align*}
    &\int_{0}^{\infty} \PP\del{ \hmuB - \mu_1 > \eps,\, \hr \in \cR} \difeps \\
    \stackrel{}{=}\,& \tilcO \del{\del{\max_{r \in \cR} \max_{k=0}^{d(r)} \max_{\substack{S: \cBs(r) \subseteq \Scal \subseteq \cBp(r) \\ \abs{\Scal}=n_*(r)+k }}\fr{k\sum_{j \in \Scal \sm \cBs(r)}N_j}{\del{\sum_{j \in \Scal}N_j}^2}} \wedge {\fr{1}{N_1}}} \\
    &+ \tilcO \del{\del{\fr{1}{\min_{r \in \cR}\sum_{j \in \cBs(r)}N_j}} \wedge {\fr{1}{N_1}}}  \\
    &+ \tilcO \del{\fr{1}{KN_1}}.
  \end{align*}
  Combining all the results completes our proof. %

\end{proof}

\begin{lemma}\label{lemma_ga:haver21count2}
  HAVER achieves
  \begin{align*}
    &\int_{0}^{\infty} \PP\del{ \hmuB - \mu_1 < -\eps,\, \hr \in \cR} \difeps \\
    \stackrel{}{=}\,& \tilcO \del{\del{\max_{r \in \cR} \fr{1}{\sum_{j \in \cBs(r)}N_j}\sum_{i \in \cBp(r)}N_i\Delta_i}^2 \wedge {\fr{1}{N_1}}} \\
    &+ \tilcO \del{\del{\max_{r \in \cR} \max_{k=0}^{d(r)} \max_{\substack{S: \cBs(r) \subseteq \Scal \subseteq \cBp(r) \\ \abs{\Scal}=n_*(r)+k }}\fr{k\sum_{j \in \Scal \sm \cBs(r)}N_j}{\del{\sum_{j \in \Scal}N_j}^2}} \wedge {\fr{1}{N_1}}} \\
    &+ \tilcO \del{\del{\fr{1}{\min_{r \in \cR}\sum_{j \in \cBs(r)}N_j}} \wedge {\fr{1}{N_1}}} \\
    &+ \tilcO \del{\fr{1}{KN_1}},
  \end{align*}
  where $d(r) = \abs{\cBp(r)} - \abs{\cBs(r)}$ and $n_*(r) = \abs{\cBs(r)}$.
\end{lemma}

\begin{proof}
  We consider 4 complementary events:
  \begin{itemize}
  \item $G_0 = \cbr{\cB = \cBs(\hr)}$
  \item $G_1 = \cbr{\cBs(\hr) \subset \cB \subseteq \cBp(\hr)}$
  \item $G_2 = \cbr{\exists i \in \cBs(\hr)\, \textrm{s.t.}\,  i \not\in \cB}$
  \item $G_3 = \cbr{\cBs(\hr) \subseteq \cB,\, \exists i \not\in \cBp(\hr)\, \textrm{s.t.}\, i \in \cB}$
  \end{itemize}
  We have 
  \begin{align*}
    &\int_{0}^{\infty} \PP\del{ \hmuB - \mu_1 < -\eps,\, \hr \in \cR} \difeps \\
    \stackrel{}{=}\,& \int_{0}^{\infty} \PP\del{ \hmuB - \mu_1 < -\eps,\, \hr \in \cR,\, G_0} \difeps + \int_{0}^{\infty} \PP\del{ \hmuB - \mu_1 < -\eps,\, \hr \in \cR,\, G_1} \difeps \\
    &+ \int_{0}^{\infty} \PP\del{ \hmuB - \mu_1 < -\eps,\, \hr \in \cR,\, G_2} \difeps + \int_{0}^{\infty} \PP\del{ \hmuB - \mu_1 < -\eps,\, \hr \in \cR,\, G_3} \difeps.
  \end{align*}
  Using Lemma~\ref{lemma_g0a:haver21count2} and ~\ref{lemma_g1a:haver21count2} for event $G_0$ and $G_1$ respectively, we have 
  \begin{align*}
    &\int_{0}^{\infty} \PP\del{ \hmuB - \mu_1 < -\eps,\, \hr \in \cR,\, G_0} \difeps \\
    \stackrel{}{=}\,& \cO\del{\del{\max_{r \in \cR}\fr{1}{\sum_{j \in \cBs(r)}N_j}\sum_{i \in \cBs(r)}N_i\Delta_i}^2 + \fr{\log(K)}{\min_{r \in \cR}\sum_{j \in \cBs(r)}N_j}}
  \end{align*}
  and 
  \begin{align*}
    &\int_{0}^{\infty} \PP\del{ \hmuB - \mu_1 < -\eps,\, \hr \in \cR,\, G_1} \difeps \\
    \stackrel{}{=}\,& \cO \del{\del{\displaystyle \max_{r \in \cR} \max_{\substack{S: \cBs(r) \subseteq \Scal \subseteq \cBp(r) }}\fr{1}{\sum_{j \in \Scal}N_j}\sum_{i \in \Scal}N_i\Delta_i}^2} \\
    &+ \cO \del{\max_{r \in \cR} \max_{k=0}^{d(r)} \max_{\substack{S: \cBs(r) \subseteq \Scal \subseteq \cBp(r) \\ \abs{\Scal}=n_*(r)+k }}\fr{k\sum_{j \in \Scal \sm \cBs(r)}N_j}{\del{\sum_{j \in \Scal}N_j}^2}\log\del{\fr{ed(r)}{k}}\log(K)} \\
    &+ \cO\del{\fr{\log(K)}{\min_{r \in \cR}\sum_{j \in \cBs(r)}N_j}}. 
  \end{align*}
  Combining the first two events, we have
  \begin{align*}
    &\int_{0}^{\infty} \PP\del{ \hmuB - \mu_1 < -\eps,\, \hr \in \cR,\, G_0} \difeps \\
    &+ \int_{0}^{\infty} \PP\del{ \hmuB - \mu_1 < -\eps,\, \hr \in \cR,\, G_1} \difeps \\
    \stackrel{}{=}\,& \cO \del{\del{\displaystyle \max_{r \in \cR} \max_{\substack{S: \cBs(r) \subseteq \Scal \subseteq \cBp(r) }}\fr{1}{\sum_{j \in \Scal}N_j}\sum_{i \in \Scal}N_i\Delta_i}^2} \\
    &+ \cO \del{\max_{r \in \cR} \max_{k=0}^{d(r)} \max_{\substack{S: \cBs(r) \subseteq \Scal \subseteq \cBp(r) \\ \abs{\Scal}=n_*(r)+k }}\fr{k\sum_{j \in \Scal \sm \cBs(r)}N_j}{\del{\sum_{j \in \Scal}N_j}^2}\log\del{\fr{ed(r)}{k}}\log(K)} \\
    &+ \cO\del{\fr{\log(K)}{\min_{r \in \cR}\sum_{j \in \cBs(r)}N_j}} \\
    \stackrel{}{=}\,& \cO \del{\del{\displaystyle \max_{r \in \cR} \fr{1}{\sum_{j \in \cBs(r)}N_j}\sum_{i \in \cBp(r)}N_i\Delta_i}^2} \\
    &+ \cO \del{\max_{r \in \cR} \max_{k=0}^{d(r)} \max_{\substack{S: \cBs(r) \subseteq \Scal \subseteq \cBp(r) \\ \abs{\Scal}=n_*(r)+k }}\fr{k\sum_{j \in \Scal \sm \cBs(r)}N_j}{\del{\sum_{j \in \Scal}N_j}^2}\log\del{3K}\log(K)} \tagcmt{use $\fr{ed(r)}{k} \le 3K$} \\
    &+ \cO\del{\fr{\log(K)}{\min_{r \in \cR}\sum_{j \in \cBs(r)}N_j}} \\
    \stackrel{}{=}\,& \tilcO \del{\del{\displaystyle \max_{r \in \cR} \fr{1}{\sum_{j \in \cBs(r)}N_j}\sum_{i \in \cBp(r)}N_i\Delta_i}^2} \\
    &+ \tilcO \del{\max_{r \in \cR} \max_{k=0}^{d(r)} \max_{\substack{S: \cBs(r) \subseteq \Scal \subseteq \cBp(r) \\ \abs{\Scal}=n_*(r)+k }}\fr{k\sum_{j \in \Scal \sm \cBs(r)}N_j}{\del{\sum_{j \in \Scal}N_j}^2}} \\
    &+ \tilcO \del{\fr{1}{\min_{r \in \cR}\sum_{j \in \cBs(r)}N_j}}.
  \end{align*}
  We use Lemma~\ref{lemma_g2:haver21count2} and ~\ref{lemma_g3:haver21count2} to show that $\PP\del{G_2} \le \fr{2K^2}{K^{2\lam}}$ and $\PP\del{G_3} \le \fr{2K^2}{K^{2\lam}}$ respectively.

  Consequently, we use Lemma~\ref{lemma_g2a:haver21count2} for event $G_2$ to obtain the integral
  \begin{align*}
    &\int_{0}^{\infty} \PP\del{ \hmuB - \mu_1 < -\eps,\, \hr \in \cR,\, G_2} \difeps \\
    \stackrel{}{=}\,& \cO \del{\min_m \del{\fr{1}{K^2N_m}\log\del{\fr{KS}{N_m}} + \fr{1}{K^{2}}\Delta_m^2}} \\
    \stackrel{}{\le}\,& \cO \del{\del{\fr{1}{K^2N_1}\log\del{\fr{KS}{N_1}}}} \tagcmt{upper bounded by $m = 1$} \\
    \stackrel{}{\le}\,& \tilcO \del{\fr{1}{KN_1}}.
  \end{align*}
  We use Lemma~\ref{lemma_g2a:haver21count2} for event $G_3$ and get the same result. Furthermore, Lemma~\ref{lemma102.a:haver} states that 
  \begin{align*}
    &\int_{0}^{\infty} \PP\del{ \hmuB - \mu_1 < -\eps} \difeps \\
    \stackrel{}{=}\,& \cO \del{\min_m \del{\fr{1}{N_m}\log\del{\fr{KS}{N_m}} + \Delta_m^2}} \\
    \stackrel{}{=}\,& \cO \del{\fr{1}{N_1}\log\del{\fr{KS}{N_1}}} \tagcmt{upper bounded by $m = 1$} \\
    \stackrel{}{=}\,& \tilcO \del{\fr{1}{N_1}}.
  \end{align*}

  Combining all the results completes our proof.

\end{proof}


\begin{lemma}\label{lemma_g0a:haver21count2}
  In the event of $G_0 = \cbr{\cB = \cBs(\hr)}$, HAVER achieves
  \begin{align*}
    &\int_{0}^{\infty} \PP\del{ \hmuB - \mu_1 < -\eps,\, \hr \in \cR,\, G_0 } \difeps \\
    \stackrel{}{\le}\,& 4\del{\max_{r \in \cR}\fr{1}{\sum_{j \in \cBs(r)}N_j}\sum_{i \in \cBs(r)}N_i\Delta_i}^2 + 48\del{\fr{\log(K)}{\min_{r \in \cR}\sum_{j \in \cBs(r)}N_j}}.
  \end{align*}
\end{lemma}

\begin{proof}
  We have
  \begin{align*}
    &\PP\del{ \hmuB - \mu_1 < -\eps,\, \hr \in \cR,\, G_0 } \\
    \stackrel{}{=}\,& \PP\del{ \fr{1}{\sum_{j \in \cB}N_j}\sum_{i \in \cB}N_i\hmu_i - \mu_1 < -\eps,\, \hr \in \cR,\, G_0} \\
    \stackrel{}{=}\,& \PP\del{ \fr{1}{\sum_{j \in \cB}N_j}\sum_{i \in \cB}N_i\del{\hmu_i - \mu_1} < -\eps,\, \hr \in \cR,\, G_0} \\
    \stackrel{}{=}\,& \PP\del{ \fr{1}{\sum_{j \in \cB}N_j}\sum_{i \in \cB}N_i\del{\hmu_i - \mu_i} < -\eps + \fr{1}{\sum_{j \in \cB}N_j}\sum_{i \in \cB}N_i\Delta_i,\, \hr \in \cR,\, G_0} \\
    \stackrel{}{=}\,& \PP\del{ \fr{1}{\sum_{j \in \cBs(\hr)}N_j}\sum_{i \in \cBs(\hr)}N_i\del{\hmu_i - \mu_i} < -\eps + \fr{1}{\sum_{j \in \cBs(\hr)}N_j}\sum_{i \in \cBs(\hr)}N_i\Delta_i,\, \hr \in \cR,\, G_0} \tagcmt{use $G_0 = \cbr{\cB = \cBs(\hr)}$} \\
    \stackrel{}{\le}\,& \sum_{r \in \cR}\PP\del{ \fr{1}{\sum_{j \in \cBs(r)}N_j}\sum_{i \in \cBs(r)}N_i\del{\hmu_i - \mu_i} < -\eps + \fr{1}{\sum_{j \in \cBs(r)}N_j}\sum_{i \in \cBs(r)}N_i\Delta_i}  \\
    \stackrel{}{\le}\,& \sum_{r \in \cR} \exp\del{-\fr{1}{2}\del{\sum_{j \in \cBs(r)}N_j}\del{\eps - \fr{1}{\sum_{j \in \cBs(r)}N_j}\sum_{i \in \cBs(r)}N_i\Delta_i}_+^2} \\
    \stackrel{}{\le}\,& K\exp\del{-\fr{1}{2}\del{\min_{r \in \cR} \sum_{j \in \cBs(r)}N_j}\del{\eps - \max_{r \in \cR}\fr{1}{\sum_{j \in \cBs(r)}N_j}\sum_{i \in \cBs(r)}N_i\Delta_i}_+^2}.
  \end{align*}

  Let $q_1 = q_2 = 1$, $z_1 = \fr{1}{16}\del{\min_{r \in \cR}\sum_{j \in \cBs(r)}N_j}$, and
  \begin{align*}
    \eps_0 = \del{2\max_{r \in \cR}\fr{1}{\sum_{j \in \cBs(r)}N_j}\sum_{i \in \cBs(r)}N_i\Delta_i} \vee \del{\sqrt{\fr{16\log(K)}{\min_{r \in \cR}\sum_{j \in \cBs(r)}N_j}}}.
  \end{align*} We claim that, in the regime of $\eps \ge \eps_0$, we have
  \begin{align*}
    \PP\del{ \hmuB - \mu_1 < -\eps,\, \hr \in \cR,\, G_0 } \le q_1\exp\del{-z_1\eps^2}
  \end{align*}
  and in the regime of $\eps < \eps_0$, we have $\PP\del{ \hmuB - \mu_1 < -\eps,\, \hr \in \cR,\, G_0 } \le q_2$.
  
  The second claim is trivial. We prove the first claim as follows. In the regime of $\eps \ge \eps_0$, we have
  \begin{align*}
    &\PP\del{ \hmuB - \mu_1 < -\eps,\, \hr \in \cR,\, G_0 } \\
    \stackrel{}{\le}\,& K\exp\del{-\fr{1}{2}\del{\min_{r \in \cR} \sum_{j \in \cBs(r)}N_j}\del{\eps - \max_{r \in \cR}\fr{1}{\sum_{j \in \cBs(r)}N_j}\sum_{i \in \cBs(r)}N_i\Delta_i}_+^2} \\
    \stackrel{}{\le}\,& K\exp\del{-\fr{1}{2}\del{\min_{r \in \cR} \sum_{j \in \cBs(r)}N_j}\del{\fr{\eps}{2}}^2} \tagcmt{use $\eps \ge \eps_0$} \\
    \stackrel{}{=}\,& K\exp\del{-\fr{1}{8}\del{\min_{r \in \cR} \sum_{j \in \cBs(r)}N_j}\eps^2} \\
    \stackrel{}{=}\,& \exp\del{-\fr{1}{8}\del{\min_{r \in \cR} \sum_{j \in \cBs(r)}N_j}\eps^2 + \log(K)} \\
    \stackrel{}{\le}\,& \exp\del{-\fr{1}{16}\del{\min_{r \in \cR} \sum_{j \in \cBs(r)}N_j}\eps^2} \tagcmt{use $\eps \ge \eps_0$}.
  \end{align*}
  
  With the above claim, we use Lemma~\ref{lemma401:utility_integral_with_eps} to bound the integral
  \begin{align*}
    &\int_{0}^{\infty} \PP\del{ \hmuB - \mu_1 < -\eps,\, \hr \in \cR,\, G_0 } \difeps \\
    \stackrel{}{\le}\,& q_1\eps_0^2 + q_2\fr{1}{z_1} \\
    \stackrel{}{=}\,& \del{\del{2\max_{r \in \cR}\fr{1}{\sum_{j \in \cBs(r)}N_j}\sum_{i \in \cBs(r)}N_i\Delta_i} \vee \del{\sqrt{\fr{16\log(K)}{\min_{r \in \cR}\sum_{j \in \cBs(r)}N_j}}}}^2 + 16\del{\fr{1}{\min_{r \in \cR} \sum_{j \in \cBs(r)}N_j}} \\
    \stackrel{}{\le}\,& 4\del{\max_{r \in \cR}\fr{1}{\sum_{j \in \cBs(r)}N_j}\sum_{i \in \cBs(r)}N_i\Delta_i}^2 + 16\del{\fr{\log(K)}{\min_{r \in \cR}\sum_{j \in \cBs(r)}N_j}} + 16\del{\fr{1}{\min_{r \in \cR} \sum_{j \in \cBs(r)}N_j}} \\
    \stackrel{}{=}\,& 4\del{\max_{r \in \cR}\fr{1}{\sum_{j \in \cBs(r)}N_j}\sum_{i \in \cBs(r)}N_i\Delta_i}^2 + 8\del{\fr{\log(K^2)}{\min_{r \in \cR}\sum_{j \in \cBs(r)}N_j}} + 16\del{\fr{1}{\min_{r \in \cR} \sum_{j \in \cBs(r)}N_j}} \\
    \stackrel{}{\le}\,& 4\del{\max_{r \in \cR}\fr{1}{\sum_{j \in \cBs(r)}N_j}\sum_{i \in \cBs(r)}N_i\Delta_i}^2 + 24\del{\fr{\log(K^2)}{\min_{r \in \cR}\sum_{j \in \cBs(r)}N_j}} \tagcmt{use $1 \le \log(K^2)$ with $K \ge 2$} \\
    \stackrel{}{\le}\,& 4\del{\max_{r \in \cR}\fr{1}{\sum_{j \in \cBs(r)}N_j}\sum_{i \in \cBs(r)}N_i\Delta_i}^2 + 48\del{\fr{\log(K)}{\min_{r \in \cR}\sum_{j \in \cBs(r)}N_j}}.
  \end{align*}

\end{proof}


\begin{lemma}\label{lemma_g1a:haver21count2}
  In the event of $G_1 = \cbr{\cBs(\hr) \subset \cB \subseteq \cBp(\hr)}$, HAVER achieves
  \begin{align*}
    &\int_{0}^{\infty} \PP\del{ \hmuB - \mu_1  < -\eps,\, \hr \in \cR,\, G_1 } \difeps \\
    \stackrel{}{\le}\,& 64\del{ \max_{r \in \cR} \max_{\substack{S: \cBs(r) \subseteq \Scal \subseteq \cBp(r) }}\fr{1}{\sum_{j \in \Scal}N_j}\sum_{i \in \Scal}N_i\Delta_i}^2 \\
  &+ 512\del{\max_{r \in \cR} \max_{k=0}^{d(r)} \max_{\substack{S: \cBs(r) \subseteq \Scal \subseteq \cBp(r) \\ \abs{\Scal}=n_*(r)+k }}\fr{k\sum_{j \in \Scal \sm \cBs(r)}N_j}{\del{\sum_{j \in \Scal}N_j}^2}\log\del{\fr{ed(r)}{k}}\log(K)} \\
  &+ 192\del{\fr{\log(K)}{\min_{r \in \cR}\sum_{j \in \cBs(r)}N_j}},
  \end{align*}
  where $d(r) = \abs{\cBp(r)} - \abs{\cBs(r)}$ and $n_*(r) = \abs{\cBs(r)}$.
\end{lemma}

\begin{proof}
  We decompose the integral as follows
  \begin{align*}
    &\int_{0}^{\infty} \PP\del{ \hmuB - \mu_1 < -\eps,\, \hr \in \cR,\, G_1 } \difeps \\
    \stackrel{}{=}\,& \int_{0}^{\infty} \PP\del{ \fr{1}{\sum_{j \in \cB}N_j}\sum_{i \in \cB}N_i\del{\muhat_i - \mu_1} < -\eps,\, \hr \in \cR,\, G_1 } \difeps \\
    \stackrel{}{\le}\,& \int_{0}^{\infty} \PP\del{ \fr{1}{\sum_{j \in \cB}N_j}\sum_{i \in \cBs(\hr)}N_i\del{\muhat_i - \mu_1} < -\fr{\eps}{2},\, \hr \in \cR,\, G_1 } \difeps \\
    &+ \int_{0}^{\infty} \PP\del{ \fr{1}{\sum_{j \in \cB}N_j}\sum_{i \in \cB \sm \cBs(\hr)}N_i\del{\muhat_i - \mu_1} < -\fr{\eps}{2},\, \hr \in \cR,\, G_1 } \difeps. 
  \end{align*}

  We bound the probability and integral of these terms respectively. For the first term, we have
  \begin{align*}
    &\PP\del{ \fr{1}{\sum_{j \in \cB}N_j}\sum_{i \in \cBs(\hr)}N_i\del{\muhat_i - \mu_1} < -\fr{\eps}{2},\, \hr \in \cR,\, G_1 } \\
    \stackrel{}{=}\,& \PP\del{ \sum_{i \in \cBs(\hr)}N_i\del{\muhat_i - \mu_1} < -\fr{\eps}{2}\sum_{j \in \cB}N_j,\, \hr \in \cR,\, G_1 } \\
    \stackrel{}{\le}\,& \PP\del{ \sum_{i \in \cBs(\hr)}N_i\del{\muhat_i - \mu_1} < -\fr{\eps}{2}\sum_{j \in \cBs(\hr)}N_j,\, \hr \in \cR,\, G_1 } \tagcmt{since $\cBs(\hr) \subset \cB$, $\sum_{j \in \cBs(\hr)}N_j < \sum_{j \in \cB}N_j$}\\
    \stackrel{}{=}\,& \PP\del{ \sum_{i \in \cBs(\hr)}N_i\del{\muhat_i - \mu_i} < -\fr{\eps}{2}\sum_{j \in \cBs(\hr)}N_j + \sum_{i \in \cBs(\hr)}N_i\Delta_i,\, \hr \in \cR,\, G_1 } \\
    \stackrel{}{\le}\,& \sum_{r \in \cR}\PP\del{ \sum_{i \in \cBs(r)}N_i\del{\muhat_i - \mu_i} < -\fr{\eps}{2}\sum_{j \in \cBs(r)}N_j + \sum_{i \in \cBs(r)}N_i\Delta_i} \\
    \stackrel{}{\le}\,& \sum_{r \in \cR} \exp\del{-\fr{1}{2}\fr{1}{\sum_{j \in \cBs(r)}N_j}\del{\fr{\eps}{2}\sum_{j \in \cBs(r)}N_j - \sum_{i \in \cBs(r)}N_i\Delta_i}_+^2} \\
    \stackrel{}{=}\,& \sum_{r \in \cR} \exp\del{-\fr{1}{8}\del{\sum_{j \in \cBs(r)}N_j}\del{\eps - \fr{2}{\sum_{j \in \cBs(r)}N_j}\sum_{i \in \cBs(r)}N_i\Delta_i}_+^2} \\
    \stackrel{}{\le}\,& K \exp\del{-\fr{1}{8}\del{\min_{r \in \cR} \sum_{j \in \cBs(r)}N_j}\del{\eps - \max_{r \in \cR}\fr{2}{\sum_{j \in \cBs(r)}N_j}\sum_{i \in \cBs(r)}N_i\Delta_i}_+^2}. 
  \end{align*}

  For this term, let $q_1 = 1$, $q_2 = 1$, $z_1 = \fr{1}{32}\del{\min_{r \in \cR}\sum_{j \in \cBs(r)}N_j}$, and
  \begin{align*}
    \eps_0 = \del{4\max_{r \in \cR}\fr{1}{\sum_{j \in \cBs(r)}N_j}\sum_{i \in \cBs(r)}N_i\Delta_i} \vee \del{\sqrt{\fr{64\log(K)}{\min_{r \in \cR}\sum_{j \in \cBs(r)}N_j}}}.
  \end{align*}
  
  We claim that, in the regime of $\eps \ge \eps_0$, we have
  \begin{align*}
    \PP\del{ \fr{1}{\sum_{j \in \cB}N_j}\sum_{i \in \cBs(r)}N_i\del{\muhat_i - \mu_1} < -\fr{\eps}{2},\, \hr \in \cR,\, G_1 } \le q_1\exp\del{-z_1\eps^2}
  \end{align*}
  and in the regime of $\eps \ge \eps_0$, we have
  \begin{align*}
    \PP\del{ \fr{1}{\sum_{j \in \cB}N_j}\sum_{i \in \cBs(\hr)}N_i\del{\muhat_i - \mu_1} < -\fr{\eps}{2},\, \hr \in \cR,\, G_1 } \le q_2.
  \end{align*}
  The second claim is trivial. We prove the first claim as follows. In the regime of $\eps \ge \eps_0$, we have 
  \begin{align*}
    &\PP\del{ \fr{1}{\sum_{j \in \cB}N_j}\sum_{i \in \cBs}N_i\del{\muhat_i - \mu_1} < -\fr{\eps}{2},\, \hr \in \cR,\, G_1 } \\
    \stackrel{}{\le}\,& K\exp\del{-\fr{1}{8}\del{\min_{r \in \cR} \sum_{j \in \cBs(r)}N_j}\del{\eps - \max_{r \in \cR} \fr{2}{\sum_{j \in \cBs(r)}N_j}\sum_{i \in \cBs}N_i\Delta_i}_+^2} \\
    \stackrel{}{\le}\,& K\exp\del{-\fr{1}{8}\del{\min_{r \in \cR} \sum_{j \in \cBs(r)}N_j}\del{\fr{\eps}{2}}^2} \tagcmt{use $\eps \ge \eps_0$} \\
    \stackrel{}{=}\,& K\exp\del{-\fr{1}{32}\del{\min_{r \in \cR} \sum_{j \in \cBs(r)}N_j}\eps^2} \\
    \stackrel{}{=}\,& \exp\del{-\fr{1}{32}\del{\min_{r \in \cR} \sum_{j \in \cBs(r)}N_j}\eps^2 + \log(K)} \\
    \stackrel{}{\le}\,& \exp\del{-\fr{1}{64}\del{\min_{r \in \cR} \sum_{j \in \cBs(r)}N_j}\eps^2 } \tagcmt{use $\eps \ge \eps_0$}.
  \end{align*}
  
  With the above claim, we use Lemma~\ref{lemma401:utility_integral_with_eps} to bound the integral
  \begin{align*}
    &\int_{0}^{\infty} \PP\del{ \fr{1}{\sum_{j \in \cB}N_j}\sum_{i \in \cBs}N_i\del{\muhat_i - \mu_1} < -\fr{\eps}{2},\, \hr \in \cR,\, G_1 } \difeps \\
    \stackrel{}{\le}\,& q_1\eps_0^2 + q_2\fr{1}{z_1} \\
    \stackrel{}{=}\,& \del{\del{4\max_{r \in \cR}\fr{1}{\sum_{j \in \cBs(r)}N_j}\sum_{i \in \cBs(r)}N_i\Delta_i} \vee \del{\sqrt{\fr{64\log(K)}{\min_{r \in \cR}\sum_{j \in \cBs(r)}N_j}}}}^2 + 64\del{\fr{1}{\min_{r \in \cR}\sum_{j \in \cBs(r)}N_j}} \\
    \stackrel{}{\le}\,& 16\del{\max_{r \in \cR}\fr{1}{\sum_{j \in \cBs(r)}N_j}\sum_{i \in \cBs(r)}N_i\Delta_i}^2 + 64\del{\fr{\log(K)}{\min_{r \in \cR}\sum_{j \in \cBs(r)}N_j}} + 64\del{\fr{1}{\min_{r \in \cR}\sum_{j \in \cBs(r)}N_j}} \\
    \stackrel{}{=}\,& 16\del{\max_{r \in \cR}\fr{1}{\sum_{j \in \cBs(r)}N_j}\sum_{i \in \cBs(r)}N_i\Delta_i}^2 + 32\del{\fr{\log(K^2)}{\min_{r \in \cR}\sum_{j \in \cBs(r)}N_j}} + 64\del{\fr{1}{\min_{r \in \cR}\sum_{j \in \cBs(r)}N_j}} \\
    \stackrel{}{\le}\,& 16\del{\max_{r \in \cR}\fr{1}{\sum_{j \in \cBs(r)}N_j}\sum_{i \in \cBs(r)}N_i\Delta_i}^2 + 96\del{\fr{\log(K^2)}{\min_{r \in \cR}\sum_{j \in \cBs(r)}N_j}} \tagcmt{use $1 \le \log(K^2)$ with $K \ge 2$} \\
    \stackrel{}{=}\,& 16\del{\max_{r \in \cR}\fr{1}{\sum_{j \in \cBs(r)}N_j}\sum_{i \in \cBs(r)}N_i\Delta_i}^2 + 192\del{\fr{\log(K)}{\min_{r \in \cR}\sum_{j \in \cBs(r)}N_j}}.
  \end{align*}

  For the second term, we have 
  \begin{align*}
    &\PP\del{\fr{1}{\sum_{j \in \cB}N_j} \sum_{i \in \cB \sm \cBs(\hr)}N_i\del{\muhat_i - \mu_1} < -\fr{\eps}{2},\, \hr \in \cR,\, G_1} \\
    \stackrel{}{=}\,& \PP\del{\fr{1}{\sum_{j \in \cB}N_j} \sum_{i \in \cB \sm \cBs(\hr)}N_i\del{\muhat_i - \mu_i} < -\fr{\eps}{2} + \fr{1}{\sum_{j \in \cB}N_j} \sum_{i \in \cB \sm \cBs(\hr)}N_i\Delta_i,\, \hr \in \cR,\, G_1} \\
    \stackrel{}{\le}\,& \sum_{r \in \cR} \sum_{k=1}^{d(r)}\sum_{\substack{S: \cBs(r) \subset \Scal \subseteq \cBp(r) \\ \abs{\Scal}=n_*(r)+k }} \PP\del{\fr{1}{\sum_{j \in \Scal}N_j} \sum_{i \in \Scal \sm \cBs(r)}N_i\del{\muhat_i - \mu_i} < -\fr{\eps}{2} + \fr{1}{\sum_{j \in \Scal}N_j} \sum_{i \in \Scal \sm \cBs(r)}N_i\Delta_i} \tagcmt{use $d(r) = \abs{\cBp(r)} - \abs{\cBs(r)}$ and $n_*(r) = \abs{\cBs(r)}$} \\
    \stackrel{}{=}\,& \sum_{r \in \cR} \sum_{k=1}^{d(r)}\sum_{\substack{S: \cBs(r) \subset \Scal \subseteq \cBp(r) \\ \abs{\Scal}=n_*(r)+k }} \exp\del{-\fr{1}{2}\del{\fr{\del{\sum_{j \in \Scal}N_j}^2}{\sum_{j \in \Scal \sm \cBs(r)}N_j}}\del{\fr{\eps}{2} - \fr{1}{\sum_{j \in \Scal}N_j}\sum_{i \in \Scal \sm \cBs(r)}N_i\Delta_i}_{+}^2} \\
    \stackrel{}{=}\,& \sum_{r \in \cR} \sum_{k=1}^{d(r)}\sum_{\substack{S: \cBs(r) \subset \Scal \subseteq \cBp(r) \\ \abs{\Scal}=n_*(r)+k }} \exp\del{-\fr{1}{8}\del{\fr{\del{\sum_{j \in \Scal}N_j}^2}{\sum_{j \in \Scal \sm \cBs(r)}N_j}}\del{\eps - \fr{2}{\sum_{j \in \Scal}N_j}\sum_{i \in \Scal \sm \cBs(r)}N_i\Delta_i}_{+}^2}.
  \end{align*}
  
  For brevity, we denote
  \begin{align*}
    M = \del{\min_{r \in \cR}\min_{k=1}^{d(r)}\min_{\substack{S: \cBs(r) \subset \Scal \subseteq \cBp(r) \\ \abs{\Scal}=n_*(r)+k }}\fr{\del{\sum_{j \in \Scal}N_j}^2}{\sum_{j \in \Scal \sm \cBs(r)}N_j}}
  \end{align*}
  and
  \begin{align*}
    H = \max_{r \in \cR}\max_{\substack{S: \cBs(r) \subset \Scal \subseteq \cBp(r) }} \fr{1}{\sum_{j \in \Scal}N_j}\sum_{i \in \Scal \sm \cBs(r)}N_i\Delta_i.
  \end{align*}
  For $\forall r \in \cR$,$\forall k \in \cbr{1,\ldots,d(r)}$, and $\forall S,\, \cBs(r) \subset \Scal \subseteq \cBp(r),\, \abs{\Scal}=n_*(r)+k$, we have
  \begin{align*}
    \del{\fr{\del{\sum_{j \in \Scal}N_j}^2}{\sum_{j \in \Scal \sm \cBs(r)}N_j}} \ge M
  \end{align*} and
  \begin{align*}
    \fr{1}{\sum_{j \in \Scal}N_j}\sum_{i \in \Scal \sm \cBs(r)}N_i\Delta_i \le H.
  \end{align*}
  Thus, we have
  \begin{align*}
    &\PP\del{\fr{1}{\sum_{j \in \cB}N_j} \sum_{i \in \cB \sm \cBs(\hr)}N_i\del{\muhat_i - \mu_1} < -\fr{\eps}{2},\, \hr \in \cR,\, G_1} \\
    \stackrel{}{\le}\,& \sum_{r \in \cR} \sum_{k=1}^{d(r)}\sum_{\substack{S: \cBs(r) \subset \Scal \subseteq \cBp(r) \\ \abs{\Scal}=n_*(r)+k }} \exp\del{-\fr{1}{8}\del{\fr{\del{\sum_{j \in \Scal}N_j}^2}{\sum_{j \in \Scal \sm \cBs(r)}N_j}}\del{\eps - \fr{2}{\sum_{j \in \Scal}N_j}\sum_{i \in \Scal \sm \cBs(r)}N_i\Delta_i}_{+}^2} \\
    \stackrel{}{\le}\,& \sum_{r \in \cR} \sum_{k=1}^{d(r)}\sum_{\substack{S: \cBs(r) \subset \Scal \subseteq \cBp(r) \\ \abs{\Scal}=n_*(r)+k }} \exp\del{-\fr{1}{8}M\del{\eps - 2H}_{+}^2}\\
    \stackrel{}{=}\,& \sum_{r \in \cR} \sum_{k=1}^{d(r)} \binom{d(r)}{k} \exp\del{-\fr{1}{8}M\del{\eps - 2H}_{+}^2}.
  \end{align*}

  For this term, let $q_1 = q_2 = 1$, $z_1 = \fr{M}{128}$, and
  \begin{align*}
    \eps_0 = \del{4H} \vee \del{\sqrt{\displaystyle \max_{r  \in \cR}  \max_{k=1}^{d(r)} \max_{\substack{S: \cBs(r) \subset \Scal \subseteq \cBp(r) \\ \abs{\Scal}=n_*(r)+k }}\fr{128k\sum_{j \in \Scal \sm \cBs(r)}N_j}{\del{\sum_{j \in \Scal}N_j}^2}\log\del{\fr{ed(r)}{k}}\log(K^2)}}.
  \end{align*}
  We claim that, in the regime of $\eps \ge \eps_0$, we have
  \begin{align*}
    \PP\del{\fr{1}{\sum_{j \in \cB}N_j} \sum_{i \in \cB \sm \cBs(r)}N_i\del{\muhat_i - \mu_1} < -\fr{\eps}{2},\, \hr \in \cR,\, G_1} \le q_1\exp\del{-z_1\eps^2}
  \end{align*}
  and in the regime of $\eps < \eps_0$, we have
  \begin{align*}
    \PP\del{\fr{1}{\sum_{j \in \cB}N_j} \sum_{i \in \cB \sm \cBs(r)}N_i\del{\muhat_i - \mu_1} < -\fr{\eps}{2},\, \hr \in \cR,\, G_1} \le q_2.
  \end{align*}
  The second claim is trivial. We prove the first claim as follows. In the regime of $\eps \ge \eps_0$, we have 
  \begin{align*}
    &\PP\del{\fr{1}{\sum_{j \in \cB}N_j} \sum_{i \in \cB \sm \cBs(r)}N_i\del{\muhat_i - \mu_1} < -\fr{\eps}{2},\, \hr \in \cR,\, G_1} \\
    \stackrel{}{\le}\,& \sum_{r \in \cR} \sum_{k=1}^{d(r)}\binom{d(r)}{k} \exp\del{- \fr{1}{8}M\del{\eps - 2H}_{+}^2} \\
    \stackrel{}{\le}\,& \sum_{r \in \cR} \sum_{k=1}^{d(r)}\binom{d(r)}{k} \exp\del{-\fr{1}{8}M\del{\fr{\eps}{2}}^2} \tagcmt{use $\eps \ge \eps_0$} \\
  \stackrel{}{=}\,& \sum_{r \in \cR} \sum_{k=1}^{d(r)}\binom{d(r)}{k} \exp\del{-\fr{1}{32}M\eps^2} \\
  \stackrel{}{\le}\,& \sum_{r \in \cR} \sum_{k=1}^{d(r)}\del{\fr{ed(r)}{k}}^k \exp\del{-\fr{1}{32}M\eps^2} \tagcmt{use Stirling's formula, Lemma~\ref{xlemma:stirling_formula}} \\
  \stackrel{}{=}\,& \sum_{r \in \cR} \sum_{k=1}^{d(r)} \exp\del{-\fr{1}{32}M\eps^2 + k\log\del{\fr{ed(r)}{k}}} \\
  \stackrel{}{\le}\,& \sum_{r \in \cR} \sum_{k=1}^{d(r)} \exp\del{-\fr{1}{64}M\eps^2} \tagcmt{use $\eps \ge \eps_0$} \\
  \stackrel{}{\le}\,& K d \exp\del{-\fr{1}{64}M\eps^2} \\
  \stackrel{}{=}\,& \exp\del{-\fr{1}{64}M\eps^2 + \ln(Kd)} \\
  \stackrel{}{\le}\,& \exp\del{-\fr{1}{128}M\eps^2 } \tagcmt{use $\eps \ge \eps_0 $}.
  \end{align*}
  With the above claim, we use Lemma~\ref{lemma401:utility_integral_with_eps} to bound the integral
  \begin{align*}
    &\int_{0}^{\infty} \PP\del{\fr{1}{\sum_{j \in \cB}N_j} \sum_{i \in \cB \sm \cBs(r)}N_i\del{\muhat_i - \mu_1} < -\fr{\eps}{2},\, \hr \in \cR,\, G_1} \difeps \\
    \stackrel{}{\le}\,& q_1\eps_0^2 + q_2\fr{1}{z_1} \\
    \stackrel{}{\le}\,& \del{\del{4H} \vee \del{\sqrt{\displaystyle \max_{r \in \cR} \max_{k=1}^{d(r)} \max_{\substack{S: \cBs(r) \subset \Scal \subseteq \cBp(r) \\ \abs{\Scal}=n_*(r)+k }}\fr{128k\sum_{j \in \Scal \sm \cBs(r)}N_j}{\del{\sum_{j \in \Scal}N_j}^2}\log\del{\fr{ed(r)}{k}}\log(K^2)}}}^2 + \fr{128}{M} \\
    \stackrel{}{\le}\,& 16H^2 + 128\del{\max_{r \in \cR} \max_{k=1}^{d(r)} \max_{\substack{S: \cBs(r) \subset \Scal \subseteq \cBp(r) \\ \abs{\Scal}=n_*(r)+k }}\fr{k\sum_{j \in \Scal \sm \cBs(r)}N_j}{\del{\sum_{j \in \Scal}N_j}^2}\log\del{\fr{ed(r)}{k}}\log(K^2)} \tagcmt{use  $\del{A \vee B}^2 \le A^2 + B^2$} + \fr{128}{M} \\
    \stackrel{}{=}\,& 16H^2 + 128\del{\max_{r \in \cR} \max_{k=1}^{d(r)} \max_{\substack{S: \cBs(r) \subset \Scal \subseteq \cBp(r) \\ \abs{\Scal}=n_*(r)+k }}\fr{k\sum_{j \in \Scal \sm \cBs(r)}N_j}{\del{\sum_{j \in \Scal}N_j}^2}\log\del{\fr{ed(r)}{k}}\log(K^2)} \\
    &+ 128\del{\max_{r \in \cR} \max_{k=1}^{d(r)} \max_{\substack{S: \cBs(r) \subset \Scal \subseteq \cBp(r) \\ \abs{\Scal}=n_*(r)+k }}\fr{128\sum_{j \in \Scal \sm \cBs(r)}N_j}{\del{\sum_{j \in \Scal}N_j}^2}} \tagcmt{plug in $M$} \\
    \stackrel{}{=}\,& 16H^2 + 256\del{\max_{r \in \cR} \max_{k=1}^{d(r)} \max_{\substack{S: \cBs(r) \subset \Scal \subseteq \cBp(r) \\ \abs{\Scal}=n_*(r)+k }}\fr{k\sum_{j \in \Scal \sm \cBs(r)}N_j}{\del{\sum_{j \in \Scal}N_j}^2}\log\del{\fr{ed(r)}{k}}\log(K^2)} \tagcmt{use $1 \le \log\del{\fr{ed(r)}{k}}\,\forall k \in \cbr{1,\ldots,d}$ and $1 \le \log(K^2)\, \forall K \ge 2$} \\
    \stackrel{}{=}\,& 16H^2 + 512\del{\max_{r \in \cR} \max_{k=1}^{d(r)} \max_{\substack{S: \cBs(r) \subset \Scal \subseteq \cBp(r) \\ \abs{\Scal}=n_*(r)+k }}\fr{k\sum_{j \in \Scal \sm \cBs(r)}N_j}{\del{\sum_{j \in \Scal}N_j}^2}\log\del{\fr{ed(r)}{k}}\log(K)} \\
    \stackrel{}{=}\,& 16\del{\displaystyle \max_{r \in \cR} \max_{\substack{S: \cBs(r) \subset \Scal \subseteq \cBp(r)}}\fr{1}{\sum_{j \in \Scal}N_j}\sum_{i \in \Scal}N_i\Delta_i}^2 \tagcmt{plug in $H$} \\
    &+ 512\del{\max_{r \in \cR} \max_{k=1}^{d(r)} \max_{\substack{S: \cBs(r) \subset \Scal \subseteq \cBp(r) \\ \abs{\Scal}=n_*(r)+k }}\fr{k\sum_{j \in \Scal \sm \cBs(r)}N_j}{\del{\sum_{j \in \Scal}N_j}^2}\log\del{\fr{ed(r)}{k}}\log(K)}.
  \end{align*}

  Combining the two terms, we have 
  \begin{align*}
    &\int_{0}^{\infty} \PP\del{ \hmuB - \mu_1 < -\eps,\, \hr \in \cR,\, G_1 } \difeps \\
    \stackrel{}{\le}\,& \int_{0}^{\infty} \PP\del{ \fr{1}{\sum_{j \in \cB}N_j}\sum_{i \in \cBs(r)}N_i\del{\muhat_i - \mu_1} < -\fr{\eps}{2},\, \hr \in \cR,\, G_1 } \difeps \\
    &+ \int_{0}^{\infty} \PP\del{ \fr{1}{\sum_{j \in \cB}N_j}\sum_{i \in \cB \sm \cBs(r)}N_i\del{\muhat_i - \mu_1} < -\fr{\eps}{2},\, \hr \in \cR,\, G_1 } \difeps \\
    \stackrel{}{\le}\,& 16\del{\max_{r \in \cR}\fr{1}{\sum_{j \in \cBs(r)}N_j}\sum_{i \in \cBs(r)}N_i\Delta_i}^2 + 192\del{\fr{\log(K)}{\min_{r \in \cR}\sum_{j \in \cBs(r)}N_j}} \\
    &+ 16\del{\displaystyle \max_{r \in \cR} \max_{\substack{S: \cBs(r) \subset \Scal \subseteq \cBp(r)}}\fr{1}{\sum_{j \in \Scal}N_j}\sum_{i \in \Scal}N_i\Delta_i}^2 \\
    &+ 512\del{\max_{r \in \cR} \max_{k=1}^{d(r)} \max_{\substack{S: \cBs(r) \subset \Scal \subseteq \cBp(r) \\ \abs{\Scal}=n_*(r)+k }}\fr{k\sum_{j \in \Scal \sm \cBs(r)}N_j}{\del{\sum_{j \in \Scal}N_j}^2}\log\del{\fr{ed(r)}{k}}\log(K)} \\
    \stackrel{}{\le}\,& 16\del{\displaystyle \max_{r \in \cR} \max_{\substack{S: \cBs(r) \subset \Scal \subseteq \cBp(r) }}\fr{1}{\sum_{j \in \Scal}N_j}\sum_{i \in \Scal}N_i\Delta_i + \max_{r \in \cR} \fr{1}{\sum_{j \in \cBs(r)}N_j}\sum_{i \in \cBs(r)}N_i\Delta_i}^2 \tagcmt{use $\forall a,b \ge 0,\, a^2 + b^2 \le (a+b)^2$} \\
    &+ 512\del{\max_{r \in \cR} \max_{k=1}^{d(r)} \max_{\substack{S: \cBs(r) \subset \Scal \subseteq \cBp(r) \\ \abs{\Scal}=n_*(r)+k }}\fr{k\sum_{j \in \Scal \sm \cBs(r)}N_j}{\del{\sum_{j \in \Scal}N_j}^2}\log\del{\fr{ed(r)}{k}}\log(K)} \\
    &+ 192\del{\fr{\log(K)}{\min_{r \in \cR}\sum_{j \in \cBs(r)(r)}N_j}} \\
    \stackrel{}{\le}\,& 64\del{\displaystyle \max_{r \in \cR} \max_{\substack{S: \cBs(r) \subseteq \Scal \subseteq \cBp(r) }}\fr{1}{\sum_{j \in \Scal}N_j}\sum_{i \in \Scal}N_i\Delta_i}^2 \\
    &+ 512\del{\max_{r \in \cR} \max_{k=0}^{d(r)} \max_{\substack{S: \cBs(r) \subseteq \Scal \subseteq \cBp(r) \\ \abs{\Scal}=n_*(r)+k }}\fr{k\sum_{j \in \Scal \sm \cBs(r)}N_j}{\del{\sum_{j \in \Scal}N_j}^2}\log\del{\fr{ed(r)}{k}}\log(K)} \\
    &+ 192\del{\fr{\log(K)}{\min_{r \in \cR}\sum_{j \in \cBs(r)}N_j}}.
  \end{align*}

\end{proof}

\begin{lemma}\label{lemma101.a:haver}
  Let $m \in [K]$ be any arm. In HAVER, we have 
  \begin{align*}
    \PP\del{ \hmuB - \mu_1 < -\eps } \le 2\exp\del{-\fr{1}{2}N_{m}\del{\eps - \gam_m - \Delta_m}_+^2}.
  \end{align*}
\end{lemma}

\begin{proof}
  We separate the probability into two terms 
  \begin{align*}
    &\PP\del{ \hmuB - \mu_1 < -\eps } \\
    \stackrel{}{=}\,& \PP\del{\hr =  m,\, \hmuB - \mu_1 < -\eps } + \PP\del{\hr \in [K] \sm \cbr{m},\, \hmuB - \mu_1 < -\eps }. 
  \end{align*}

  For the first term, we have
  \begin{align*}
    & \PP\del{\hr =  m,\, \hmuB - \mu_1 < -\eps } \\
    \stackrel{}{=}\,& \PP\del{\hr =  m,\, \fr{N_m}{\sum_{j \in \cB}N_j}\muhat_m + \sum_{i \in \cB \setminus \cbr{m}}\fr{N_i}{\sum_{j \in \cB}N_j}\muhat_i - \mu_1 < -\eps } \\
    \stackrel{}{\le}\,& \PP\del{\hr =  m,\, \fr{N_m}{\sum_{j \in \cB}N_j}\muhat_m + \sum_{i \in \cB \setminus \cbr{m}}\fr{N_i}{\sum_{j \in \cB}N_j}\del{\muhat_m - \gam_m} - \mu_1 < -\eps } \tagcmt{use $\forall i \in \cB,\, \muhat_m - \gam_m \le \muhat_i$} \\
    \stackrel{}{=}\,& \PP\del{\muhat_m - \mu_1 < -\eps + \sum_{i \in \cB \setminus \cbr{m}}\fr{N_i}{\sum_{j \in \cB}N_j}\gam_m } \\
    \stackrel{}{\le}\,& \PP\del{\muhat_m - \mu_1 < -\eps + \gam_m} \tagcmt{use $\sum_{i \in \cB \setminus \cbr{m}}\fr{N_i}{\sum_{j \in \cB}N_j} \le 1$} \\
    \stackrel{}{=}\,& \PP\del{\muhat_m - \mu_m < -\eps + \gam_m + \mu_1 - \mu_m} \\
    \stackrel{}{=}\,& \PP\del{\muhat_m - \mu_m < -\eps + \gam_m + \Delta_m} \\
    \stackrel{}{\le}\,& \exp\del{-\fr{N_m\del{\eps - \gam_m - \Delta_m}_+^2}{2}}. 
  \end{align*}

  For the second term, we have
  \begin{align*}
    & \PP\del{\hr \in [K] \sm \cbr{m},\, \hmuB - \mu_1 < -\eps } \\
    \stackrel{}{=}\,& \PP\del{\hr \in [K] \sm \cbr{m},\, \fr{N_{\hr}}{\sum_{j \in \cB}N_j}\muhat_{\hr} + \sum_{i \in \cB \setminus \cbr{\hr}}\fr{N_i}{\sum_{j \in \cB}N_j}\muhat_i - \mu_1 < -\eps } \\
    \stackrel{}{\le}\,& \PP\del{\hr \in [K] \sm \cbr{m},\, \fr{N_{\hr}}{\sum_{j \in \cB}N_j}\muhat_{\hr} + \sum_{i \in \cB \setminus \cbr{\hr}}\fr{N_i}{\sum_{j \in \cB}N_j}\del{\muhat_{\hr} - \gam_{\hr}} - \mu_1 < -\eps } \tagcmt{use $\forall i \in \cB,\, \muhat_{i} \ge \muhat_{\hr} - \gam_{\hr}$} \\
    \stackrel{}{=}\,& \PP\del{\hr \in [K] \sm \cbr{m},\, \muhat_{\hr} - \mu_1 < -\eps + \sum_{i \in \cB \setminus \cbr{\hr}}\fr{N_i}{\sum_{j \in \cB}N_j}\gam_{\hr}} \\
    \stackrel{}{\le}\,& \PP\del{\hr \in [K] \sm \cbr{m},\, \muhat_{m} - \gam_{m} + \gam_{\hr} - \mu_1 < -\eps + \sum_{i \in \cB \setminus \cbr{\hr}}\fr{N_i}{\sum_{j \in \cB}N_j}\gam_{\hr}} \tagcmt{by def of $\hr$, $\muhat_{\hr} > \muhat_{m} - \gam_{m} + \gam_{\hr}$}\\
    \stackrel{}{=}\,& \PP\del{\hr \in [K] \sm \cbr{m},\, \muhat_{m}  - \mu_1 < -\eps + \gam_{m} - \gam_{\hr} + \sum_{i \in \cB \setminus \cbr{\hr}}\fr{N_i}{\sum_{j \in \cB}N_j}\gam_{\hr}} \\
    \stackrel{}{\le}\,& \PP\del{\hr \in [K] \sm \cbr{m},\, \muhat_{m}  - \mu_1 < -\eps + \gam_{m} - \gam_{\hr} + \gam_{\hr}} \tagcmt{use $\sum_{i \in \cB \setminus \cbr{m}}\fr{N_i}{\sum_{j \in \cB}N_j} \le 1$} \\
    \stackrel{}{=}\,& \PP\del{\hr \in [K] \sm \cbr{m},\, \muhat_{m}  - \mu_1 < -\eps + \gam_{m}} \\
    \stackrel{}{\le}\,& \PP\del{\muhat_m - \mu_1 < -\eps + \gam_m} \\
    \stackrel{}{=}\,& \PP\del{\muhat_m - \mu_m < -\eps + \gam_m + \mu_1 - \mu_m} \\
    \stackrel{}{=}\,& \PP\del{\muhat_m - \mu_m < -\eps + \gam_m + \Delta_m} \\
    \stackrel{}{\le}\,& \exp\del{-\fr{N_m\del{\eps - \gam_m - \Delta_m}_+^2}{2}}.
  \end{align*}

  Combining the two terms, we have
  \begin{align*}
    &\PP\del{ \hmuB - \mu_1 < -\eps } \\
    \stackrel{}{=}\,& \PP\del{\hr =  m,\, \hmuB - \mu_1 < -\eps } + \PP\del{\hr \in [K] \sm \cbr{m},\, \hmuB - \mu_1 < -\eps } \\
    \stackrel{}{\le}\,& 2\exp\del{-\fr{N_m\del{\eps - \gam_m - \Delta_m}_+^2}{2}}.
  \end{align*}
  
\end{proof}

\begin{lemma}\label{lemma102.a:haver}
  In HAVER, we have
  \begin{align*}
    \int_{0}^{\infty} \PP\del{\hmuB - \mu_1 < -\eps} \difeps \le \min_m \del{\fr{352}{N_m}\log\del{\fr{KS}{N_m}}  + 4\Delta_m^2}.
  \end{align*}
\end{lemma}

\begin{proof}
  Lemma~\ref{lemma101.a:haver} states that, for any arm $m \in [K]$,
  \begin{align*}
    \PP\del{ \hmuB - \mu_1 < -\eps } \le 2\exp\del{-\fr{1}{2}N_{m}\del{\eps - \gam_m - \Delta_m}_+^2} 
  \end{align*}
  Let $q_1 = q_2 = 1$, $\eps_0 = 2\del{\gam_m + \Delta_m}$, and $z_1 = \fr{1}{16}N_m$. We claim that, in the regime of $\eps \ge \eps_0$, we have
  \begin{align*}
    \PP\del{ \hmuB - \mu_1 < -\eps } \le q_1\exp\del{-z_1\eps^2}
  \end{align*}
  and in the regime of $\eps < \eps_0$, we have $\PP\del{ \hmuB - \mu_1 < -\eps } \le q_2$.

  The second claim is trivial. We prove the first claim as follows. In the regime of $\eps \ge \eps_0$, we have
  \begin{align*}
    & \PP\del{ \hmuB - \mu_1 < -\eps }  \\
    \stackrel{}{\le}\,& 2\exp\del{-\fr{1}{2}N_{m}\del{\eps-\gam_{m}-\Delta_{m}}_+^2} \\
    \stackrel{}{\le}\,& 2\exp\del{-\fr{1}{2}N_{m}\del{\fr{\eps}{2}}^2} \tagcmt{use $\eps \ge \eps_0 = 2\del{\gam_m + \Delta_m}$} \\
    \stackrel{}{=}\,& 2\exp\del{-\fr{1}{8}N_{m}\eps^2} \\
    \stackrel{}{=}\,& \exp\del{-\fr{1}{8}N_{m}\eps^2 + \log(2)} \\
    \stackrel{}{\le}\,& \exp\del{-\fr{1}{16}N_{m}\eps^2} \tagcmt{use $\eps \ge \eps_0 \ge \gam_m$}.
  \end{align*}
  
  With the above claim, we use Lemma~\ref{lemma401:utility_integral_with_eps} to bound the integral
  \begin{align*}
    &\int_{0}^{\infty} \PP\del{ \hmuB - \mu_1 < -\eps } \difeps \\
    \stackrel{}{\le}\,& q_2\eps_0^2 + q_1\fr{1}{z_1} \\
    \stackrel{}{=}\,& \del{2\del{\gam_m + \Delta_m}}^2 + \fr{16}{N_m} \\
    \stackrel{}{\le}\,& 4\gam_m^2  + 4\Delta_m^2 + \fr{16}{N_m} \\
    \stackrel{}{\le}\,& 4\fr{18}{N_m}\log\del{\del{\fr{KS}{N_m}}^{2\lam}}  + 4\Delta_m^2 + \fr{16}{N_m} \tagcmt{use $\gam_m = \sqrt{\fr{18}{N_m}\log\del{\del{\fr{KS}{N_m}}^{2\lam}}}$} \\
    \stackrel{}{=}\,& \fr{72}{N_m}\log\del{\del{\fr{KS}{N_m}}^{4}}  + 4\Delta_m^2 + \fr{16}{N_m} \tagcmt{recall $\lam = 2$} \\
    \stackrel{}{\le}\,& \fr{88}{N_m}\log\del{\del{\fr{KS}{N_m}}^4}  + 4\Delta_m^2 \tagcmt{use $1 \le \log\del{\del{\fr{KS}{N_m}}^4}$ with $K \ge 2$} \\
    \stackrel{}{\le}\,& \fr{352}{N_m}\log\del{\fr{KS}{N_m}}  + 4\Delta_m^2 \\
    \stackrel{}{\le}\,& \min_m \del{\fr{352}{N_m}\log\del{\fr{KS}{N_m}}  + 4\Delta_m^2}.
  \end{align*}
  
\end{proof}

\begin{lemma}\label{lemma_g2a:haver21count2}
  If an event $G$ sastisfies $\PP\del{G} \le \fr{2K^2}{K^{2\lam}}$, then HAVER achieves
  \begin{align*}
    \int_{0}^{\infty} \PP\del{ \hmuB - \mu_1 < -\eps,\, G } \dif \eps \le \min_{m \in [K]} \del{\fr{704}{K^2N_m}\log\del{\fr{KS}{N_m}}  + \fr{8}{K^{2}}\Delta_m^2}.
  \end{align*}
\end{lemma}

\begin{proof}
  Using Lemma~\ref{lemma101.a:haver}, for any arm $m$, we have 
  \begin{align*}
    &\PP\del{ \hmuB - \mu_1 < -\eps,\, G} \\
    \stackrel{}{\le}\,& \PP\del{ \hmuB - \mu_1 < -\eps} \\
    \stackrel{}{\le}\,& 2\exp\del{-\fr{1}{2}{N_{m}\del{\eps - \gam_{m} - \Delta_{m}}_+^2}}.
  \end{align*}
  We also have
  \begin{align*}
    &\PP\del{ \hmuB - \mu_1 < -\eps,\, G} \le \PP\del{G} \le \fr{2K^2}{K^{2\lam}}.
  \end{align*}
  Thus,
  \begin{align*}
    &\PP\del{ \hmuB - \mu_1 < -\eps,\, G} \le 2\exp\del{-\fr{1}{2}{N_{m}\del{\eps - \gam_{m} - \Delta_{m}}_+^2}} \wedge \fr{2K^2}{K^{2\lam}}.
  \end{align*}

  Let $q_1 = \fr{2K}{K^{\lam}}$, $q_2 = \fr{2K^2}{K^{2\lam}}$, $\eps_0 = 2\del{\gam_m + \Delta_m}$ and $z_1 = \fr{1}{16}N_m$. We claim that, in the regime of $\eps \ge \eps_0$, we have
  \begin{align*}
    \PP\del{ \hmuB - \mu_1 < -\eps,\, G} \le q_1\exp\del{-z_1\eps^2}
  \end{align*}
  and in the regime of $\eps < \eps_0$, we have $\PP\del{ \hmuB - \mu_1 < -\eps,\, G} \le q_2$.

  We prove the claim as follows. In the regime of $\eps \ge \eps_0$, we have
  \begin{align*}
    &\PP\del{ \hmuB - \mu_1 < -\eps,\, G}  \\
    \stackrel{}{\le}\,& 2\exp\del{-\fr{1}{2}N_{m}\del{\eps-\gam_{m}-\Delta_{m}}_+^2} \wedge \fr{2K^2}{K^{2\lam}} \\
    \stackrel{}{\le}\,& \sqrt{2\exp\del{-\fr{1}{2}N_{m}\del{\eps-\gam_{m}-\Delta_{m}}_+^2} \cdot \fr{2K^2}{K^{2\lam}}} \\
    \stackrel{}{=}\,& \fr{2K}{K^{\lam}}\exp\del{-\fr{1}{4}N_{m}\del{\eps-\gam_{m}-\Delta_{m}}_+^2} \\
    \stackrel{}{\le}\,& \fr{2K}{K^{\lam}}\exp\del{-\fr{1}{4}N_{m}\del{\fr{\eps}{2}}^2} \tagcmt{use $\eps \ge \eps_0 = 2\del{\gam_m + \Delta_m}$} \\
    \stackrel{}{=}\,& \fr{2K}{K^{\lam}}\exp\del{-\fr{1}{16}N_{m}\eps^2}. 
  \end{align*}
  In the regime of $\eps < \eps_0$, we have
  \begin{align*}
    &\PP\del{ \hmuB - \mu_1 < -\eps,\, G} \\
    \stackrel{}{\le}\,& 2\exp\del{-\fr{1}{2}N_{m}\del{\eps - \gam_{m} + \Delta_{m}}_+^2} \wedge \fr{2K^2}{K^{2\lam}} \\
    \stackrel{}{\le}\,& \fr{2K^2}{K^{2\lam}}.
  \end{align*}
  With the above claim, we use Lemma~\ref{lemma401:utility_integral_with_eps} to bound the integral (todo fix the constant)
  \begin{align*}
    &\int_{0}^{\infty} \PP\del{ \hmuB - \mu_1 < -\eps,\, G} \difeps \\
    \stackrel{}{\le}\,& q_2\eps_0^2 + q_1\fr{1}{z_1} \\
    \stackrel{}{=}\,& \fr{2K^2}{K^{2\lam}}\del{2\del{\gam_m + \Delta_m}}^2 + \fr{2K}{K^{\lam}}\fr{16}{N_m} \\
    \stackrel{}{\le}\,& \fr{8K^2}{K^{2\lam}}\gam_m^2  + \fr{8K}{K^{2\lam}}\Delta_m^2 + \fr{32}{K^{\lam}N_m} \\
    \stackrel{}{\le}\,& \fr{8K^2}{K^{2\lam}}\fr{18}{N_m}\log\del{\del{\fr{KS}{N_m}}^{2\lam}}  + \fr{8K}{K^{2\lam}}\Delta_m^2 + \fr{32}{K^{\lam}N_m} \tagcmt{use $\gam_m = \sqrt{\fr{18}{N_m}\log\del{\del{\fr{KS}{N_m}}^{2\lam}}}$} \\
    \stackrel{}{=}\,& \fr{144K^2}{K^{4}N_m}\log\del{\del{\fr{KS}{N_m}}^{4}}  + \fr{8K}{K^{4}}\Delta_m^2 + \fr{32}{K^{2}N_m} \tagcmt{recall $\lam = 2$} \\
    \stackrel{}{\le}\,& \fr{176}{K^2N_m}\log\del{\del{\fr{KS}{N_m}}^4}  + \fr{8}{K^{3}}\Delta_m^2 \tagcmt{use $1 \le \log\del{\del{\fr{KS}{N_m}}^4}$ with $K \ge 2$} \\
    \stackrel{}{\le}\,& \fr{704}{K^2N_m}\log\del{\fr{KS}{N_m}}  + \fr{8}{K^{3}}\Delta_m^2.
  \end{align*}
  We conclude the proof by using the fact that the result above applies to any choice of $m \in [K]$.
  
\end{proof}


\begin{lemma}\label{lemma_gb:haver21count2}
  HAVER achieves
  \begin{align*}
    &\int_{0}^{\infty} \PP\del{ \hmuB - \mu_1 > \eps,\, \hr \in \cR} \difeps \\
    \stackrel{}{=}\,& \tilcO \del{\del{\max_{r \in \cR} \max_{k=0}^{d(r)} \max_{\substack{S: \cBs(r) \subseteq \Scal \subseteq \cBp(r) \\ \abs{\Scal}=n_*(r)+k }}\fr{k\sum_{j \in \Scal \sm \cBs(r)}N_j}{\del{\sum_{j \in \Scal}N_j}^2}} \wedge {\fr{1}{N_1}}} \\
    &+ \tilcO \del{\del{\fr{1}{\min_{r \in \cR}\sum_{j \in \cBs(r)}N_j}} \wedge {\fr{1}{N_1}}}  \\
    &+ \tilcO \del{\fr{1}{KN_1}},
  \end{align*}
  where $d(r) = \abs{\cBp(r)} - \abs{\cBs(r)}$ and $n_*(r) = \abs{\cBs(r)}$.
\end{lemma}

\begin{proof}
  
  We consider 4 complementary events:
  \begin{itemize}
  \item $G_0 = \cbr{\cB = \cBs(\hr)}$
  \item $G_1 = \cbr{\cBs(\hr) \subset \cB \subseteq \cBp(\hr)}$
  \item $G_2 = \cbr{\exists i \in \cBs(\hr)\, \textrm{s.t.}\,  i \not\in \cB}$
  \item $G_3 = \cbr{\cBs(\hr) \subseteq \cB,\, \exists i \not\in \cBp(\hr)\, \textrm{s.t.}\, i \in \cB}$
  \end{itemize}

  We have
  \begin{align*}
    &\int_{0}^{\infty} \PP\del{ \hmuB - \mu_1 > \eps,\, \hr \in \cR} \difeps \\
    \stackrel{}{=}\,& \int_{0}^{\infty} \PP\del{ \hmuB - \mu_1 > \eps,\, \hr \in \cR,\, G_0} \difeps + \int_{0}^{\infty} \PP\del{ \hmuB - \mu_1 > \eps,\, \hr \in \cR,\, G_1} \difeps \\
    &+ \int_{0}^{\infty} \PP\del{ \hmuB - \mu_1 > \eps,\, \hr \in \cR,\, G_2} \difeps + \int_{0}^{\infty} \PP\del{ \hmuB - \mu_1 > \eps,\, \hr \in \cR,\, G_3} \difeps.
  \end{align*}
  
  Use Lemma~\ref{lemma_g0b:haver21count2} and ~\ref{lemma_g1b:haver21count2} for event $G_0$ and $G_1$ respectively, we have 
  \begin{align*}
    &\int_{0}^{\infty} \PP\del{ \hmuB - \mu_1 > \eps,\, \hr \in \cR,\, G_0} \difeps \\
    \stackrel{}{=}\,& \cO\del{\fr{\log(K)}{\min_{r \in \cR}\sum_{j \in \cBs(r)}N_j}}
  \end{align*}
  and 
  \begin{align*}
    &\int_{0}^{\infty} \PP\del{ \hmuB - \mu_1 > \eps,\, \hr \in \cR,\, G_1} \difeps \\
    \stackrel{}{=}\,& \cO \del{\max_{r \in \cR} \max_{k=0}^{d(r)} \max_{\substack{S: \cBs(r) \subseteq \Scal \subseteq \cBp(r) \\ \abs{\Scal}=n_*(r)+k }}\fr{k\sum_{j \in \Scal \sm \cBs(r)}N_j}{\del{\sum_{j \in \Scal}N_j}^2}\log\del{\fr{ed(r)}{k}}\log(K)} \\
    &+ \cO \del{\fr{\log(K)}{\min_{r \in \cR}\sum_{j \in \cBs(r)}N_j}}.
  \end{align*}
  Combining the first two events, we have
  \begin{align*}
    &\int_{0}^{\infty} \PP\del{ \hmuB - \mu_1 > \eps,\, \hr \in \cR,\, G_0} \difeps \\
    &+ \int_{0}^{\infty} \PP\del{ \hmuB - \mu_1 > \eps,\, \hr \in \cR,\, G_1} \difeps \\
    \stackrel{}{=}\,& \cO \del{\max_{r \in \cR} \max_{k=0}^{d(r)} \max_{\substack{S: \cBs(r) \subseteq \Scal \subseteq \cBp(r) \\ \abs{\Scal}=n_*(r)+k }}\fr{k\sum_{j \in \Scal \sm \cBs(r)}N_j}{\del{\sum_{j \in \Scal}N_j}^2}\log\del{\fr{ed(r)}{k}}\log(K)} \\
    &+ \cO\del{\fr{\log(K)}{\min_{r \in \cR}\sum_{j \in \cBs(r)}N_j}} \\
    \stackrel{}{\le}\,& \cO \del{\max_{r \in \cR} \max_{k=0}^{d(r)} \max_{\substack{S: \cBs(r) \subseteq \Scal \subseteq \cBp(r) \\ \abs{\Scal}=n_*(r)+k }}\fr{k\sum_{j \in \Scal \sm \cBs(r)}N_j}{\del{\sum_{j \in \Scal}N_j}^2}\log\del{3K}\log(K)} \tagcmt{use $\fr{ed(r)}{k} \le 3K$} \\
    &+ \cO\del{\fr{\log(K)}{\min_{r \in \cR}\sum_{j \in \cBs(r)}N_j}} \\
    \stackrel{}{=}\,& \tilcO \del{\max_{r \in \cR} \max_{k=0}^{d(r)} \max_{\substack{S: \cBs(r) \subseteq \Scal \subseteq \cBp(r) \\ \abs{\Scal}=n_*(r)+k }}\fr{k\sum_{j \in \Scal \sm \cBs(r)}N_j}{\del{\sum_{j \in \Scal}N_j}^2}} \\
    &+ \tilcO \del{\fr{1}{\min_{r \in \cR}\sum_{j \in \cBs(r)}N_j}}.
  \end{align*}
  
  We use Lemma~\ref{lemma_g2:haver21count2} and ~\ref{lemma_g3:haver21count2} to show that $\PP\del{G_2} \le \fr{2K^2}{K^{2\lam}}$ and $\PP\del{G_3} \le \fr{2K^2}{K^{2\lam}}$ respectively.

  Consequently, we use Lemma~\ref{lemma_g2b:haver21count2} for event $G_2$ to obtain the integral
  \begin{align*}
    \int_{0}^{\infty} \PP\del{ \hmuB - \mu_1 > \eps,\, \hr \in \cR,\, G_2} \difeps 
    \stackrel{}{=}\, \tilcO \del{\fr{1}{KN_1}}.
  \end{align*}
  We use use Lemma~\ref{lemma_g2b:haver21count2} for event $G_3$ and get the same result. Furthermore, Lemma~\ref{lemma102.b:haver21count2} states that
  \begin{align*}
    \int_{0}^{\infty} \PP\del{ \hmuB - \mu_1 > \eps} \difeps = \tilcO \del{\fr{1}{N_1}}.
  \end{align*}
  Combining all the results completes our proof.
\end{proof}


\begin{lemma}\label{lemma_g0b:haver21count2}
  In the event of $G_0 = \cbr{\cB = \cBs(\hr)}$, HAVER achieves
  \begin{align*}
    &\int_{0}^{\infty} \PP\del{ \hmuB - \mu_1 > \eps,\, \hr \in \cR,\, G_0 } \difeps \le 10\del{\fr{\log(K)}{\min_{r \in \cR}\sum_{j \in \cBs(r)}N_j}}.
  \end{align*}
\end{lemma}

\begin{proof}
  We have
  \begin{align*}
    &\PP\del{ \hmuB - \mu_1 > \eps,\, \hr \in \cR,\, G_0 } \\
    \stackrel{}{=}\,& \PP\del{ \fr{1}{\sum_{j \in \cB}N_j}\sum_{i \in \cB}N_i\hmu_i - \mu_1 > \eps,\, \hr \in \cR,\, G_0} \\
    \stackrel{}{=}\,& \PP\del{ \fr{1}{\sum_{j \in \cB}N_j}\sum_{i \in \cB}N_i\del{\hmu_i - \mu_1} > \eps,\, \hr \in \cR,\, G_0} \\
    \stackrel{}{=}\,& \PP\del{ \fr{1}{\sum_{j \in \cB}N_j}\sum_{i \in \cB}N_i\del{\hmu_i - \mu_i} > \eps + \fr{1}{\sum_{j \in \cB}N_j}\sum_{i \in \cB}N_i\Delta_i,\, \hr \in \cR,\, G_0} \\
    \stackrel{}{=}\,& \PP\del{ \fr{1}{\sum_{j \in \cBs(\hr)}N_j}\sum_{i \in \cBs(\hr)}N_i\del{\hmu_i - \mu_i} > \eps + \fr{1}{\sum_{j \in \cBs(\hr)}N_j}\sum_{i \in \cBs(\hr)}N_i\Delta_i,\, \hr \in \cR,\, G_0} \tagcmt{use $G_0 = \cbr{\cB = \cBs(\hr)}$} \\
    \stackrel{}{\le}\,& \sum_{r \in \cR} \PP\del{ \fr{1}{\sum_{j \in \cBs(r)}N_j}\sum_{i \in \cBs(r)}N_i\del{\hmu_i - \mu_i} > \eps + \fr{1}{\sum_{j \in \cBs(r)}N_j}\sum_{i \in \cBs(r)}N_i\Delta_i}  \\
    \stackrel{}{\le}\,& \sum_{r \in \cR}\exp\del{-\fr{1}{2}\del{\sum_{j \in \cBs(\hr)}N_j}\del{\eps + \fr{1}{\sum_{j \in \cBs(r)}N_j}\sum_{i \in \cBp(r)}N_i\Delta_i}^2} \\
    \stackrel{}{\le}\,& \sum_{r \in \cR}\exp\del{-\fr{1}{2}\del{\sum_{j \in \cBs(r)}N_j}\eps^2} \\
    \stackrel{}{\le}\,& K\exp\del{-\fr{1}{2}\del{\min_{r \in \cR}\sum_{j \in \cBs(r)}N_j}\eps^2}.
  \end{align*}

  Let $q_1 = q_2 = 1$, $z_1 = \fr{1}{4}\del{\min_{r \in \cR}\sum_{j \in \cBs(r)}N_j}$, and $\eps_0 = \sqrt{\fr{4\log(K)}{\min_{r \in \cR}\sum_{j \in \cBs(r)}N_j}}$. We claim that, in the regime of $\eps \ge \eps_0$, we have
  \begin{align*}
    \PP\del{ \hmuB - \mu_1 > \eps,\, \hr \in \cR,\, G_0 } \le q_1\exp\del{-z_1\eps^2}
  \end{align*}
  and in the regime of $\eps < \eps_0$, we have
  \begin{align*}
    \PP\del{ \hmuB - \mu_1 > \eps,\, \hr \in \cR,\, G_0 } \le q_2.
  \end{align*}
  The second claim is trivial. We prove the first claim as follows. In the regime of $\eps \ge \eps_0$, we have
  \begin{align*}
    &\PP\del{ \hmuB - \mu_1 > \eps,\, \hr \in \cR,\, G_0 } \\
    \stackrel{}{\le}\,& K\exp\del{-\fr{1}{2}\del{\min_{r \in \cR}\sum_{j \in \cBs(r)}N_j}\eps^2} \\
    \stackrel{}{\le}\,& \exp\del{-\fr{1}{2}\del{\min_{r \in \cR}\sum_{j \in \cBs(r)}N_j}\eps^2 + \log(K)} \\
    \stackrel{}{\le}\,& \exp\del{-\fr{1}{4}\del{\min_{r \in \cR}\sum_{j \in \cBs(r)}N_j}\eps^2} \tagcmt{use $\eps \ge \eps_0$}.
  \end{align*}
  
  With the above claim, we use Lemma~\ref{lemma401:utility_integral_with_eps} to bound the integral
  \begin{align*}
    &\int_{0}^{\infty} \PP\del{ \hmuB - \mu_1 > \eps,\, G_0 } \difeps \\
    \stackrel{}{\le}\,& q_1\eps_0^2 + q_2\fr{1}{z_1} \\
    \stackrel{}{=}\,& 4\del{\fr{\log(K)}{\min_{r \in \cR}\sum_{j \in \cBs(r)}N_j}} + 4\del{\fr{4}{\sum_{r \in \cR}\sum_{j \in \cBs(r)}N_j}} \\
    \stackrel{}{=}\,& 2\del{\fr{\log(K^2)}{\min_{r \in \cR}\sum_{j \in \cBs(r)}N_j}} + 4\del{\fr{4}{\sum_{r \in \cR}\sum_{j \in \cBs(r)}N_j}} \\
    \stackrel{}{\le}\,& 5\del{\fr{\log(K^2)}{\min_{r \in \cR}\sum_{j \in \cBs(r)}N_j}} \tagcmt{use $K \ge 2$} \\
    \stackrel{}{\le}\,& 10\del{\fr{\log(K)}{\min_{r \in \cR}\sum_{j \in \cBs(r)}N_j}}.
  \end{align*}

\end{proof}


\begin{lemma}\label{lemma_g1b:haver21count2}
  In the event of $G_1 = \cbr{\cBs(\hr) \subset \cB \subseteq \cBp(\hr)}$, HAVER achieves
  \begin{align*}
    &\int_{0}^{\infty} \PP\del{ \hmuB - \mu_1  > \eps,\, \hr \in \cR,\, G_1 } \difeps \\
    \stackrel{}{=}\,& 128\del{\max_{r \in \cR} \max_{k=0}^{d(r)} \max_{\substack{S: \cBs(r) \subseteq \Scal \subseteq \cBp(r) \\ \abs{\Scal}=n_*(r)+k }}\fr{k\sum_{j \in \Scal \sm \cBs(r)}N_j}{\del{\sum_{j \in \Scal}N_j}^2}\log\del{\fr{ed(r)}{k}}\log(K)} + 48\del{\fr{\log(K)}{\min_{r \in \cR}\sum_{j \in \cBs(r)}N_j}},
  \end{align*}
  where $d(r) = \abs{\cBp(r)} - \abs{\cBs(r)}$ and $n_*(r) = \abs{\cBs(r)}$.
\end{lemma}

\begin{proof}
  We decompose the integral as follow:
  \begin{align*}
    &\int_{0}^{\infty} \PP\del{ \hmuB - \mu_1 > \eps,\, \hr \in \cR,\, G_1 } \difeps \\
    \stackrel{}{=}\,& \int_{0}^{\infty} \PP\del{ \fr{1}{\sum_{j \in \cB}N_j}\sum_{i \in \cB}N_i\del{\muhat_i - \mu_1} > \eps,\, \hr \in \cR,\, G_1 } \difeps \\
    \stackrel{}{\le}\,& \int_{0}^{\infty} \PP\del{ \fr{1}{\sum_{j \in \cB}N_j}\sum_{i \in \cBs(\hr)}N_i\del{\muhat_i - \mu_1} > \fr{\eps}{2},\, \hr \in \cR,\, G_1 } \difeps \\
    &+ \int_{0}^{\infty} \PP\del{ \fr{1}{\sum_{j \in \cB}N_j}\sum_{i \in \cB \sm \cBs(\hr)}N_i\del{\muhat_i - \mu_1} > \fr{\eps}{2},\, \hr \in \cR,\, G_1 } \difeps. 
  \end{align*}

  We bound the probability and integral of these terms respectively. For the first term, we have  
  \begin{align*}
    &\PP\del{ \fr{1}{\sum_{j \in \cB}N_j}\sum_{i \in \cBs(\hr)}N_i\del{\muhat_i - \mu_1} > \fr{\eps}{2},\, \hr \in \cR,\, G_1 } \\
    \stackrel{}{=}\,& \PP\del{ \sum_{i \in \cBs(\hr)}N_i\del{\muhat_i - \mu_1} > \fr{\eps}{2}\sum_{j \in \cB}N_j,\, \hr \in \cR,\, G_1 } \\
    \stackrel{}{\le}\,& \PP\del{ \sum_{i \in \cBs(\hr)}N_i\del{\muhat_i - \mu_1} > \fr{\eps}{2}\sum_{j \in \cBs(\hr)}N_j,\, \hr \in \cR,\, G_1 } \tagcmt{since $\cBs(\hr) \subset \cB$, $\sum_{j \in \cBs(\hr)}N_j < \sum_{j \in \cB}N_j$}\\
    \stackrel{}{=}\,& \PP\del{ \sum_{i \in \cBs(\hr)}N_i\del{\muhat_i - \mu_i} > \fr{\eps}{2}\sum_{j \in \cBs(\hr)}N_j + \sum_{i \in \cBs(\hr)}N_i\Delta_i,\, \hr \in \cR,\, G_1 } \\
    \stackrel{}{\le}\,& \sum_{r \in \cR} \PP\del{ \sum_{i \in \cBs(r)}N_i\del{\muhat_i - \mu_i} \ge \fr{\eps}{2}\sum_{j \in \cBs(r)}N_j + \sum_{i \in \cBs(r)}N_i\Delta_i} \\
    \stackrel{}{\le}\,& \sum_{r \in \cR} \exp\del{-\fr{1}{2}\fr{1}{\sum_{j \in \cBs(r)}N_j}\del{\fr{\eps}{2}\sum_{j \in \cBs(r)}N_j + \sum_{i \in \cBs(r)}N_i\Delta_i}^2} \\
    \stackrel{}{\le}\,& \sum_{r \in \cR} \exp\del{-\fr{1}{2}\fr{1}{\sum_{j \in \cBs(r)}N_j}\del{\fr{\eps}{2}\sum_{j \in \cBs(r)}N_j}^2} \\
    \stackrel{}{=}\,& \sum_{r \in \cR} \exp\del{-\fr{1}{8}\del{\sum_{j \in \cBs(r)}N_j}\eps^2} \\
    \stackrel{}{\le}\,& K\exp\del{-\fr{1}{8}\del{\min_{r \in \cR}\sum_{j \in \cBs(r)}N_j}\eps^2}.
  \end{align*}

  For this term, let $q_1 = q_2 = 1$, $z_1 = \fr{1}{16}\del{\min_{r \in \cR}\sum_{j \in \cBs(r)}N_j}$, and $\eps_0 = \sqrt{\fr{16\log(K)}{\min_{r \in \cR}\sum_{j \in \cBs(r)}N_j}}$. We claim that, in the regime of $\eps \ge \eps_0$, we have
  \begin{align*}
    \PP\del{ \fr{1}{\sum_{j \in \cB}N_j}\sum_{i \in \cBs(\hr)}N_i\del{\muhat_i - \mu_1} > \fr{\eps}{2},\, \hr \in \cR,\, G_1 } \le q_1\exp\del{-z_1\eps^2}
  \end{align*}
  and in the regime of $\eps < \eps_0$, we have
  \begin{align*}
    \PP\del{ \fr{1}{\sum_{j \in \cB}N_j}\sum_{i \in \cBs(\hr)}N_i\del{\muhat_i - \mu_1} > \fr{\eps}{2},\, \hr \in \cR,\, G_1 } \le q_2.
  \end{align*}
  The second claim is trivial. We prove the first claim as follows. In the regime of $\eps \ge \eps_0$, we have
  \begin{align*}
    & \PP\del{ \fr{1}{\sum_{j \in \cB}N_j}\sum_{i \in \cBs(\hr)}N_i\del{\muhat_i - \mu_1} > \fr{\eps}{2},\, \hr \in \cR,\, G_1 } \\
    \stackrel{}{\le}\,& K\exp\del{-\fr{1}{8}\del{\min_{r \in \cR}\sum_{j \in \cBs(r)}N_j}\eps^2} \\
    \stackrel{}{=}\,& \exp\del{-\fr{1}{8}\del{\min_{r \in \cR}\sum_{j \in \cBs(r)}N_j}\eps^2 + \log(K)} \\
    \stackrel{}{\le}\,& \exp\del{-\fr{1}{16}\del{\min_{r \in \cR}\sum_{j \in \cBs(r)}N_j}\eps^2 } \tagcmt{use $\eps \ge \eps_0$}
  \end{align*}
  With the above claim, we use Lemma~\ref{lemma401:utility_integral_with_eps} to bound the integral
  \begin{align*}
    &\int_{0}^{\infty} \PP\del{ \fr{1}{\sum_{j \in \cB}N_j}\sum_{i \in \cBs(\hr)}N_i\del{\muhat_i - \mu_1} > \fr{\eps}{2},\, \hr \in \cR,\, G_1 } \difeps \\
    \stackrel{}{\le}\,& q_1\eps_0^2 + q_2\fr{1}{z_1} \\
    \stackrel{}{=}\,& 16\del{\fr{\log(K)}{\min_{r \in \cR}\sum_{j \in \cBs(r)}N_j}} + 16\del{\fr{1}{\min_{r \in \cR}\sum_{j \in \cBs(r)}N_j}} \\
    \stackrel{}{=}\,& 8\del{\fr{\log(K^2)}{\min_{r \in \cR}\sum_{j \in \cBs(r)}N_j}} + 16\del{\fr{1}{\min_{r \in \cR}\sum_{j \in \cBs(r)}N_j}} \\
    \stackrel{}{\le}\,& 24\del{\fr{\log(K^2)}{\min_{r \in \cR}\sum_{j \in \cBs(r)}N_j}} \tagcmt{use $K \ge 2$} \\
    \stackrel{}{=}\,& 48\del{\fr{\log(K)}{\min_{r \in \cR}\sum_{j \in \cBs(r)}N_j}}.
  \end{align*}

  For the second term, we have
  \begin{align*}
    &\PP\del{\fr{1}{\sum_{j \in \cB}N_j} \sum_{i \in \cB \sm \cBs(\hr)}N_i\del{\muhat_i - \mu_1} > \fr{\eps}{2},\, \hr \in \cR,\, G_1} \\
    \stackrel{}{=}\,& \PP\del{\fr{1}{\sum_{j \in \cB}N_j} \sum_{i \in \cB \sm \cBs(\hr)}N_i\del{\muhat_i - \mu_i} > \fr{\eps}{2} + \fr{1}{\sum_{j \in \cB}N_j} \sum_{i \in \cB \sm \cBs(\hr)}N_i\Delta_i,\, \hr \in \cR,\, G_1} \\
    \stackrel{}{\le}\,& \PP\del{\fr{1}{\sum_{j \in \cB}N_j} \sum_{i \in \cB \sm \cBs(\hr)}N_i\del{\muhat_i - \mu_i} > \fr{\eps}{2},\, \hr \in \cR,\, G_1} \\
    \stackrel{}{\le}\,& \sum_{r \in \cR} \sum_{k=1}^{d(r)}\sum_{\substack{S: \cBs(r) \subset \Scal \subseteq \cBp(r) \\ \abs{\Scal}=n_*(r)+k }} \PP\del{\fr{1}{\sum_{j \in \Scal}N_j} \sum_{i \in \Scal \sm \cBs(r)}N_i\del{\muhat_i - \mu_i} > \fr{\eps}{2},\, G_1} \tagcmt{use $d(r) = \abs{\cBp(r)} - \abs{\cBs(r)}$ and $n_*(r) = \abs{\cBs(r)}$} \\
    \stackrel{}{\le}\,& \sum_{r \in \cR} \sum_{k=1}^{d(r)}\sum_{\substack{S: \cBs(r) \subset \Scal \subseteq \cBp(r) \\ \abs{\Scal}=n_*(r)+k }} \exp\del{-\fr{1}{2}\del{\fr{\del{\sum_{j \in \Scal}N_j}^2}{\sum_{j \in \Scal \sm \cBs(\hr)}N_j}}\del{\fr{\eps}{2}}^2} \\
    \stackrel{}{=}\,& \sum_{r \in \cR} \sum_{k=1}^{d(r)}\sum_{\substack{S: \cBs(r) \subset \Scal \subseteq \cBp(r) \\ \abs{\Scal}=n_*(r)+k }} \exp\del{-\fr{1}{8}\del{\fr{\del{\sum_{j \in \Scal}N_j}^2}{\sum_{j \in \Scal \sm \cBs(r)}N_j}}\eps^2} \\
    \stackrel{}{\le}\,& \sum_{r \in \cR}\sum_{k=1}^{d(r)}\sum_{\substack{S: \cBs(r) \subset \Scal \subseteq \cBp(r) \\ \abs{\Scal}=n_*(r)+k }} \exp\del{-\fr{1}{8}\del{\min_{r \in \cR} \min_{k=1}^{d(r)}\min_{\substack{S: \cBs(r) \subset \Scal \subseteq \cBp(r) \\ \abs{\Scal}=n_*(r)+k }}\fr{\del{\sum_{j \in \Scal}N_j}^2}{\sum_{j \in \Scal \sm \cBs(r)}N_j}}\eps^2} \\
    \stackrel{}{=}\,& \sum_{r \in \cR}\sum_{k=1}^{d(r)}\binom{d(r)}{k} \exp\del{-\fr{1}{8}\del{\min_{r \in \cR} \min_{k=1}^{d(r)}\min_{\substack{S: \cBs(r) \subset \Scal \subseteq \cBp(r) \\ \abs{\Scal}=n_*(r)+k }}\fr{\del{\sum_{j \in \Scal}N_j}^2}{\sum_{j \in \Scal \sm \cBs(r)}N_j}}\eps^2}.
  \end{align*}

  For brevity, we denote
  \begin{align*}
    M = \min_{r \in \cR} \min_{k=1}^{d(r)}\min_{\substack{S: \cBs(r) \subset \Scal \subseteq \cBp(r) \\ \abs{\Scal}=n_*(r)+k }}\fr{\del{\sum_{j \in \Scal}N_j}^2}{\sum_{j \in \Scal \sm \cBs(r)}N_j}
  \end{align*}
  
  For this term, let $q_1 = q_2 = 1$, $z_1 = \fr{M}{32}$, and 
  \begin{align*}
    \eps_0 = \sqrt{\displaystyle \max_{r \in \cR} \max_{k=1}^{d(r)} \max_{\substack{S: \cBs(r) \subset \Scal \subseteq \cBp(r) \\ \abs{\Scal}=n_*(r)+k }}\fr{32k\sum_{j \in \Scal \sm \cBs(r)}N_j}{\del{\sum_{j \in \Scal}N_j}^2}\log\del{\fr{ed(r)}{k}}\log\del{K^2}}.
  \end{align*}

  We claim that, in the regime of $\eps \ge \eps_0$, we have 
  \begin{align*}
    \PP\del{\fr{1}{\sum_{j \in \cB}N_j} \sum_{i \in \cB \sm \cBs(r)}N_i\del{\muhat_i - \mu_1} > \fr{\eps}{2},\, \hr \in \cR,\, G_1} \le q_1\exp\del{-z_1\eps^2}
  \end{align*}
  and in the regime of $\eps < \eps_0$, we have
  \begin{align*}
    \PP\del{\fr{1}{\sum_{j \in \cB}N_j} \sum_{i \in \cB \sm \cBs(r)}N_i\del{\muhat_i - \mu_1} > \fr{\eps}{2},\, \hr \in \cR,\, G_1} \le q_2.
  \end{align*}
  The second claim is trivial. We prove the first claim as follows. In the regime of $\eps \ge \eps_0$, we have 
  \begin{align*}
    &\PP\del{\fr{1}{\sum_{j \in \cB}N_j} \sum_{i \in \cB \sm \cBs(r)}N_i\del{\muhat_i - \mu_1} > \fr{\eps}{2},\, \hr \in \cR,\, G_1} \\
    \stackrel{}{\le}\,& \sum_{r \in \cR} \sum_{k=1}^{d(r)}\binom{d(r)}{k} \exp\del{-\fr{1}{8}M\eps^2} \\
    \stackrel{}{\le}\,& \sum_{r \in \cR} \sum_{k=1}^{d(r)}\del{\fr{ed(r)}{k}}^k \exp\del{-\fr{1}{8}M\eps^2} \tagcmt{use Stirling's formula, Lemma ~\ref{xlemma:stirling_formula}} \\
    \stackrel{}{=}\,& \sum_{r \in \cR} \sum_{k=1}^{d(r)} \exp\del{-\fr{1}{8}M\eps^2 + k\log\del{\fr{ed(r)}{k}}} \\
    \stackrel{}{\le}\,& \sum_{r \in \cR} \sum_{k=1}^{d(r)} \exp\del{-\fr{1}{16}M\eps^2} \tagcmt{use $\eps \ge \eps_0$}\\
    \stackrel{}{\le}\,& \sum_{r \in \cR} \sum_{k=1}^{d(r)} \exp\del{-\fr{1}{16}M\eps^2}  \\
    \stackrel{}{=}\,& Kd \exp\del{-\fr{1}{16}M\eps^2} \\
    \stackrel{}{=}\,& \exp\del{-\fr{1}{16}M\eps^2 + \ln(Kd)} \\
    \stackrel{}{\le}\,& \exp\del{-\fr{1}{32}M\eps^2 } \tagcmt{use $\eps \ge \eps_0$}.
  \end{align*}

  With the above claim, we use Lemma~\ref{lemma401:utility_integral_with_eps} to bound the integral 
  \begin{align*}
    &\int_{0}^{\infty} \PP\del{\fr{1}{\sum_{j \in \cB}N_j} \sum_{i \in \cB \sm \cBs(r)}N_i\del{\muhat_i - \mu_1} > \fr{\eps}{2},\, \hr \in \cR,\, G_1} \difeps \\
    \stackrel{}{\le}\,& q_1\eps_0^2 + q_2\fr{1}{z_1} \\
    \stackrel{}{=}\,& \del{\sqrt{\displaystyle \max_{r \in \cR} \max_{k=1}^{d(r)} \max_{\substack{S: \cBs(r) \subset \Scal \subseteq \cBp(r) \\ \abs{\Scal}=n_*(r)+k }}\fr{32k\sum_{j \in \Scal \sm \cBs(r)}N_j}{\del{\sum_{j \in \Scal}N_j}^2}\log\del{\fr{ed(r)}{k}}\log\del{K^2}}}^2 + \fr{32}{M} \\
    \stackrel{}{=}\,& 32\del{\max_{r \in \cR} \max_{k=1}^{d(r)} \max_{\substack{S: \cBs(r) \subset \Scal \subseteq \cBp(r) \\ \abs{\Scal}=n_*(r)+k }}\fr{32k\sum_{j \in \Scal \sm \cBs(r)}N_j}{\del{\sum_{j \in \Scal}N_j}^2}\log\del{\fr{ed(r)}{k}}\log\del{K^2}} \\
    &+ 32\del{\max_{r \in \cR} \max_{k=1}^{d(r)} \max_{\substack{S: \cBs(r) \subset \Scal \subseteq \cBp(r) \\ \abs{\Scal}=n_*(r)+k }}\fr{\sum_{j \in \Scal \sm \cBs(r)}N_j}{\del{\sum_{j \in \Scal}N_j}^2}} \tagcmt{plug in $M$} \\
    \stackrel{}{\le}\,& 64\del{\displaystyle \max_{r \in \cR} \max_{k=1}^{d(r)} \max_{\substack{S: \cBs(r) \subset \Scal \subseteq \cBp(r) \\ \abs{\Scal}=n_*(r)+k }}\fr{k\sum_{j \in \Scal \sm \cBs(r)}N_j}{\del{\sum_{j \in \Scal}N_j}^2}\log\del{\fr{ed(r)}{k}}\log\del{K^2}} \tagcmt{use $1 \le \log\del{\fr{ed(r)}{k}}\, \forall k \in [1,d]$ and $1 \le \log(K^2)\, \forall K \ge 2$} \\
    \stackrel{}{=}\,& 128\del{\displaystyle \max_{r \in \cR} \max_{k=1}^{d(r)} \max_{\substack{S: \cBs(r) \subset \Scal \subseteq \cBp(r) \\ \abs{\Scal}=n_*(r)+k }}\fr{k\sum_{j \in \Scal \sm \cBs(r)}N_j}{\del{\sum_{j \in \Scal}N_j}^2}\log\del{\fr{ed(r)}{k}}\log(K)}.
  \end{align*}

Combining all the terms, we have 
\begin{align*}
  &\int_{0}^{\infty} \PP\del{ \hmuB - \mu_1 > \eps,\, \hr \in \cR,\, G_1 } \difeps \\
  \stackrel{}{\le}\,& \int_{0}^{\infty} \PP\del{ \fr{1}{\sum_{j \in \cB}N_j}\sum_{i \in \cBs(r)}N_i\del{\muhat_i - \mu_1} > \fr{\eps}{2},\, \hr \in \cR,\, G_1 } \difeps \\
  &+ \int_{0}^{\infty} \PP\del{ \fr{1}{\sum_{j \in \cB}N_j}\sum_{i \in \cB \sm \cBs(r)}N_i\del{\muhat_i - \mu_1} > \fr{\eps}{2},\, \hr \in \cR,\, G_1 } \difeps \\
  \stackrel{}{\le}\,& 48\del{\fr{\log(K)}{\min_{r \in \cR}\sum_{j \in \cBs(r)}N_j}} + 128\del{\max_{r \in \cR} \max_{k=1}^{d(r)} \max_{\substack{S: \cBs(r) \subset \Scal \subseteq \cBp(r) \\ \abs{\Scal}=n_*(r)+k }}\fr{k\sum_{j \in \Scal \sm \cBs(r)}N_j}{\del{\sum_{j \in \Scal}N_j}^2}\log\del{\fr{ed(r)}{k}}\log(K)} \\
  \stackrel{}{\le}\,& 48\del{\fr{\log(K)}{\min_{r \in \cR}\sum_{j \in \cBs(r)}N_j}} + 128\del{\max_{r \in \cR} \max_{k=0}^{d(r)} \max_{\substack{S: \cBs(r) \subseteq \Scal \subseteq \cBp(r) \\ \abs{\Scal}=n_*(r)+k }}\fr{k\sum_{j \in \Scal \sm \cBs(r)}N_j}{\del{\sum_{j \in \Scal}N_j}^2}\log\del{\fr{ed(r)}{k}}\log(K)}
\end{align*}

\end{proof}

\begin{lemma}\label{lemma_g2b:haver21count2}
  If event $G$ sastisfies $\PP\del{G} \le \fr{2K^2}{K^{2\lam}}$, then HAVER achieves
  \begin{align*}
    \int_{0}^{\infty} \PP\del{ \hmuB - \mu_1 > \eps,\, G } \difeps \le \fr{3088}{KN_1}\log\del{KS}.
  \end{align*}
\end{lemma}

\begin{proof}
  We have
  \begin{align*}
    &\int_{0}^{\infty} \PP\del{ \hmuB - \mu_1 > \eps, G } \difeps \\
    \stackrel{}{\le}\,& \int_{0}^{\infty} \PP\del{\hr =  1,\, \hmuB - \mu_1 > \eps, G } \difeps \\
    &+ \int_{0}^{\infty} \sum_{\substack{i: i \ne 1, \\ \gam_i \le 2\gam_1}}\PP\del{\hr = i,\, \hmuB - \mu_{i} > \eps, G} \difeps \\
    &+ \int_{0}^{\infty} \sum_{\substack{i: i \ne 1, \\ \gam_i > 2\gam_1}}\PP\del{\hr = i,\, \hmuB - \mu_{i} > \eps, G} \difeps.
  \end{align*}

  For the first term, we have
  \begin{align*}
    &\PP\del{\hr =  1,\, \hmuB - \mu_1 > \eps, G } \\
    \stackrel{}{\le}\,& \PP\del{\hr =  1,\, \hmuB - \mu_1 > \eps } \\
    \stackrel{}{\le}\,& \exp\del{-\fr{1}{2}N_1\del{\eps - \fr{1}{2}\gam_1}_+^2} \tagcmt{use Lemma~\ref{lemma101.b.1:haver}} .
  \end{align*}
  We also have $\PP\del{G} \le \fr{2K^2}{K^{2\lam}}$, thus
  \begin{align*}
    \PP\del{\hr =  1,\, \hmuB - \mu_1 > \eps, G } \le \exp\del{-\fr{1}{2}N_1\del{\eps - \gam_1}_+^2} \wedge \fr{2K^2}{K^{2\lam}}.
  \end{align*}
  Let $q_1 = \fr{2K}{K^{\lam}}$, $q_2 = \fr{2K^2}{K^{2\lam}}$, $z_1 = \fr{N_1}{16}$, and $\eps_0 = \gam_1$.
  We claim that in the regime of $\eps \ge \eps_0$, we have
  \begin{align*}
    \PP\del{\hr =  1,\, \hmuB - \mu_1 > \eps, G } \le q_1\exp\del{-z_1\eps^2}
  \end{align*}
  and in the regime of $\eps < \eps_0$, we have $\PP\del{\hr =  1,\, \hmuB - \mu_1 > \eps, G } \le q_2$.

  We prove the claim as follows. In the regime of $\eps \ge \eps_0$, we have 
  \begin{align*}
    &\PP\del{\hr =  1,\, \hmuB - \mu_1 > \eps, G } \\
    \stackrel{}{\le}\,& \exp\del{-\fr{1}{2}N_1\del{\eps - \fr{1}{2}\gam_1}_+^2} \wedge \fr{2K^2}{K^{2\lam}} \\
    \stackrel{}{\le}\,& \sqrt{\exp\del{-\fr{1}{2}N_1\del{\eps - \fr{1}{2}\gam_1}_+^2} \cdot \fr{2K^2}{K^{2\lam}}} \\
    \stackrel{}{=}\,& \fr{2K}{K^{\lam}}\exp\del{-\fr{1}{4}N_1\del{\eps - \fr{1}{2}\gam_1}_+^2} \\
    \stackrel{}{\le}\,& \fr{2K}{K^{\lam}}\exp\del{-\fr{1}{4}N_1\del{\fr{\eps}{2}}^2} \tagcmt{use $\eps \ge \gam_1$}\\
    \stackrel{}{=}\,& \fr{2K}{K^{\lam}}\exp\del{-\fr{1}{16}N_1\eps^2}.
  \end{align*}
  In the regime of $\eps < \eps_0$, we have
  \begin{align*}
    &\PP\del{\hr = 1,\, \hmuB - \mu_1 > \eps, G } \\
    \stackrel{}{\le}\,& \exp\del{-\fr{1}{2}N_1\del{\eps - \fr{1}{2}\gam_1}_+^2} \wedge \fr{2K^2}{K^{2\lam}} \\
    \stackrel{}{\le}\,& \fr{2K^2}{K^{2\lam}}. 
  \end{align*}
  With the above claim, we use Lemma~\ref{lemma401:utility_integral_with_eps} to bound the integral
  \begin{align*}
    &\int_{0}^{\infty}\PP\del{\hr = 1,\, \hmuB - \mu_1 > \eps, G } \difeps \\
    \stackrel{}{\le}\,& q_2\eps_0^2 + q_1\fr{1}{z_1} \\
    \stackrel{}{=}\,& \fr{2K^2}{K^{2\lam}}\gam_1^2 + \fr{2K}{K^{\lam}}\fr{16}{N_1}  \\
    \stackrel{}{=}\,& \fr{2K^2}{K^{2\lam}}\fr{18}{N_1}\log\del{\del{\fr{KS}{N_1}}^{2\lam}} + \fr{2K}{K^{\lam}}\fr{16}{N_1} \tagcmt{use $\gam_i = \sqrt{\fr{18}{N_i}\log\del{\del{\fr{KS}{N_i}}^{2\lam}}}$} \\
    \stackrel{}{=}\,& \fr{36K^2}{K^{4}N_1}\log\del{\del{\fr{KS}{N_1}}^{4}} + \fr{32K}{K^{2}N_1} \tagcmt{recall $\lam = 2$} \\
    \stackrel{}{=}\,& \fr{36}{K^{2}N_1}\log\del{\del{\fr{KS}{N_1}}^{4}} + \fr{32}{KN_1}  \\
    \stackrel{}{=}\,& \fr{68}{KN_1}\log\del{\del{\fr{KS}{N_1}}^{4}} \tagcmt{use $1 \le \log\del{\del{\fr{KS}{N_1}}^{4}}$ with $K \ge 2$}  \\
    \stackrel{}{=}\,& \fr{272}{KN_1}\log\del{\fr{KS}{N_1}}.
  \end{align*}

  For the second term, we denote 
  \begin{align*}
    h(\eps) = \sum_{\substack{i: i \ne 1, \\ \gam_i \le 2\gam_1}}\PP\del{\hr = i,\, \hmuB - \mu_{i} > \eps, G}
  \end{align*}
  by the probability of interest. We have
  \begin{align*}
    &h(\eps) \\
    \stackrel{}{=}\,& \sum_{\substack{i: i \ne 1, \\ \gam_i \le 2\gam_1}}\PP\del{\hr = i,\, \hmuB - \mu_{i} > \eps, G} \\
    \stackrel{}{\le}\,& \sum_{\substack{i: i \ne 1, \\ \gam_i \le 2\gam_1}}\PP\del{\hr = i,\, \hmuB - \mu_{i} > \eps} \\
    \stackrel{}{\le}\,& \sum_{\substack{i: i \ne 1, \\ \gam_i \le 2\gam_1}} \exp\del{-\fr{1}{2}N_i\del{\eps - \gam_1 + \Delta_i}_+^2} \tagcmt{use Lemma~\ref{lemma101.b.2:haver}} \\
    \stackrel{}{\le}\,& \sum_{\substack{i: i \ne 1, \\ \gam_i \le 2\gam_1}} \exp\del{-\fr{1}{2}N_i\del{\eps - \gam_1}_+^2} \\
    \stackrel{}{\le}\,& K \exp\del{-\fr{1}{2}\min_{\substack{i: i \ne 1, \\ \gam_i \le 2\gam_1}}\del{\eps - \gam_1}_+^2} \\
    \stackrel{}{=}\,& K \exp\del{-\fr{1}{2}N_u\del{\eps - \gam_1}_+^2}. 
  \end{align*}
  We also have $\PP\del{G} \le \fr{2K^2}{K^{2\lam}}$ thus
  \begin{align*}
    &h(\eps) \\
    \stackrel{}{=}\,& \sum_{\substack{i: i \ne 1, \\ \gam_i \le 2\gam_1}}\PP\del{\hr = i,\, \hmuB - \mu_{i} > \eps, G} \\
    \stackrel{}{\le}\,& \sum_{\substack{i: i \ne 1, \\ \gam_i \le 2\gam_1}}\PP\del{G} \\
    \stackrel{}{\le}\,& \sum_{\substack{i: i \ne 1, \\ \gam_i \le 2\gam_1}} \fr{2K^2}{K^{2\lam}} \tagcmt{use $\PP\del{G} \le \fr{2K^2}{K^{2\lam}}$} \\
    \stackrel{}{\le}\,& \fr{2K^3}{K^{2\lam}}.  
  \end{align*}
  Let $q_1 = \fr{2K}{K^{\lam}}$, $q_2 = \fr{2K^3}{K^{2\lam}}$, $z_1 = \fr{N_u}{32}$, and $\eps_0 = 2\gam_1 \vee \sqrt{\fr{32\log(K)}{N_u}}$.
  We claim that in the regime of $\eps \ge \eps_0$, we have
  \begin{align*}
    h(\eps) \le q_1\exp\del{-z_1\eps^2}
  \end{align*}
  and in the regime of $\eps < \eps_0$, we have $h(\eps) \le q_2$.

  We prove the claim as follows. In the regime of $\eps \ge \eps_0$, we have 
  \begin{align*}
    &h(\eps) \\
    \stackrel{}{\le}\,& K\exp\del{-\fr{1}{2}N_u\del{\eps - \gam_1}_+^2} \wedge \fr{2K^3}{K^{2\lam}} \\
    \stackrel{}{\le}\,& \sqrt{K\exp\del{-\fr{1}{2}N_u\del{\eps - \gam_1}_+^2} \cdot \fr{2K^3}{K^{2\lam}}} \\
    \stackrel{}{=}\,& \exp\del{-\fr{1}{4}N_u\del{\eps - \gam_1}_+^2} \\
    \stackrel{}{\le}\,& \fr{2K^2}{K^{\lam}}\exp\del{-\fr{1}{4}N_u\del{\fr{\eps}{2}}^2} \tagcmt{use $\eps \ge \eps_0 = 2\gam_1$}\\
    \stackrel{}{=}\,& \fr{2K^2}{K^{\lam}}\exp\del{-\fr{1}{16}N_u\eps^2} \\
    \stackrel{}{=}\,& \fr{2K}{K^{\lam}}\exp\del{-\fr{1}{16}N_u\eps^2 + \log(K)} \\
    \stackrel{}{\le}\,& \fr{2K}{K^{\lam}}\exp\del{-\fr{1}{32}N_u\eps^2} \tagcmt{use $\eps \ge \eps_0 \ge \sqrt{\fr{32\log(K)}{N_u}}$}. 
  \end{align*}
  In the regime of $\eps < \eps_0$, we have
  \begin{align*}
    &h(\eps) \\
    \stackrel{}{\le}\,& \exp\del{-\fr{1}{2}N_u\del{\eps - \gam_1}_+^2} \wedge \fr{2K^3}{K^{2\lam}} \\
    \stackrel{}{\le}\,& \fr{2K^3}{K^{2\lam}}. 
  \end{align*}
  With the above claim, we use Lemma~\ref{lemma401:utility_integral_with_eps} to bound the integral
  \begin{align*}
    &\int_{0}^{\infty}h(\eps) \difeps \\
    \stackrel{}{\le}\,& q_2\eps_0^2 + q_1\fr{1}{z_1} \\
    \stackrel{}{=}\,& \fr{2K^3}{K^{2\lam}}\del{2\gam_1 \vee \sqrt{\fr{32\log(K)}{N_u}}}^2 + \fr{2K}{K^{\lam}}\fr{32}{N_u} \\
    \stackrel{}{\le}\,& \gam_1^2 + \fr{8K^3}{K^{2\lam}}\fr{32\log(K)}{N_u} + \fr{2K}{K^{\lam}}\fr{32}{N_u}  \\
    \stackrel{}{=}\,& \fr{8K^3}{K^{2\lam}}\fr{18}{N_1}\log\del{\del{\fr{KS}{N_1}}^{2\lam}} + \fr{8K^3}{K^{2\lam}}\fr{32\log(K)}{N_u} + \fr{2K}{K^{\lam}}\fr{32}{N_u} \tagcmt{use $\gam_i = \sqrt{\fr{18}{N_i}\log\del{\del{\fr{KS}{N_i}}^{2\lam}}}$} \\
    \stackrel{}{=}\,& \fr{144K^3}{K^{4}N_1}\log\del{\del{\fr{KS}{N_1}}^{4}} + \fr{256K^3\log(K)}{K^{4}N_u} + \fr{64K}{K^{2}N_u} \tagcmt{recall $\lam = 2$} \\
    \stackrel{}{=}\,& \fr{144}{KN_1}\log\del{\del{\fr{KS}{N_1}}^{4}} + \fr{128\log(K^2)}{KN_u} + \fr{64}{KN_u} \\
    \stackrel{}{\le}\,& \fr{144}{KN_1}\log\del{\del{\fr{KS}{N_1}}^{4}} + \fr{192\log(K^2)}{KN_u} \tagcmt{use $1 \le \log(K^2)$ with $K \ge 2$} \\
    \stackrel{}{\le}\,& \fr{144}{KN_1}\log\del{\del{\fr{KS}{N_1}}^{4}} + \fr{384\log(K)}{KN_u}. 
  \end{align*}

  Lemma~\ref{lemma:maxlcb_101} states that for any $i \ne 1$ the condition $\gam_i < 2\gam_1$ implies that $N_i > \fr{N_1}{4} \fr{\log\del{K}}{\log\del{\fr{KS}{N_1}}}$. Thus, $N_u \ge \fr{N_1}{4} \fr{\log\del{K}}{\log\del{\fr{KS}{N_1}}}$.
  \begin{align*}
    &\fr{144}{KN_1}\log\del{\del{\fr{KS}{N_1}}^{4}} + \fr{384\log(K)}{KN_u} \\
    \stackrel{}{\le}\,& \fr{144}{KN_1}\log\del{\del{\fr{KS}{N_1}}^{4}} + \fr{384\log(K)}{KN_1}\fr{4\log\del{\fr{KS}{N_1}}}{\log\del{K}} \\
    \stackrel{}{=}\,& \fr{144}{KN_1}\log\del{\del{\fr{KS}{N_1}}^{4}} + \fr{1436}{KN_1}\log\del{\fr{KS}{N_1}} \\
    \stackrel{}{=}\,& \fr{2112}{KN_1}\log\del{\fr{KS}{N_1}}. 
  \end{align*}

  For the third term, we focus on arm $i$ such that $i \ne 1,\, \gam_i > 2\gam_1$. We have 
  \begin{align*}
    &\PP\del{\hr = i, \hmuB - \mu_{i} > \eps, G} \\
    \stackrel{}{\le}\,& \PP\del{\hr = i, \hmuB - \mu_{i} > \eps} \\
    \stackrel{}{\le}\,& \PP\del{\hr = i, \hmu_{i} - \mu_{i} > \eps - \fr{1}{2}\gam_{i} + \Delta_{i}} \tagcmt{use Lemma~\ref{lemma100.b:haver21count2}} \\
    \stackrel{}{\le}\,& \exp\del{-\fr{1}{2}N_i\del{\eps - \gam_i + \Delta_i}_+^2}. 
  \end{align*}
  Lemma~\ref{lemma109:haver} states that
  \begin{align*}
    \PP\del{\hr = i,\, \hmu_{\hr} - \gam_{\hr} > \hmu_1 - \gam_1 } \le \del{\fr{N_i}{KS}}^{2\lam}. 
  \end{align*}
  We also have 
  \begin{align*}
    &\PP\del{\hr = i, \hmuB - \mu_{i} > \eps, G} \\
    \stackrel{}{\le}\,& \PP\del{G} \\
    \stackrel{}{\le}\,& \fr{2K^2}{K^{2\lam}}. 
  \end{align*}
  Thus, we have
  \begin{align*}
    &\PP\del{\hr = i, \hmuB - \mu_{i} > \eps, G} \\
    \stackrel{}{\le}\,& \exp\del{-\fr{1}{2}N_i\del{\eps - \gam_i + \Delta_i}_+^2} \wedge \del{\fr{N_i}{KS}}^{2\lam} \wedge \fr{2K^2}{K^{2\lam}}. 
  \end{align*}

  Let $q_1 = \fr{2K}{K^{\lam}}\del{\fr{N_i}{KS}}^{\lam}$, $q_2 = \fr{2K}{K^{\lam}}\del{\fr{N_i}{KS}}^{\lam}$, $z_1 = \fr{N_i}{16}$, and $\eps_0 = 2\gam_i$.
  We claim that in the regime of $\eps \ge \eps_0$, we have
  \begin{align*}
    \PP\del{\hr = i, \hmuB - \mu_{i} > \eps, G} \le q_1\exp\del{-z_1\eps^2}
  \end{align*}
  and in the regime of $\eps < \eps_0$, we have $\PP\del{\hr = i, \hmuB - \mu_{i} > \eps, G} \le q_2$.

  We prove the claim as follows. In the regime of $\eps \ge \eps_0$, we have 
  \begin{align*}
    &\PP\del{\hr = i, \hmuB - \mu_{i} > \eps, G} \\
    \stackrel{}{\le}\,& \exp\del{-\fr{1}{2}N_i\del{\eps - \gam_i}_+^2} \wedge \del{\fr{N_i}{KS}}^{2\lam} \wedge \fr{2K^2}{K^{2\lam}} \\
    \stackrel{}{\le}\,& \sqrt{\exp\del{-\fr{1}{2}N_i\del{\eps - \gam_i}_+^2} \cdot \del{\fr{N_i}{KS}}^{2\lam} \cdot \fr{2K^2}{K^{2\lam}}} \\
    \stackrel{}{\le}\,& \del{\fr{N_i}{KS}}^{\lam}\fr{2K}{K^{\lam}}\exp\del{-\fr{1}{4}N_i\del{\eps - \gam_i}_+^2} \\
    \stackrel{}{\le}\,& \fr{2K}{K^{\lam}}\del{\fr{N_i}{KS}}^{\lam}\exp\del{-\fr{1}{16}N_i\eps^2} \tagcmt{use $\eps \ge eps_0 = 2\gam_i$}.
  \end{align*}

  In the regime of $\eps < \eps_0$, we have 
  \begin{align*}
    &\PP\del{\hr = i, \hmuB - \mu_{i} > \eps, G} \\
    \stackrel{}{\le}\,& \del{\fr{N_i}{KS}}^{2\lam} \wedge \fr{2K^2}{K^{2\lam}} \\
    \stackrel{}{\le}\,& \sqrt{\del{\fr{N_i}{KS}}^{2\lam} \cdot \fr{2K^2}{K^{2\lam}}} \\
    \stackrel{}{\le}\,& \fr{2K}{K^{\lam}} \del{\fr{N_i}{KS}}^{\lam}. 
  \end{align*}
  With the claim above, we use Lemma~\ref{lemma401:utility_integral_with_eps} to bound the integral
  \begin{align*}
    &\int_{0}^{\infty} \PP\del{\hr = i,\, \hmuB - \mu_1 > \eps, G} \difeps \\
    \stackrel{}{\le}\,& q_2\eps_0^2 + q_1\fr{1}{z_1} \\
    \stackrel{}{\le}\,& \fr{2K}{K^{\lam}}\del{\fr{N_i}{KS}}^{\lam}4\gam_i^2 + \fr{2K}{K^{\lam}}\del{\fr{N_i}{KS}}^{\lam}\fr{16}{N_i} \\
    \stackrel{}{\le}\,& \fr{8K}{K^{\lam}}\del{\fr{N_i}{KS}}^{\lam}\fr{18}{N_i}\log\del{\del{\fr{KS}{N_i}}^{2\lam}} + \fr{2K}{K^{\lam}}\del{\fr{N_i}{KS}}^{\lam}\fr{16}{N_i} \\
    \stackrel{}{=}\,& \fr{8K}{K^{\lam}}\del{\fr{N_i}{K\Nmax\sum_{j \in [K]}N_j}}^{\lam}\fr{18}{N_i}\log\del{\del{\fr{KS}{N_i}}^{2\lam}} + \fr{2K}{K^{\lam}}\del{\fr{N_i}{K\Nmax\sum_{j \in [K]}N_j}}^{\lam}\fr{16}{N_i} \tagcmt{use $S = \Nmax\sum_{j \in [K]}N_j$} \\
    \stackrel{}{\le}\,& \fr{144K}{K^{\lam}N_i}\del{\fr{1}{K\sum_{j \in [K]}N_j}}^{\lam}\log\del{\del{\fr{KS}{N_i}}^{2\lam}} + \fr{32K}{K^{\lam}N_i}\del{\fr{1}{K\sum_{j \in [K]}N_j}}^{\lam} \\
    \stackrel{}{=}\,& \fr{144K}{K^{2}N_i}\del{\fr{1}{K\sum_{j \in [K]}N_j}}^{2}\log\del{\del{\fr{KS}{N_i}}^{4}} + \fr{32K}{K^{2}N_i}\del{\fr{1}{K\sum_{j \in [K]}N_j}}^{2} \tagcmt{recall $\lam = 2$} \\
    \stackrel{}{=}\,& \fr{144}{K^3\del{\sum_{j \in [K]}N_j}^2}\log\del{\del{KS}^{4}} + \fr{32}{K^3\del{\sum_{j \in [K]}N_j}^2} \tagcmt{use $N_i \ge 1$} \\
    \stackrel{}{\le}\,& \fr{176}{K^3\del{\sum_{j \in [K]}N_j}^2}\log\del{\del{KS}^{4}} \tagcmt{use $1 \le \log(\del{KS}^4)$ with $K \ge 2$} \\
    \stackrel{}{=}\,& \fr{704}{K^3\del{\sum_{j \in [K]}N_j}^2}\log\del{\del{KS}}.
  \end{align*}

  Thus, for the third term, we obtain the integral
  \begin{align*}
    &\sum_{\substack{i: i \ne 1, \\ \gam_i > 2\gam_1}} \int_{0}^{\infty} \PP\del{\hr = i,\, \hmuB - \mu_1 > \eps, G} \difeps \\
    \stackrel{}{\le}\,& \sum_{\substack{i: i \ne 1, \\ \gam_i > 2\gam_1}} \fr{704}{K^3\del{\sum_{j \in [K]}N_j}^2}\log\del{KS} \\
    \stackrel{}{\le}\,& \fr{704}{K^2\del{\sum_{j \in [K]}N_j}^2}\log\del{KS}. 
  \end{align*}

  Combining all the terms, we have
  \begin{align*}
    &\int_{0}^{\infty} \PP\del{ \hmuB - \mu_1 > \eps, G } \difeps \\
    \stackrel{}{\le}\,& \int_{0}^{\infty} \PP\del{\hr =  1,\, \hmuB - \mu_1 > \eps, G } \difeps \\
    &+ \int_{0}^{\infty} \sum_{\substack{i: i \ne 1, \\ \gam_i \le 2\gam_1}}\PP\del{\hr = i,\, \hmuB - \mu_{i} > \eps, G} \difeps \\
    &+ \int_{0}^{\infty} \sum_{\substack{i: i \ne 1, \\ \gam_i > 2\gam_1}}\PP\del{\hr = i,\, \hmuB - \mu_{i} > \eps, G} \difeps \\
    \stackrel{}{\le}\,& \fr{272}{KN_1}\log\del{\fr{KS}{N_1}} \\
    &+ \fr{2112}{KN_1}\log\del{\fr{KS}{N_1}} \\
    &+ \fr{704}{K^2\del{\sum_{j \in [K]}N_j}^2}\log\del{KS} \\
    \stackrel{}{\le}\,& \fr{3088}{KN_1}\log\del{KS}.
  \end{align*}
  
\end{proof}

\begin{lemma}\label{lemma100.b:haver21count2}
  In HAVER, we have
  \begin{align*}
    \hmu_{\hr} - \gam_{\hr} \le \hmuB \le \hmu_{\hr} + \fr{1}{2}\gam_{\hr}.
  \end{align*}
\end{lemma}

\begin{proof}
  For the first inequality, by the definition of $\cB$, $\forall i \in \cB,\, \hmu_i \ge \hmu_{\hr} - \gam_{\hr}$, therefore
  \begin{align*}
    &\sum_{i \in \cB} w_i \hmu_i \ge \hmu_{\hr} - \gam_{\hr} \\
    \stackrel{}{\Leftrightarrow}\,& \hmuB \ge \hmu_{\hr} - \gam_{\hr}.
  \end{align*}

  For the second inequality, by the definition of $\hr$, $\forall i \in \cB,\, \hmu_{\hr} - \gam_{\hr} \ge \hmu_i - \gam_i$, therefore $\forall i \in \cB$,
  \begin{align*}
    &\hmu_i \le \hmu_{\hr} - \gam_{\hr} + \gam_i \\
    \stackrel{}{\Rightarrow}\,& \hmu_i \le \hmu_{\hr} - \gam_{\hr} + \fgam \gam_{\hr}  \tagcmt{use $\forall i \in \cB,\, \gam_i  \le \fgam \gam_{\hr}$} \\
    \stackrel{}{\Leftrightarrow}\,& \hmu_i \le \hmu_{\hr} + \fr{1}{2} \gam_{\hr} \\
    \stackrel{}{\Leftrightarrow}\,& \sum_{i \in \cB} w_i \hmu_i \le \hmu_{\hr} + \fr{1}{2} \gam_{\hr} \\
    \stackrel{}{\Leftrightarrow}\,& \hmuB \le \hmu_{\hr} + \fr{1}{2} \gam_{\hr}.
  \end{align*}
  
\end{proof}

\begin{lemma}\label{lemma101.b.1:haver}
  In HAVER, we have
  \begin{align*}
    \PP\del{\hr = 1,\, \hmuB - \mu_1 > \eps } \le \exp\del{-\fr{1}{2}N_1\del{\eps - \fr{1}{2}\gam_1}_+^2}.
  \end{align*}
\end{lemma}

\begin{proof}
  We have
  \begin{align*}
    &\PP\del{\hr =  1,\, \hmuB - \mu_1 > \eps } \\
    \stackrel{}{\le}\,& \PP\del{\hr =  1,\, \hmu_{\hr} + \fr{1}{2}\gam_{\hr} - \mu_1 > \eps} \tagcmt{use Lemma~\ref{lemma100.b:haver21count2}, $\hmuB \le \hmu_{\hr} + \fr{1}{2}\gam_{\hr}$} \\
    \stackrel{}{=}\,& \PP\del{\hr =  1,\, \hmu_{1} - \mu_1 > \eps - \fr{1}{2}\gam_1} \\
    \stackrel{}{\le}\,& \exp\del{-\fr{1}{2}N_1\del{\eps - \fr{1}{2}\gam_1}_+^2}.
  \end{align*}
\end{proof}

\begin{lemma}\label{lemma101.b.2:haver}
  Let $i \in [K]$ be any arm such that $i \ne 1,\, \gam_i \le 2\gam_1$. In HAVER, we have
  \begin{align*}
    \PP\del{\hr = i,\, \hmuB - \mu_1 > \eps } \le \exp\del{-\fr{1}{2}N_i\del{\eps - \gam_1 + \Delta_i}_+^2}.
  \end{align*}
\end{lemma}

\begin{proof}
  We have
  \begin{align*}
    &\PP\del{\hr = i,\, \hmuB - \mu_{1} > \eps} \\
    \stackrel{}{\le}\,& \PP\del{\hr = i,\, \hmu_{\hr} + \fr{1}{2}\gam_{\hr} - \mu_{1} > \eps} \tagcmt{use Lemma~\ref{lemma100.b:haver21count2}, $\hmuB \le \hmu_{\hr} + \fr{1}{2}\gam_{\hr}$} \\
    \stackrel{}{=}\,& \PP\del{\hr = i,\, \hmu_{i} + \fr{1}{2}\gam_{i} - \mu_{1} > \eps } \\
    \stackrel{}{=}\,& \PP\del{\hr = i,\, \hmu_{i} - \mu_{1} > \eps - \fr{1}{2}\gam_{i}} \\
    \stackrel{}{\le}\,& \PP\del{\hr = i,\, \hmu_{i} - \mu_{i} > \eps - \fr{1}{2}\gam_{i} + \Delta_{i}} \\
    \stackrel{}{\le}\,& \PP\del{\hr = i,\, \hmu_{i} - \mu_{i} > \eps - \gam_{1} + \Delta_{i}} \tagcmt{use  $\gam_i \le 2\gam_1$} \\
    \stackrel{}{\le}\,& \exp\del{-\fr{1}{2}N_i\del{\eps - \gam_1 + \Delta_i}_+^2}. 
  \end{align*}
\end{proof}

\begin{lemma}\label{lemma101.b.3:haver}
  Let $i \in [K]$ be any arm such that $i \ne 1,\, \gam_i > 2\gam_1$. In HAVER, we have 
  \begin{align*}
    \PP\del{\hr = i,\, \hmuB - \mu_1 > \eps } \le \exp\del{-\fr{1}{2}N_i\del{\eps - \gam_i + \Delta_i}_+^2}.
  \end{align*}
\end{lemma}

\begin{proof}
  We have 
  \begin{align*}
    &\PP\del{\hr = i, \hmuB - \mu_{1} > \eps} \\
    \stackrel{}{\le}\,& \PP\del{\hr = i,\, \hmu_{\hr} + \fr{1}{2}\gam_{\hr} - \mu_{1} > \eps} \tagcmt{use Lemma~\ref{lemma100.b:haver21count2}, $\hmuB \le \hmu_{\hr} + \fr{1}{2}\gam_{\hr}$} \\
    \stackrel{}{=}\,& \PP\del{\hr = i,\, \hmu_{i} + \fr{1}{2}\gam_{i} - \mu_{1} > \eps} \\
    \stackrel{}{=}\,& \PP\del{\hr = i,\, \hmu_{i} - \mu_{1} > \eps - \fr{1}{2}\gam_{i} } \\
    \stackrel{}{=}\,& \PP\del{\hr = i,\, \hmu_{i} - \mu_{i} > \eps - \fr{1}{2}\gam_{i} + \Delta_i } \\
    \stackrel{}{\le}\,& \exp\del{-\fr{1}{2}N_i\del{\eps - \fr{1}{2}\gam_i + \Delta_i}_+^2} \\
    \stackrel{}{\le}\,& \exp\del{-\fr{1}{2}N_i\del{\eps - \gam_i + \Delta_i}_+^2}. 
  \end{align*}
\end{proof}

\begin{lemma}\label{lemma102.b:haver21count2}
  In HAVER, we have
  \begin{align*}
    \int_{0}^{\infty} \PP\del{ \hmuB - \mu_1 > \eps } \difeps \le \fr{720}{N_1}\log\del{KS}.
  \end{align*}
\end{lemma}

\begin{proof}
  We have
  \begin{align*}
    &\int_{0}^{\infty} \PP\del{ \hmuB - \mu_1 > \eps } \difeps \\
    \stackrel{}{\le}\,& \int_{0}^{\infty} \PP\del{\hr =  1,\, \hmuB - \mu_1 > \eps } \difeps \\
    &+ \int_{0}^{\infty} \sum_{\substack{i: i \ne 1, \\ \gam_i \le 2\gam_1}}\PP\del{\hr = i,\, \hmuB - \mu_{i} > \eps, G} \difeps \\
    &+ \int_{0}^{\infty} \sum_{\substack{i: i \ne 1, \\ \gam_i > 2\gam_1}}\PP\del{\hr = i,\, \hmuB - \mu_{i} > \eps, G} \difeps.
  \end{align*}

  For the first term, from Lemma~\ref{lemma101.b.1:haver}, we have
  \begin{align*}
    &\PP\del{\hr =  1,\, \hmuB - \mu_1 > \eps } \\
    \stackrel{}{\le}\,& \exp\del{-\fr{1}{2}N_1\del{\eps - \fr{1}{2}\gam_1}_+^2} 
  \end{align*}
  Let $q_1 = q_2 = 1$, $z_1 = \fr{N_1}{8}$, and $\eps_0 = \gam_1$.
  We claim that in the regime of $\eps \ge \eps_0$, we have
  \begin{align*}
    \PP\del{\hr =  1,\, \hmuB - \mu_1 > \eps} \le q_1\exp\del{-z_1\eps^2}
  \end{align*}
  and in the regime of $\eps < \eps_0$, we have $\PP\del{\hr =  1,\, \hmuB - \mu_1 > \eps, G } \le q_2$.
  
  The second claim is trivial. We prove the first claim as follows. In the regime of $\eps \ge \eps_0$, we have 
  \begin{align*}
    &\PP\del{\hr =  1,\, \hmuB - \mu_1 > \eps} \\
    \stackrel{}{\le}\,& \exp\del{-\fr{1}{2}N_1\del{\eps - \fr{1}{2}\gam_1}_+^2} \\
    \stackrel{}{\le}\,& \exp\del{-\fr{1}{2}N_1\del{\fr{\eps}{2}}^2} \tagcmt{use $\eps \ge \eps_0 = \gam_1$} \\
    \stackrel{}{=}\,& \exp\del{-\fr{1}{8}N_1\eps^2}. 
  \end{align*}
 
  With the above claim, we use Lemma~\ref{lemma401:utility_integral_with_eps} to bound the integral
  \begin{align*}
    &\int_{0}^{\infty}\PP\del{\hr = 1,\, \hmuB - \mu_1 > \eps} \difeps \\
    \stackrel{}{\le}\,& q_2\eps_0^2 + q_1\fr{1}{z_1} \\
    \stackrel{}{=}\,& \gam_1^2 + \fr{16}{N_1}  \\
    \stackrel{}{=}\,& \fr{18}{N_1}\log\del{\del{\fr{KS}{N_1}}^{2\lam}} + \fr{8}{N_1} \tagcmt{use $\gam_i = \sqrt{\fr{18}{N_i}\log\del{\del{\fr{KS}{N_i}}^{2\lam}}}$} \\
    \stackrel{}{=}\,& \fr{18}{N_1}\log\del{\del{\fr{KS}{N_1}}^{4}} + \fr{8}{N_1} \tagcmt{recall $\lam = 2$} \\
    \stackrel{}{=}\,& \fr{26}{N_1}\log\del{\del{\fr{KS}{N_1}}^{4}} \tagcmt{use $1 \le \log\del{\del{\fr{KS}{N_1}}^{4}}$ with $K \ge 2$}  \\
    \stackrel{}{=}\,& \fr{104}{N_1}\log\del{\fr{KS}{N_1}}.
  \end{align*}

  For the second term, we denote 
  \begin{align*}
    h(\eps) = \sum_{\substack{i: i \ne 1, \\ \gam_i \le 2\gam_1}}\PP\del{\hr = i,\, \hmuB - \mu_{i} > \eps}
  \end{align*}
  by the probability of interest. We have
  \begin{align*}
    &h(\eps) \\
    \stackrel{}{=}\,& \sum_{\substack{i: i \ne 1, \\ \gam_i \le 2\gam_1}}\PP\del{\hr = i,\, \hmuB - \mu_{i} > \eps} \\
    \stackrel{}{\le}\,& \sum_{\substack{i: i \ne 1, \\ \gam_i \le 2\gam_1}} \exp\del{-\fr{1}{2}N_i\del{\eps - \gam_1 + \Delta_i}_+^2} \tagcmt{use Lemma~\ref{lemma101.b.2:haver}} \\
    \stackrel{}{\le}\,& \sum_{\substack{i: i \ne 1, \\ \gam_i \le 2\gam_1}} \exp\del{-\fr{1}{2}N_i\del{\eps - \gam_1}_+^2} \\
    \stackrel{}{\le}\,& K \exp\del{-\fr{1}{2}\min_{\substack{i: i \ne 1, \\ \gam_i \le 2\gam_1}}\del{\eps - \gam_1}_+^2} \\
    \stackrel{}{=}\,& K \exp\del{-\fr{1}{2}N_u\del{\eps - \gam_1}_+^2}. 
  \end{align*}
  Let $q_1 = q_2 = 1$, $z_1 = \fr{N_u}{16}$, and $\eps_0 = 2\gam_1 \vee \sqrt{\fr{16\log(K)}{N_u}}$.
  We claim that in the regime of $\eps \ge \eps_0$, we have
  \begin{align*}
    h(\eps) \le q_1\exp\del{-z_1\eps^2}
  \end{align*}
  and in the regime of $\eps < \eps_0$, we have $h(\eps) \le q_2$.

  The second claim is trivial. We prove the first claim as follows. In the regime of $\eps \ge \eps_0$, we have 
  \begin{align*}
    &h(\eps) \\
    \stackrel{}{=}\,& \exp\del{-\fr{1}{2}N_u\del{\eps - \gam_1}_+^2} \\
    \stackrel{}{\le}\,& \exp\del{-\fr{1}{2}N_u\del{\fr{\eps}{2}}^2} \tagcmt{use $\eps \ge \eps_0 = 2\gam_1$}\\
    \stackrel{}{=}\,& \exp\del{-\fr{1}{8}N_u\eps^2} \\
    \stackrel{}{=}\,& \exp\del{-\fr{1}{8}N_u\eps^2 + \log(K)} \\
    \stackrel{}{\le}\,& \exp\del{-\fr{1}{16}N_u\eps^2} \tagcmt{use $\eps \ge \eps_0 \ge \sqrt{\fr{16\log(K)}{N_u}}$}. 
  \end{align*}
  
  With the above claim, we use Lemma~\ref{lemma401:utility_integral_with_eps} to bound the integral
  \begin{align*}
    &\int_{0}^{\infty}h(\eps) \difeps \\
    \stackrel{}{\le}\,& q_2\eps_0^2 + q_1\fr{1}{z_1} \\
    \stackrel{}{=}\,& \del{2\gam_1 \vee \sqrt{\fr{16\log(K)}{N_u}}}^2 + \fr{16}{N_u} \\
    \stackrel{}{\le}\,& 4\gam_1^2 + \fr{16\log(K)}{N_u} + \fr{16}{N_u}  \\
    \stackrel{}{=}\,& 4\fr{18}{N_1}\log\del{\del{\fr{KS}{N_1}}^{2\lam}} + \fr{16\log(K)}{N_u} + \fr{16}{N_u} \tagcmt{use $\gam_i = \sqrt{\fr{18}{N_i}\log\del{\del{\fr{KS}{N_i}}^{2\lam}}}$} \\
    \stackrel{}{=}\,& \fr{72}{N_1}\log\del{\del{\fr{KS}{N_1}}^{4}} + \fr{16\log(K)}{N_u} + \fr{16}{N_u} \tagcmt{recall $\lam = 2$} \\
    \stackrel{}{=}\,& \fr{72}{N_1}\log\del{\del{\fr{KS}{N_1}}^{4}} + \fr{8\log(K^2)}{N_u} + \fr{16}{N_u} \\
    \stackrel{}{\le}\,& \fr{72}{N_1}\log\del{\del{\fr{KS}{N_1}}^{4}} + \fr{24\log(K^2)}{N_u} \tagcmt{use $1 \le \log(K^2)$ with $K \ge 2$}\\
    \stackrel{}{=}\,& \fr{72}{N_1}\log\del{\del{\fr{KS}{N_1}}^{4}} + \fr{48\log(K)}{N_u}. 
  \end{align*}

  Lemma~\ref{lemma:maxlcb_101} states that for any $i \ne 1$ the condition $\gam_i < 2\gam_1$ implies that $N_i > \fr{N_1}{4} \fr{\log\del{K}}{\log\del{\fr{KS}{N_1}}}$. Thus, $N_u \ge \fr{N_1}{4} \fr{\log\del{K}}{\log\del{\fr{KS}{N_1}}}$.
  \begin{align*}
    & \fr{72}{N_1}\log\del{\del{\fr{KS}{N_1}}^{4}} + \fr{48\log(K)}{N_u} \\
    \stackrel{}{\le}\,& \fr{72}{N_1}\log\del{\del{\fr{KS}{N_1}}^{4}} + \fr{48\log(K)}{N_1}\fr{4\log\del{\fr{KS}{N_1}}}{\log\del{K}} \\
    \stackrel{}{=}\,& \fr{72}{N_1}\log\del{\del{\fr{KS}{N_1}}^{4}} + \fr{192}{N_1}\log\del{\fr{KS}{N_1}} \\
    \stackrel{}{=}\,& \fr{480}{N_1}\log\del{\fr{KS}{N_1}}. 
  \end{align*}

  For the third term, we focus on arm $i \in [K]$ such that $i \ne 1,\, \gam_i > 2\gam_1$. From Lemma~\ref{lemma101.b.3:haver}, we have 
  \begin{align*}
    &\PP\del{\hr = i, \hmuB - \mu_{i} > \eps} \\
    \stackrel{}{\le}\,& \exp\del{-\fr{1}{2}N_i\del{\eps - \gam_i + \Delta_i}_+^2}. 
  \end{align*}
  Lemma~\ref{lemma109:haver} states that
  \begin{align*}
    \PP\del{\hr = i,\, \hmu_{\hr} - \gam_{\hr} > \hmu_1 - \gam_1 } \le \del{\fr{N_i}{KS}}^{2\lam}. 
  \end{align*}
  Thus, we have
  \begin{align*}
    &\PP\del{\hr = i, \hmuB - \mu_{i} > \eps} \\
    \stackrel{}{\le}\,& \exp\del{-\fr{1}{2}N_i\del{\eps - \gam_i + \Delta_i}_+^2} \wedge \del{\fr{N_i}{KS}}^{2\lam}. 
  \end{align*}
  Let $q_1 = \del{\fr{N_i}{KS}}^{\lam}$, $q_2 = \del{\fr{N_i}{KS}}^{2\lam}$, $z_1 = \fr{N_i}{16}$, and $\eps_0 = 2\gam_i$
  We claim that in the regime of $\eps \ge \eps_0$, we have
  \begin{align*}
    \PP\del{\hr = i, \hmuB - \mu_{i} > \eps} \le q_1\exp\del{-z_1\eps^2}
  \end{align*}
  and in the regime of $\eps < \eps_0$, we have $\PP\del{\hr = i, \hmuB - \mu_{i} > \eps} \le q_2$.

  We prove the claim as follows. In the regime of $\eps \ge \eps_0$, we have 
  \begin{align*}
    &\PP\del{\hr = i, \hmuB - \mu_{i} > \eps} \\
    \stackrel{}{\le}\,& \exp\del{-\fr{1}{2}N_i\del{\eps - \gam_i}_+^2} \wedge \del{\fr{N_i}{KS}}^{2\lam} \\
    \stackrel{}{\le}\,& \sqrt{\exp\del{-\fr{1}{2}N_i\del{\eps - \gam_i}_+^2} \cdot \del{\fr{N_i}{KS}}^{2\lam}} \\
    \stackrel{}{\le}\,& \del{\fr{N_i}{KS}}^{\lam}\exp\del{-\fr{1}{4}N_i\del{\eps - \gam_i}_+^2} \\
    \stackrel{}{\le}\,& \del{\fr{N_i}{KS}}^{\lam}\exp\del{-\fr{1}{8}N_i\eps^2} \tagcmt{use $\eps \ge \eps_0 = 2\gam_i$}.
  \end{align*}

  In the regime of $\eps < \eps_0$, we have $\PP\del{\hr = i, \hmuB - \mu_{i} > \eps} \le \del{\fr{N_i}{KS}}^{2\lam}$.
 
  With the above claim, we use Lemma~\ref{lemma401:utility_integral_with_eps} to bound the integral
  \begin{align*}
    &\int_{0}^{\infty} \PP\del{\hr = i,\, \hmuB - \mu_1 > \eps, G} \difeps \\
    \stackrel{}{\le}\,& q_2\eps_0^2 + q_1\fr{1}{z_1} \\
    \stackrel{}{\le}\,& \del{\fr{N_i}{KS}}^{\lam}4\gam_i^2 + \del{\fr{N_i}{KS}}^{2\lam}\fr{16}{N_i} \\
    \stackrel{}{\le}\,& \del{\fr{N_i}{KS}}^{\lam}\fr{18}{N_i}\log\del{\del{\fr{KS}{N_i}}^{2\lam}} + \del{\fr{N_i}{KS}}^{2\lam}\fr{16}{N_i} \\
    \stackrel{}{=}\,& \del{\fr{N_i}{K\Nmax\sum_{j \in [K]}N_j}}^{\lam}\fr{18}{N_i}\log\del{\del{\fr{KS}{N_i}}^{2\lam}} + \del{\fr{N_i}{K\Nmax\sum_{j \in [K]}N_j}}^{\lam}\fr{16}{N_i} \tagcmt{use $S = \Nmax\sum_{j \in [K]}N_j$} \\
    \stackrel{}{\le}\,& \fr{18}{N_i}\del{\fr{1}{K\sum_{j \in [K]}N_j}}^{\lam}\log\del{\del{\fr{KS}{N_i}}^{2\lam}} + \fr{16}{N_i}\del{\fr{1}{K\sum_{j \in [K]}N_j}}^{\lam} \tagcmt{use $N_i \le \Nmax$} \\
    \stackrel{}{=}\,& \fr{18}{N_i}\del{\fr{1}{K\sum_{j \in [K]}N_j}}^{2}\log\del{\del{\fr{KS}{N_i}}^{4}} + \fr{16}{N_i}\del{\fr{1}{K\sum_{j \in [K]}N_j}}^{2} \tagcmt{recall $\lam = 2$} \\
    \stackrel{}{=}\,& \fr{18}{K\del{\sum_{j \in [K]}N_j}^2}\log\del{\del{KS}^{4}} + \fr{16}{K\del{\sum_{j \in [K]}N_j}^2} \tagcmt{use $N_i \ge 1$} \\
    \stackrel{}{\le}\,& \fr{34}{K\del{\sum_{j \in [K]}N_j}^2}\log\del{\del{KS}^{4}} \tagcmt{use $1 \le \log(\del{KS}^4)$ with $K \ge 2$} \\
    \stackrel{}{=}\,& \fr{136}{K\del{\sum_{j \in [K]}N_j}^2}\log\del{\del{KS}}.
  \end{align*}

  Thus, for the third term, we obtain the integral
  \begin{align*}
    &\sum_{\substack{i: i \ne 1, \\ \gam_i > 2\gam_1}} \int_{0}^{\infty} \PP\del{\hr = i,\, \hmuB - \mu_1 > \eps, G} \difeps \\
    \stackrel{}{\le}\,& \sum_{\substack{i: i \ne 1, \\ \gam_i > 2\gam_1}} \fr{136}{K\del{\sum_{j \in [K]}N_j}^2}\log\del{\del{KS}} \\
    \stackrel{}{\le}\,& \fr{136}{\del{\sum_{j \in [K]}N_j}^2}\log\del{\del{KS}}. 
  \end{align*}

  Combining all the terms, we have
  \begin{align*}
    &\int_{0}^{\infty} \PP\del{ \hmuB - \mu_1 > \eps, G } \difeps \\
    \stackrel{}{\le}\,& \int_{0}^{\infty} \PP\del{\hr =  1,\, \hmuB - \mu_1 > \eps, G } \difeps \\
    &+ \int_{0}^{\infty} \sum_{\substack{i: i \ne 1, \\ \gam_i \le 2\gam_1}}\PP\del{\hr = i,\, \hmuB - \mu_{i} > \eps, G} \difeps \\
    &+ \int_{0}^{\infty} \sum_{\substack{i: i \ne 1, \\ \gam_i > 2\gam_1}}\PP\del{\hr = i,\, \hmuB - \mu_{i} > \eps, G} \difeps \\
    \stackrel{}{\le}\,& \fr{104}{N_1}\log\del{\fr{KS}{N_1}} \\
    &+ \fr{480}{N_1}\log\del{\fr{KS}{N_1}}  \\
    &+ \fr{136}{\del{\sum_{j \in [K]}N_j}^2}\log\del{\del{KS}} \\
    \stackrel{}{\le}\,& \fr{720}{N_1}\log\del{KS}.
  \end{align*}

\end{proof}


\begin{lemma}\label{lemma_g2:haver21count2}
  In the event of $G_2 = \cbr{\exists i \in \cBs(\hr)\, \textrm{s.t.}\,  i \not\in \cB}$, HAVER achieves
  \begin{align*}
    \PP\del{G_2} \le \fr{2K^2}{K^{2\lam}}.
  \end{align*}
\end{lemma}

\begin{proof}
  Let $i$ be an arm that satisfies the event $G_2$, i.e., $i \in \cBs(\hr),\, i \not\in \cB$.

  By the definition of $\cB$, $i \not\in \cB$ means either (1) $\del{\hmu_i < \hmu_{\hr} - \gam_{\hr},\, \gam_i \le \fgam \gam_{\hr}}$ or (2) $(\gam_i > f \gam_{\hr})$.

  The condition (1) implies that
  \begin{align*}
    &\hmu_i < \hmu_{\hr} - \gam_{\hr} \\
    \stackrel{}{\Leftrightarrow}\,& \hmu_i - \mu_i - \hmu_{\hr} + \mu_{\hr} \le -\gam_{\hr} - \mu_i + \mu_{\hr} \\
    \stackrel{}{\Rightarrow}\,& \hmu_i - \mu_i - \hmu_{\hr} + \mu_{\hr} \le -\gam_{\hr} + \mu_{\hr} - \mu_s + \cstar\gam_s \tagcmt{use $i \in \cBs(\hr),\, \mu_i \ge \mu_s - \cstar\gam_s$} \\
    \stackrel{}{\Rightarrow}\,& \hmu_i - \mu_i - \hmu_{\hr} + \mu_{\hr} \le -\gam_{\hr}  + \cstar\gam_s - \cs\gam_s + \cs\gam_{\hr} \tagcmt{by def of $s$, $\mu_s - \cs\gam_s \ge \mu_{\hr} - \cs\gam_{\hr}$}\\
    \stackrel{}{\Leftrightarrow}\,& \hmu_i - \mu_i - \hmu_{\hr} + \mu_{\hr} \le -\fr{5}{6}\gam_{\hr}. 
  \end{align*}
  Thus, we have
  \begin{align*}
    & \PP\del{ \exists i \in \cBs(\hr)\, \textrm{s.t.}\,  i \not\in \cB,\, \hmu_{i} - \mu_{i} + \mu_{\hr} - \hmu_{\hr} < -\fr{5}{6}\gam_{\hr} } \\
    \stackrel{}{\le}\,& \sum_{\substack{r \in [K]}} \sum_{i \in \cBs(r)} \PP\del{ \hmu_{i} - \mu_{i} + \mu_{r} - \hmu_{r} < -\fr{5}{6}\gam_{r} } \\
    \stackrel{}{\le}\,& \sum_{\substack{r \in [K]}} \sum_{i \in \cBs(r)} \del{\PP\del{ \mu_{r} - \hmu_{r} < -\fr{1}{3}\gam_{r} } + \PP\del{ \hmu_{i} - \mu_{i}  < -\fr{1}{2}\gam_{r} }} \\
    \stackrel{}{\le}\,& \sum_{\substack{r \in [K]}} \sum_{i \in \cBs(r)} \del{ \exp\del{\fr{-N_{r}\gam_{r}^2}{18}} + \exp\del{\fr{-N_{i}\gam_{r}^2}{8}}} \\
    \stackrel{}{\le}\,& \sum_{\substack{r \in [K]}} \sum_{i \in \cBs(r)} \del{ \exp\del{\fr{-N_{r}\gam_{r}^2}{18}} + \exp\del{\fr{- N_{i}\del{\fgaminv}^{2} \gam_{i}^2}{8}}} \tagcmt{use $i \in \cBs(r),\, \gam_i \le \fgam\gam_{r}$} \\
    \stackrel{}{=}\,& \sum_{\substack{r \in [K]}} \sum_{i \in \cBs(r)} \del{ \exp\del{\fr{-N_{r}\gam_{r}^2}{18}} + \exp\del{\fr{-N_{i}\gam_{i}^2}{18}}} \\
    \stackrel{}{\le}\,& \sum_{\substack{r \in [K]}} \sum_{i \in \cBs(r)}\del{\del{\fr{N_r}{KS}}^{2\lam} + \del{\fr{N_s}{KS}}^{2\lam}} \tagcmt{by def $\forall i \in [K],\, \gam_{i} = \sqrt{\fr{18}{N_{i}}\log\del{\del{\fr{KS}{N_i}}^{2\lam}}}$} \\
    \stackrel{}{\le}\,& \sum_{\substack{r \in [K]}} \sum_{i \in \cBs(r)} \fr{2}{K^{2\lam}} \tagcmt{use $\forall i,\, N_i \le S$} \\
    \stackrel{}{\le}\,& \fr{2K^2}{K^{2\lam}}. 
  \end{align*}
  
  The condition (2) contradicts with $i \in \cBs(\hr)$ (i.e, $\gam_i \le \fgam \gam_{\hr}$), thus, 
  \begin{align*}
    \PP\del{\exists i \in \cBs(\hr)\, \textrm{s.t.}\,  i \not\in \cB,\, \gam_i > f \gam_{\hr}} = 0.
  \end{align*}
  Therefore,
  \begin{align*}
    &\PP\del{G_2} \\
    \stackrel{}{\le}\,& \PP\del{ \exists i \in \cBs(\hr)\, \textrm{s.t.}\,  i \not\in \cB,\, \hmu_i < \hmu_{\hr} - \gam_{\hr},\, \gam_i \le \fgam \gam_{\hr} } + \PP\del{\exists i \in \cBs(\hr)\, \textrm{s.t.}\,  i \not\in \cB,\, \gam_i > \fgam \gam_{\hr}} \\
    \stackrel{}{\le}\,& \fr{2K^2}{K^{2\lam}}.
  \end{align*}
  
\end{proof}

\begin{lemma}\label{lemma_g3:haver21count2}
  In the event of $G_3 = \cbr{\cBs(\hr) \subseteq \cB,\, \exists i \not\in \cBp(\hr)\, \textrm{s.t.}\, i \in \cB}$, HAVER achieves
  \begin{align*}
    \PP\del{G_3} \le \fr{2K^2}{K^{2\lam}}.
  \end{align*}
\end{lemma}

\begin{proof}
  Let $i$ be an arm that satisfies the event $G_3$, i.e., $i \not\in \cBp(r),\, i \in \cB$.

  By the definition of $\cBp(r)$, $i \not\in \cBp(r)$ means either (1) ($\mu_i < \mu_s - \cplus\gam_s - \cplus\gam_i,\, \gam_i \le \fgam \gam_{\hr}$) or (2) ($\gam_i > \fgam \gam_{\hr}$).
  
  By the definition of $\cB$, since $i \in \cB$, we have $\hmu_i \ge \hmu_{\hr} - \gam_{\hr}$. With the condition (1), we have 
  \begin{align*}
    &\hmu_i \ge \hmu_{\hr} - \gam_{\hr} \\
    \stackrel{}{\Leftrightarrow}\,&  \hmu_i - \hmu_{\hr} \ge - \gam_{\hr} \\
    \stackrel{}{\Rightarrow}\,& \hmu_i - \hmu_s + \gam_{s} - \gam_{\hr} \ge -\gam_{\hr} \tagcmt{by def of $\hr$, $\hmu_{\hr} \ge \hmu_{s} - \gam_{s} + \gam_{\hr}$} \\
    \stackrel{}{\Leftrightarrow}\,& \hmu_i - \hmu_s \ge -\gam_s \\
    \stackrel{}{\Leftrightarrow}\,& \hmu_i - \mu_i + \mu_s - \hmu_s \ge -\gam_s + \mu_s - \mu_i \\
    \stackrel{}{\Rightarrow}\,& \hmu_i - \mu_i + \mu_s - \hmu_s \ge -\gam_s + \cplus\gam_s + \cplus\gam_i \tagcmt{use condition (1)} \\
    \stackrel{}{\Leftrightarrow}\,& \hmu_i - \mu_i + \mu_s - \hmu_s \ge \fr{1}{3}\gam_s + \fr{4}{3}\gam_i. 
  \end{align*}
  Thus, we have
  \begin{align*}
    &\PP\del{G_3} \\
    \stackrel{}{=}\,& \PP\del{\exists i \not\in \cBp,\, i \in \cB,\, \hmu_{i} - \hmu_{\hr} \ge -\gam_{\hr} } \\
    \stackrel{}{\le}\,& \PP\del{\exists i \not\in \cBp,\, i \in \cB,\, \hmu_i - \mu_i + \mu_s - \hmu_s \ge \fr{1}{3}\gam_s + \fr{4}{3}\gam_i} \\
    \stackrel{}{\le}\,& \sum_{i \not\in \cBp} \PP\del{\hmu_i - \mu_i + \mu_s - \hmu_s \ge \fr{1}{3}\gam_s + \fr{4}{3}\gam_i} \\
    \stackrel{}{\le}\,& \sum_{i \not\in \cBp} \del{\PP\del{\mu_s - \hmu_s \ge \fr{1}{3}\gam_s} + \PP\del{\hmu_i - \mu_i \ge \fr{4}{3}\gam_i}} \\
    \stackrel{}{\le}\,& \sum_{i \not\in \cBp}\del{\exp\del{-\fr{N_{s}\gam_{s}^2}{18}} + \exp\del{-\fr{16N_{i}\gam_{i}^2}{18}}} \\
    \stackrel{}{\le}\,& \sum_{i \not\in \cBp}\del{\exp\del{-\fr{N_{s}\gam_{s}^2}{18}} + \exp\del{-\fr{N_{i}\gam_{i}^2}{18}}} \\
    \stackrel{}{=}\,& \sum_{i \not\in \cBp}\del{\del{\fr{N_s}{KS}}^{2\lam} + \del{\fr{N_i}{KS}}^{2\lam} } \tagcmt{by def $\forall i \in [K],\, \gam_{i} = \sqrt{\fr{18}{N_{i}}\log\del{\del{\fr{KS}{N_i}}^{2\lam}}}$} \\
    \stackrel{}{\le}\,& \sum_{i \not\in \cBp} \fr{2}{K^{2\lam}} \tagcmt{use $\forall i,\, N_i \le S$} \\
    \stackrel{}{\le}\,&\fr{2K^2}{K^{2\lam}}.  
  \end{align*}
  The condition (2) contradicts with $i \in \cB$ (i.e., $\gam_i \le \fgam \gam_{\hr}$), thus,
  \begin{align*}
    \PP\del{\exists i \not\in \cBp,\, i \in \cB,\, \gam_i > \fgam \gam_{\hr}} = 0.
  \end{align*}
  Therefore,
  \begin{align*}
    &\PP\del{G_3} \\
    \stackrel{}{\le}\,& \PP\del{\exists i \not\in \cBp,\, i \in \cB,\, \mu_i < \mu_s - \cplus\gam_s - \cplus\gam_i,\, \gam_i \le \fgam \gam_{\hr}} + \PP\del{\exists i \not\in \cBp,\, i \in \cB,\, \gam_i > \fgam \gam_{\hr}} \\
    \stackrel{}{\le}\,& \fr{2K^2}{K^{2\lam}}.
  \end{align*}
\end{proof}

\begin{lemma}\label{lemma109:haver}
  Let any arm $i$ such that $i \ne 1,\, \gam_i > 2\gam_1$, HAVER achieves
  \begin{align*}
    \PP\del{\hr = i,\, \hmu_{\hr} - \gam_{\hr} > \hmu_1 - \gam_1 } \le \del{\fr{N_i}{KS}}^{2\lam}.
  \end{align*}
\end{lemma}

\begin{proof}
  By the definition of $\hr$, we have $\hmu_{\hr} - \gam_{\hr} \ge \hmu_1 - \gam_1$. Therefore, with $\hr = i$, we have
  \begin{align*}
    &\PP\del{\hr = i,\, \hmu_{\hr} - \gam_{\hr} \ge \hmu_1 - \gam_1} \\
    \stackrel{}{\le}\,& \PP\del{\hr = i,\, \hmu_i - \mu_i - \hmu_1 + \mu_1 \ge \gam_i - \gam_1 + \Delta_i} \\
    \stackrel{}{\le}\,& \PP\del{\hr = i,\, \hmu_i - \mu_i - \hmu_1 + \mu_1 \ge \fr{1}{2}\gam_i + \Delta_i} \tagcmt{use $\gam_i > 2\gam_1$} \\
    \stackrel{}{\le}\,& \exp\del{-\fr{1}{2}\fr{1}{\del{\fr{1}{N_i} + \fr{1}{N_1}}}\del{\fr{1}{2}\gam_i + \Delta_i}^2} \\
    \stackrel{}{\le}\,& \exp\del{-\fr{1}{8}\fr{1}{\del{\fr{1}{N_i} + \fr{1}{N_1}}}\del{\gam_i + \Delta_i}^2}.
  \end{align*}

  Lemma~\ref{lemma:maxlcb_102} states that for any $i \ne 1$ the condition $\gam_i > 2\gam_1$ implies that $N_i < N_1$. Therefore,
  \begin{align*}
    &\exp\del{-\fr{1}{8}\fr{1}{\del{\fr{1}{N_i} + \fr{1}{N_1}}}\del{\gam_i + \Delta_i}^2} \\
    \stackrel{}{\le}\,& \exp\del{-\fr{1}{8}\fr{1}{\del{\fr{1}{N_i} + \fr{1}{N_i}}}\del{\gam_i + \Delta_i}^2} \\
    \stackrel{}{=}\,& \exp\del{-\fr{1}{16}N_i\del{\gam_i + \Delta_i}^2} \\
    \stackrel{}{\le}\,& \exp\del{-\fr{1}{16}N_i\gam_i^2} \tagcmt{use $\Delta_i \ge 1$} \\
    \stackrel{}{\le}\,& \exp\del{-\fr{1}{18}N_i\gam_i^2} \\
    \stackrel{}{=}\,& \del{\fr{N_i}{KS}}^{2\lam} \tagcmt{use $\gam_i = \sqrt{\fr{18}{N_i}\log\del{\del{\fr{KS}{N_i}}^{2\lam}}}$}.
  \end{align*}
  
\end{proof}


\begin{lemma}\label{lemma401:utility_integral_with_eps}
  Let $q_1$, $q_2$, $z_1$, and $\eps_0$ be some positive real values. Let $f$ be a function of $\eps$ such that in the regime of $\eps \ge \eps_0$, we have 
  \begin{align*}
    f(\eps) \le q_1\exp\del{-z_1\eps^2}.
  \end{align*}
  and in the regime of $\eps < \eps_0$, we have $f(\eps) \le q_2$.
  Then, we can bound the integral
  \begin{align*}
    \int_{0}^{\infty} f(\eps) \difeps \le q_2\eps_0^2 + q_1\fr{1}{z_1}.
  \end{align*}
\end{lemma}

\begin{proof}
  We bound the integral as follows
  \begin{align*}
    &\int_{0}^{\infty} f(\eps) \difeps \\
    \stackrel{}{=}\,& \int_{0}^{\eps_0} f(\eps) \difeps + \int_{\eps_0}^{\infty} f(\eps)  \difeps \\
    \stackrel{}{\le}\,& \int_{0}^{\eps_0} f(\eps) \difeps + \int_{\eps_0}^{\infty} q_1\exp\del{-z_1\eps^2} \difeps \tagcmt{in the regime $\eps \ge \eps_0$, use $f(\eps) \le q_1\exp\del{-z_1\eps^2}$} \\
    \stackrel{}{\le}\,& \int_{0}^{\eps_0} q_2\difeps + \int_{\eps_0}^{\infty} q_1\exp\del{-z_1\eps^2} \difeps \tagcmt{in the regime $\eps \le \eps_0$, use $f(\eps) \le q_2$} \\
    \stackrel{}{=}\,& q_2\eps^2 \Big|_{0}^{\eps_0} - \fr{q_1}{z_1}\exp\del{-z_1\eps^2} \Big|_{\eps_0}^{\infty} \\
    \stackrel{}{=}\,& q_2\eps_0^2 + q_1\fr{1}{z_1}\exp\del{-z_1\eps_0^2} \\
    \stackrel{}{\le}\,& q_2\eps_0^2 + q_1\fr{1}{z_1} \tagcmt{use $\exp\del{-z_1\eps_0^2} \le 1$}.
  \end{align*}
  
\end{proof}

\begin{lemma}\label{lemma401:utility_integral_without_eps}
  Let $q_1$ and $z_1$ be some positive real values. Let $f$ be a function of $\eps$. If we have 
  \begin{align*}
    f(\eps) \le q_1\exp\del{-z_1\eps^2}.
  \end{align*}
  Then, we can bound the integral
  \begin{align*}
    \int_{0}^{\infty} f(\eps) \difeps \le q_1\fr{1}{z_1}.
  \end{align*}
\end{lemma}

\begin{proof}
  We bound the integral as follows
  \begin{align*}
    &\int_{0}^{\infty} f(\eps) \difeps \\
    \stackrel{}{\le}\,& \int_{0}^{\infty} q_1\exp\del{-z_1\eps^2} \difeps \\
    \stackrel{}{=}\,& - q_1\fr{1}{z_1}\exp\del{-z_1\eps^2} \Big|_{0}^{\infty} \\
    \stackrel{}{=}\,& q_1\fr{1}{z_1}. 
  \end{align*}
  
\end{proof}

\begin{lemma}\label{lemma200.a:haver21count2}
  With $q^2 \ge \fr{1}{9}$, in HAVER, we have
  \begin{align*}
    \int_{0}^{\infty} \PP\del{ \abs{\hmuB - \mu_1} > \eps,\, \exists i \in [K],\, \abs{\hmu_i - \mu_i} \ge q\gam_i} \difeps \le \tilcO \del{\fr{1}{KN_1}}.
  \end{align*}
\end{lemma}

\begin{proof}
  We denote $G = \cbr{\exists i \in [K],\, \abs{\hmu_{i} - \mu_i} \ge q \gam_i}$. We have
  \begin{align*}
    &\PP\del{\exists i \in [K],\, \abs{\hmu_{i} - \mu_i} \ge q \gam_i} \\
    \stackrel{}{\le}\,& \sum_{i \in [K]}\PP\del{\abs{\hmu_{i} - \mu_i} \ge q \gam_i} \\
    \stackrel{}{\le}\,& \sum_{i \in [K]}2\exp\del{-\fr{1}{2}N_iq^2\gam_i^2} \\
    \stackrel{}{=}\,& \sum_{i \in [K]}2\exp\del{-\fr{1}{2}N_i q^2 \fr{18}{N_i}\ln\del{\del{\fr{KS}{N_i}}^{2\lam}}} \tagcmt{use $\gam_i = \sqrt{\fr{18}{N_i}\ln\del{\del{\fr{KS}{N_i}}^{2\lam}}}$} \\
    \stackrel{}{\le}\,& \sum_{i \in [K]}2\del{\fr{N_i}{KS}}^{2\lam} \tag{use $q^2 \ge \fr{1}{9} $} \\
    \stackrel{}{\le}\,& \sum_{i \in [K]}2\del{\fr{1}{K}}^{2\lam} \tagcmt{use $\forall i \in [K],\, N_i \le S$} \\
    \stackrel{}{\le}\,& \fr{2K^2}{K^{2\lam}} \tagcmt{reminder we set $\lam = 2$}.
  \end{align*}

  Since $G$ satisfies $\PP\del{G} \le \fr{2K^2}{K^{2\lam}}$, we use Lemma~\ref{lemma_g2a:haver21count2} to obtain
  \begin{align*}
    &\int_{0}^{\infty} \PP\del{ \hmuB - \mu_1 < -\eps,\, G } \difeps \\
    \stackrel{}{\le}\,& \min_m \del{\fr{704}{K^2N_m}\log\del{\fr{KS}{N_m}}  + \fr{8}{K^{2}}\Delta_m^2} \\
    \stackrel{}{\le}\,& \fr{704}{K^2N_1}\log\del{\fr{KS}{N_1}} \tagcmt{upper bounded by $m = 1$}.
  \end{align*}

  Also, we use Lemma~\ref{lemma_g2b:haver21count2} to obtain
  \begin{align*}
    \int_{0}^{\infty} \PP\del{ \hmuB - \mu_1 > \eps,\, G } \difeps \le \fr{3088}{KN_1}\log\del{KS}.
  \end{align*}
  Therefore,
  \begin{align*}
    &\int_{0}^{\infty} \PP\del{ \abs{\hmuB - \mu_1} < \eps,\, G } \difeps \\
    \stackrel{}{=}\,& \PP\del{ \hmuB - \mu_1 < -\eps,\, G } \difeps \\
    &+ \int_{0}^{\infty} \PP\del{ \hmuB - \mu_1 > \eps,\, G } \difeps \\
    \stackrel{}{\le}\,& \fr{704}{K^2N_1}\log\del{\fr{KS}{N_1}} + \fr{3088}{KN_1}\log\del{KS} \\
    \stackrel{}{\le}\,& \fr{3792}{KN_1}\log\del{KS}.
  \end{align*}
    
\end{proof}

\begin{lemma}\label{xlemma:max_bound}
  Let $n,d \ge 1$ be two integers. Then,
  \begin{align*}
    \max_{k=0}^{d} \fr{k^2}{\del{n+k}^2} \le \del{\log\del{\fr{n+d}{n}}}^2.
  \end{align*}
\end{lemma}

\begin{proof}
  We have
  \begin{align*}
    &\max_{k=0}^{d} \fr{k^2}{\del{n+k}^2} \\
    \stackrel{}{\le}\,& \del{\max_{k=0}^{d} \fr{k}{\del{n+k}}}^2 \\
    \stackrel{}{\le}\,& \del{\sum_{k=0}^d \fr{1}{\del{n+k}}}^2 \\
  \end{align*}
  Then, we bound the summation term with integral
  \begin{align*}
    &\sum_{k=0}^d \fr{1}{\del{n+k}} \\
    \stackrel{}{\le}\,& \int_{k=0}^{d} \fr{1}{\del{n+k}} \\
    \stackrel{}{=}\,& \log\del{n+k} \Big|_{0}^{d} \\
    \stackrel{}{=}\,& \log\del{\fr{n+d}{n}}.
  \end{align*}
  Therefore,
  \begin{align*}
    \max_{k=0}^{d} \fr{k^2}{\del{n+k}^2} \le \del{\log\del{\fr{n+d}{n}}}^2.
  \end{align*}
\end{proof}

\begin{lemma}\label{lemma305:bcals}
  A sufficient condition for
  \begin{align*}
    K\del{c_1\sqrt{\fr{1}{N}}}^{\fr{1}{\alpha}} - 1 \ge K\del{\fr{c_1}{2}\sqrt{\fr{1}{N}}}^{\fr{1}{\alpha}}
  \end{align*}
  is
  \begin{align*}
    N \le \fr{c_1^2}{4}\del{\fr{K}{\alpha}\log(2)}^{2\alpha}. 
  \end{align*}
\end{lemma}

\begin{proof}
  It suffices to prove the contraposition.
  Thus, we assume $K\del{c_1\sqrt{\fr{1}{N}}}^{\fr{1}{\alpha}} - 1 < K\del{\fr{c_1}{2}\sqrt{\fr{1}{N}}}^{\fr{1}{\alpha}}$ is true.
  Then,
  \begin{align*}
    &K\del{c_1\sqrt{\fr{1}{N}}}^{\fr{1}{\alpha}} - 1 < K\del{\fr{c_1}{2}\sqrt{\fr{1}{N}}}^{\fr{1}{\alpha}} \\
    \stackrel{}{\Leftrightarrow}\,& 1 > K\del{c_1\sqrt{\fr{1}{N}}}^{\fr{1}{\alpha}}\del{1 - \del{\fr{1}{2}}^{\fr{1}{\alpha}}} \\
    \stackrel{}{\Leftrightarrow}\,& 1 > \del{c_1\sqrt{\fr{1}{N}}}\del{K\del{1 - \del{\fr{1}{2}}^{\fr{1}{\alpha}}}}^{\alpha} \\
    \stackrel{}{\Leftrightarrow}\,& N > c_1^2\del{K\del{1 - \del{\fr{1}{2}}^{\fr{1}{\alpha}}}}^{2\alpha} \\
    \stackrel{}{\Leftrightarrow}\,& N > c_1^2\del{K\del{\fr{2^{\fr{1}{\alpha}}-1}{2^{\fr{1}{\alpha}}}}}^{2\alpha} \\
    \stackrel{}{\Leftrightarrow}\,& N > \fr{c_1^2}{4}\del{K\del{2^{\fr{1}{\alpha}}-1}}^{2\alpha} \\
    \stackrel{}{\Rightarrow}\,& N > \fr{1}{4c_1^2}\del{K\del{\fr{\log(2)}{\alpha}}}^{2\alpha} \tagcmt{use $\forall x,\, \fr{\log(x)}{\alpha} \le x^{\fr{1}{\alpha}}-1$ choose $x = 2$} \\
    \stackrel{}{\Leftrightarrow}\,& N > \fr{c_1^2}{4}\del{\fr{K}{\alpha}\log(2)}^{2\alpha}.
  \end{align*}
  Therefore, the sufficient condition for
  \begin{align*}
    K\del{\fr{1}{c_1}\sqrt{\fr{1}{N}}}^{\fr{1}{\alpha}} - 1 \ge K\del{\fr{1}{2c_1}\sqrt{\fr{1}{N}}}^{\fr{1}{\alpha}}
  \end{align*}
  is
  \begin{align*}
    N \le \fr{c_1^2}{4}\del{\fr{K}{\alpha}\log(2)}^{2\alpha}. 
  \end{align*}
\end{proof}

\begin{lemma}\label{lemma306:bcals}
  A sufficient condition for
  \begin{align*}
    K\del{c_2\sqrt{\fr{1}{N}}}^{\fr{1}{\alpha}} + 1 \le K\del{2c_2\sqrt{\fr{1}{N}}}^{\fr{1}{\alpha}}
  \end{align*}
  is
  \begin{align*}
    N \le c_2^2\del{\fr{K}{\alpha}\log(2)}^{2\alpha}.
  \end{align*}
\end{lemma}

\begin{proof}
  It suffices to prove the contraposition.
  Thus, we assume $K\del{c_2\sqrt{\fr{1}{N}}}^{\fr{1}{\alpha}} + 1 > K\del{2c_2\sqrt{\fr{1}{N}}}^{\fr{1}{\alpha}}$ is true.
  Then,
  \begin{align*}
    &K\del{c_2\sqrt{\fr{1}{N}}}^{\fr{1}{\alpha}} + 1 > K\del{2c_2\sqrt{\fr{1}{N}}}^{\fr{1}{\alpha}} \\
    \stackrel{}{\Leftrightarrow}\,&  1 > K\del{c_2\sqrt{\fr{1}{N}}}^{\fr{1}{\alpha}}\del{2^{\fr{1}{\alpha}} - 1} \\
    \stackrel{}{\Leftrightarrow}\,&  1 > \del{c_2\sqrt{\fr{1}{N}}}\del{K\del{2^{\fr{1}{\alpha}} - 1}}^{\alpha} \\
    \stackrel{}{\Leftrightarrow}\,&  N > c_2^2\del{K\del{2^{\fr{1}{\alpha}} - 1}}^{2\alpha} \\
    \stackrel{}{\Rightarrow}\,&  N > c_2^2\del{K\del{\fr{\log(2)}{\alpha}}}^{2\alpha} \tagcmt{use $\forall x,\, \fr{\log(x)}{\alpha} \le x^{\fr{1}{\alpha}}-1$ choose $x = 2$} \\
    \stackrel{}{\Leftrightarrow}\,&  N > c_2^2\del{\fr{K}{\alpha}\log(2)}^{2\alpha}.
  \end{align*}
  Therefore, the sufficient condition for
  \begin{align*}
    K\del{c_2\sqrt{\fr{1}{N}}}^{\fr{1}{\alpha}} + 1 \le \del{2c_2\sqrt{\fr{1}{N}}}^{\fr{1}{\alpha}}
  \end{align*}
  is
  \begin{align*}
    N \le c_2^2\del{\fr{K}{\alpha}\log(2)}^{2\alpha}.
  \end{align*}
\end{proof}

\begin{lemma}\label{lemma301:bcals}
  Consider the Poly($\alpha$) instance where $\forall i \ge 2,\, \Delta_i = \del{\fr{i}{K}}^{\alpha}$. Suppose $\cS = \cbr{i: \Delta_i \le c\sqrt{\fr{1}{N}}}$. Assuming $K\del{\fr{c}{\sqrt{N}}}^{\fr{1}{\alpha}} \ge 2$, we have 
  \begin{align*}
    \abs{\cS} = \floor{K\del{\fr{c}{\sqrt{N}}}^{\fr{1}{\alpha}}}.
  \end{align*}
\end{lemma}

\begin{proof}
  
We have
\begin{align*}
  &\Delta_i \le c\sqrt{\fr{1}{N}} \\
  \stackrel{}{\Leftrightarrow}\,& \del{\fr{i}{K}}^{\alpha} \le c\sqrt{\fr{1}{N}} \\
  \stackrel{}{\Leftrightarrow}\,& \fr{i}{K} \le \fr{c^{\fr{1}{\alpha}}}{N^{\fr{1}{2\alpha}}} \\
  \stackrel{}{\Leftrightarrow}\,& i \le \fr{Kc^{\fr{1}{\alpha}}}{N^{\fr{1}{2\alpha}}} \\
  \stackrel{}{\Leftrightarrow}\,& i \le K\del{\fr{c}{\sqrt{N}}}^{\fr{1}{\alpha}}.
\end{align*}

Therefore, $\abs{\cS} = \max\cbr{i: \Delta_i \le c\sqrt{\fr{1}{N}}} = \floor{K\del{\fr{c}{\sqrt{N}}}^{\fr{1}{\alpha}}} $.

\end{proof}

\begin{lemma}\label{lemma302:bcals}
  Consider the Poly($\alpha$) instance where $\forall i \ge 2,\, \Delta_i = \del{\fr{i}{K}}^{\alpha}$. Suppose $\Bcals = \cbr{i: \Delta_i \le a_1\sqrt{\fr{1}{N}}}$ and $\Bcalp = \cbr{i: \Delta_i \le a_2\sqrt{\fr{1}{N}}}$. Assuming $N \le \del{\fr{a_1^2}{4} \wedge a_2^2}\del{\fr{K}{\alpha}\log(2)}^{2\alpha}$, we have
  \begin{align*}
     \fr{1}{\abs{\cBs}} \le \fr{1}{K\del{\fr{a_1}{2}\fr{1}{\sqrt{N}}}^{\fr{1}{\alpha}}},
  \end{align*}
  \begin{align*}
    \fr{\abs{\cBp}}{\abs{\cBs}} \le \del{\fr{2a_2}{a_1}}^{\fr{1}{\alpha}},
  \end{align*}
  and
  \begin{align*}
    \fr{1}{\abs{\cBs}}\sum_{i = 1}^{\abs{\Bcalp}}\Delta_i 
    \stackrel{}{\le}\,& \fr{2a_2\del{\fr{4a_2}{a_1}}^{\fr{1}{\alpha}}}{(\alpha+1)\sqrt{N}}.
  \end{align*}
  
\end{lemma}

\begin{proof}
  We prove the first inequality as follows. From Lemma~\ref{lemma301:bcals}, we have $\abs{\Bcals} = \floor{K\del{\fr{a_1}{\sqrt{N}}}^{\fr{1}{\alpha}}} \ge K\del{\fr{c_1}{\sqrt{N}}}^{\fr{1}{\alpha}}-1$.
  
  We have the assumption $N \le \fr{a_1^2}{4}\del{\fr{K}{\alpha}\log(2)}^{2\alpha}$. Lemma~\ref{lemma305:bcals} states that condition $N \le \fr{a_1^2}{4}\del{\fr{K}{\alpha}\log(2)}^{2\alpha}$ implies 
  \begin{align*}
    K\del{\fr{a_1}{\sqrt{N}}}^{\fr{1}{\alpha}} - 1 \ge K\del{\fr{a_1}{2}\fr{1}{\sqrt{N}}}^{\fr{1}{\alpha}}. 
  \end{align*}
  Thus, 
  \begin{align}
    \abs{\Bcals} \ge K\del{\fr{a_1}{\sqrt{N}}}^{\fr{1}{\alpha}} - 1 \ge K\del{\fr{a_1}{2}\fr{1}{\sqrt{N}}}^{\fr{1}{\alpha}} \label{eq:cBs}.
  \end{align}
  Therefore,
  \begin{align*}
    \fr{1}{\abs{\cBs}} \le \fr{1}{K\del{\fr{a_1}{2}\fr{1}{\sqrt{N}}}^{\fr{1}{\alpha}}},
  \end{align*}
  which concludes the first inequality. 
  
  We prove the second inequality as follows. From Lemma~\ref{lemma301:bcals}, we have $\abs{\Bcalp} = \floor{K\del{\fr{a_2}{\sqrt{N}}}^{\fr{1}{\alpha}}} \le K\del{\fr{a_2}{\sqrt{N}}}^{\fr{1}{\alpha}}$.
  Thus,
  \begin{align*}
    \fr{\abs{\cBp}}{\abs{\cBs}}
    \le \fr{K\del{\fr{a_2}{\sqrt{N}}}^{\fr{1}{\alpha}}}{K\del{\fr{a_1}{2}\fr{1}{\sqrt{N}}}^{\fr{1}{\alpha}}} 
    = \del{\fr{2a_2}{a_1}}^{\fr{1}{\alpha}}, 
  \end{align*}
  which concludes the second inequality.

  We prove the third inequality as follows. We have the assumption $N \le a_2^2\del{\fr{K}{\alpha}\log(2)}^{2\alpha}$. Lemma~\ref{lemma306:bcals} states that condition $N \le a_2^2\del{\fr{K}{\alpha}\log(2)}^{2\alpha}$ implies
  \begin{align*}
    K\del{\fr{a_2}{\sqrt{N}}}^{\fr{1}{\alpha}} + 1 \le K\del{\fr{2a_2}{\sqrt{N}}}^{\fr{1}{\alpha}}.
  \end{align*}
  Thus, we have
  \begin{align}
    \abs{\Bcalp} + 1 \le  K\del{\fr{a_2}{\sqrt{N}}}^{\fr{1}{\alpha}} + 1 \le K\del{\fr{2a_2}{\sqrt{N}}}^{\fr{1}{\alpha}} \label{eq:cBp}.
  \end{align}

  \begin{align*}
  \fr{\abs{\Bcalp}}{\abs{\Bcals}}
  \stackrel{}{\le}\,& \fr{K\del{a_2\sqrt{\fr{1}{N}}}^{\fr{1}{\alpha}}}{K\del{\fr{1}{a_1}\sqrt{\fr{1}{N}}}^{\fr{1}{\alpha}}-1} \\
  \stackrel{}{\le}\,& \fr{K\del{a_2\sqrt{\fr{1}{N}}}^{\fr{1}{\alpha}}}{K\del{\fr{1}{2a_1}\sqrt{\fr{1}{N}}}^{\fr{1}{\alpha}}} \\
  \stackrel{}{=}\,& \del{2a_1a_2}^{\fr{1}{\alpha}}.
  \end{align*}
  
  We have 
  \begin{align*}
    \sum_{i = 1}^{\abs{\Bcalp}}\Delta_i 
    \stackrel{}{=}\,& \sum_{i = 1}^{\abs{\Bcalp}}\del{\fr{i}{K}}^{\alpha} \\
    \stackrel{}{=}\,& \fr{1}{K^{\alpha}}\sum_{i = 1}^{\abs{\Bcalp}}i^{\alpha} \\
    \stackrel{}{\le}\,& \fr{1}{K^{\alpha}}\int_{i = 1}^{\abs{\Bcalp}+1}i^{\alpha} \\
    \stackrel{}{=}\,& \fr{1}{K^{\alpha}}\fr{\del{\abs{\Bcalp}+1}^{\alpha+1} - 1}{\alpha+1} \\
    \stackrel{}{\le}\,& \fr{1}{K^{\alpha}}\fr{\del{\abs{\Bcalp}+1}^{\alpha+1}}{\alpha+1} \\
    \stackrel{}{\le}\,& \fr{1}{K^{\alpha}}\fr{\del{K\del{\fr{2a_2}{\sqrt{N}}}^{\fr{1}{\alpha}}}^{\alpha+1}}{\alpha+1} \tagcmt{from \ref{eq:cBp}}. 
  \end{align*}

  Therefore,
  \begin{align*}
    \fr{1}{\abs{\cBs}}\sum_{i = 1}^{\abs{\Bcalp}}\Delta_i 
    \stackrel{}{\le}\,& \fr{1}{\abs{\cBs}} \fr{1}{K^{\alpha}}\fr{\del{K\del{\fr{2a_2}{\sqrt{N}}}^{\fr{1}{\alpha}}}^{\alpha+1}}{\alpha+1} \\
    \stackrel{}{\le}\,& \fr{1}{K\del{\fr{a_1}{2}\fr{1}{\sqrt{N}}}^{\fr{1}{\alpha}}} \fr{1}{K^{\alpha}}\fr{\del{K\del{\fr{2a_2}{\sqrt{N}}}^{\fr{1}{\alpha}}}^{\alpha+1}}{\alpha+1} \tagcmt{from \ref{eq:cBs}} \\
    \stackrel{}{=}\,& \fr{\del{2a_2}^{1 + \fr{1}{\alpha}}}{\del{\fr{a_1}{2}}^{\fr{1}{\alpha}}} \cdot \fr{1}{(\alpha+1)\sqrt{N}} \\
    \stackrel{}{=}\,& \fr{2a_2\del{\fr{4a_2}{a_1}}^{\fr{1}{\alpha}}}{(\alpha+1)\sqrt{N}},
  \end{align*}
  which concludes the third inequality.

\end{proof}

\begin{lemma}\label{xlemma:delta_sufficient}
  A sufficient condition for $\Delta > \sqrt{\fr{c}{N}\log\del{K^2N}}$ is
  \begin{align*}
    N > \fr{2c}{\Delta^2}\log\del{\fr{2cK^2}{\Delta^2e}}.
  \end{align*}
\end{lemma}

\begin{proof}
  It suffices to prove the contraposition. Thus, we assume $\Delta \le \sqrt{\fr{c}{N}\log\del{K^2N}}$ is true.
  Then,
  \begin{align*}
    &\Delta \le \sqrt{\fr{c}{N}\log\del{K^2N}} \\
    \stackrel{}{\Leftrightarrow}\,& N \le \fr{c}{\Delta^2}\log\del{K^2N} \\
    \stackrel{}{\Leftrightarrow}\,& N = \fr{c}{\Delta^2}\log\del{N} + \fr{c}{\Delta^2}\log\del{K^2}.
  \end{align*}
  We set $A = \fr{c}{\Delta^2}$ and $B = \fr{c}{\Delta^2}\log\del{K^2}$.
  Thus, we have
  \begin{align*}
    &\Delta \le \sqrt{\fr{c}{N}\log\del{K^2N}} \\
    \stackrel{}{\Leftrightarrow}\,& N = \fr{c}{\Delta^2}\log\del{N} + \fr{c}{\Delta^2}\log\del{K^2} \\
    \stackrel{}{\Leftrightarrow}\,& N = A\log\del{N} + B \\
    \stackrel{}{\Leftrightarrow}\,& N = A\log\del{\fr{N}{2A}\cdot 2A} + B \\
    \stackrel{}{\Leftrightarrow}\,& N = A\log\del{\fr{N}{2A}} + A\log\del{2A} + B \\
    \stackrel{}{\Rightarrow}\,& N \le A\del{\fr{N}{2A}-1} + A\log\del{2A} + B \\
    \stackrel{}{\Leftrightarrow}\,& N = \fr{N}{2} - A + A\log\del{2A} + B \\
    \stackrel{}{\Leftrightarrow}\,& N = \fr{N}{2} + A\log\del{\fr{2A}{e}} + B \\
    \stackrel{}{\Leftrightarrow}\,& N = 2A\log\del{\fr{2A}{e}} + 2B \\
    \stackrel{}{\Leftrightarrow}\,& N = \fr{2c}{\Delta^2}\log\del{\fr{2c}{\Delta^2e}} + \fr{2c}{\Delta^2}\log\del{K^2} \\
    \stackrel{}{\Leftrightarrow}\,& N = \fr{2c}{\Delta^2}\log\del{\fr{2cK^2}{\Delta^2e}}.
  \end{align*}
  Therefore, a sufficient condition is
  \begin{align*}
    N > \fr{2c}{\Delta^2}\log\del{\fr{2cK^2}{\Delta^2e}}. 
  \end{align*}
\end{proof}

\section{MLCB's Theorem~\ref{thm:maxlcb}}
\label{thm_proof:maxlcb}

Recall the following definitions that would be used in MLCB.

For each arm $i \in [K]$, we define 
\begin{align*}
  \gam_{i}: = \sqrt{\fr{16}{N_{i}}\log\del{\del{\fr{KT}{N_i}}^{2}}}, 
\end{align*}
as its confidence width where $T = \sum_{j \in [K]}N_j$. We define
\begin{align*}
  \hr := \argmax_{i \in [K]} \hmu_i - \gam_i
\end{align*}
as the output arm that has the highest lower confidence bound.

\newtheorem*{thm-2}{Theorem ~\ref{thm:maxlcb}}
\begin{thm-2}
  MLCB achieves
  \begin{align*}
    \MSE(\hmumlcb) 
    \stackrel{}{=}\, \tilcO \del{\fr{1}{N_1}}.
  \end{align*}
\end{thm-2}

\begin{proof}
  We have
  \begin{align*}
    &\MSE(\hmumlcb) \\
    \stackrel{}{=}\,& \EE\sbr{\del{\hmu_{\hr} - \mu_1}^2} \\
    \stackrel{}{=}\,& \int_{0}^{\infty} \PP\del{ \del{\hmu_{\hr} - \mu_1}^2 > \eps }  \dif \eps  \\
    \stackrel{}{=}\,& \int_{0}^{\infty} \PP\del{ \abs{\hmu_{\hr} - \mu_1} > \eps } \difeps \tagcmt{change of variable} \\
    \stackrel{}{=}\,& \int_{0}^{\infty} \PP\del{ \hmu_{\hr} - \mu_1 < -\eps} \difeps \\
    &+ \int_{0}^{\infty} \PP\del{ \hmu_{\hr} - \mu_1 > \eps} \difeps.
  \end{align*}
  We use claim Lemma~\ref{lemma:maxlcb_under} for the first term
  \begin{align*}
    &\int_{0}^{\infty} \PP\del{ \hmu_{\hr} - \mu_1 < -\eps} \difeps \le \fr{84}{N_1}\log\del{\fr{KT}{N_1}}.
  \end{align*}
  and use claim Lemma~\ref{lemma:maxlcb_upper} for the second term
  \begin{align*}
    &\int_0^{\infty} \PP\del{\hmu_{\hr} - \mu_1 > \eps} \difeps \le \fr{60}{N_1}\log\del{\fr{KT}{N_1}}.
  \end{align*}
  Combining the two terms concludes our proof.
\end{proof}

\begin{lemma}\label{lemma:maxlcb_under}
  MLCB achieves
  \begin{align*}
  \int_0^{\infty} \PP\del{\hmu_{\hr} - \mu_1 <- \eps} \difeps \le \fr{84}{N_1}\log\del{\fr{KT}{N_1}},
  \end{align*}
  where $\hmu_{\hr} = \max_{i \in [K]} \hmu_i - \gam_i$.
\end{lemma}

\begin{proof}
  We have
  \begin{align*}
    &\int_0^{\infty} \PP\del{\hmu_{\hr} - \mu_1 < -\eps} \difeps \\
    \stackrel{}{\le}\,& \int_0^{\infty} \PP\del{\hr = 1,\, \hmu_{\hr} - \mu_1 < -\eps} \difeps \\
    &+ \int_0^{\infty} \PP\del{\hr \ne 1,\, \hmu_{\hr} - \mu_1 < -\eps} \difeps.
  \end{align*}

  For the first term, we have
  \begin{align*}
    &\PP\del{\hr = 1,\, \hmu_{\hr} - \mu_1 < -\eps} \\
    \stackrel{}{=}\,& \PP\del{\hr = 1,\, \hmu_{1} - \mu_1 < -\eps} \\
    \stackrel{}{\le}\,& \exp\del{-\fr{1}{2}N_1\eps^2}.
  \end{align*}
  
  We use Lemma~\ref{lemma401:utility_integral_without_eps} (with $q_1 = 1$ and $z_1 = \fr{1}{2}N_1$) to bound the integral
  \begin{align*}
    & \int_0^{\infty} \PP\del{\hr = 1,\, \hmu_{\hr} - \mu_1 < -\eps} \difeps \\
    \stackrel{}{\le}\,& q_1\fr{1}{z_1} \\
    \stackrel{}{=}\,& \fr{2}{N_1}.
  \end{align*}

  For the second term, we have
  \begin{align*}
    &\PP\del{\hr \ne 1,\, \hmu_{\hr} - \mu_1 < -\eps} \\
    \stackrel{}{\le}\,& \PP\del{\hr \ne 1,\, \hmu_{\hr} - \mu_1 < -\eps} \\
    \stackrel{}{\le}\,& \PP\del{\hr \ne 1,\, \hmu_{1} - \gam_1 + \gam_{\hr} - \mu_1 < -\eps} \tagcmt{use def of $\hr$, $\hmu_{\hr} \ge \hmu_{1} - \gam_1 + \gam_{\hr}$} \\
    \stackrel{}{=}\,& \PP\del{\hr \ne 1,\, \hmu_{1} - \mu_1 < -\eps + \gam_1 - \gam_{\hr}} \\
    \stackrel{}{\le}\,& \PP\del{\hr \ne 1,\, \hmu_{1} - \mu_1 < -\eps + \gam_1} \\
    \stackrel{}{\le}\,& \PP\del{\hmu_{1} - \mu_1 < -\eps + \gam_1} \\
    \stackrel{}{\le}\,& \exp\del{-\fr{1}{2}N_1\del{\eps - \gam_1}_+^2}.
  \end{align*}

  Let $q_1 = q_2 = 1$, $z_1 = \fr{1}{8}N_1$, and $\eps_0 = 2\gam_1$. We claim that, in the regime of $\eps \ge \eps_0$, we have
  \begin{align*}
    \PP\del{\hr \ne 1,\, \hmu_{\hr} - \mu_1 < -\eps} \le q_1\exp\del{-z_1\eps^2}
  \end{align*}
  and in the regime of $\eps < \eps_0$, we have 
  \begin{align*}
    \PP\del{\hr \ne 1,\, \hmu_{\hr} - \mu_1 < -\eps} \le q_2.
  \end{align*}
  The second claim is trivial. We prove the first claim as follows. In the regime of $\eps \ge \eps_0$, we have
  \begin{align*}
    &\PP\del{\hr \ne 1,\, \hmu_{\hr} - \mu_1 < -\eps} \\
    \stackrel{}{\le}\,& \exp\del{-\fr{1}{2}N_1\del{\eps - \gam_1}_+^2} \\
    \stackrel{}{\le}\,& \exp\del{-\fr{1}{2}N_1\del{\fr{\eps}{2}}^2} \tagcmt{use $\eps \ge \eps_0$} \\
    \stackrel{}{=}\,& \exp\del{-\fr{1}{8}N_1\eps^2}.
  \end{align*}
  With the above claim, we use Lemma~\ref{lemma401:utility_integral_with_eps} to bound the integral
  \begin{align*}
    &\int_0^{\infty} \PP\del{\hr \ne 1,\, \hmu_{\hr} - \mu_1 < -\eps} \difeps \\
    \stackrel{}{\le}\,& q_2\eps_0^2 + q_1\fr{1}{z_1} \\
    \stackrel{}{=}\,& 4\gam_1^2 + \fr{8}{N_1} \\
    \stackrel{}{=}\,& 4\fr{16}{N_1}\log\del{\fr{KT}{N_1}} + \fr{8}{N_1} \\
    \stackrel{}{=}\,& \fr{64}{N_1}\log\del{\fr{KT}{N_1}} + \fr{8}{N_1}.
  \end{align*}

  Combining the two terms, we have
  \begin{align*}
    &\int_0^{\infty} \PP\del{\hmu_{\hr} - \mu_1 < -\eps} \difeps \\
    \stackrel{}{\le}\,& \int_0^{\infty} \PP\del{\hr = 1,\, \hmu_{\hr} - \mu_1 < -\eps} \difeps \\
    &+ \int_0^{\infty} \PP\del{\hr \ne 1,\, \hmu_{\hr} - \mu_1 < -\eps} \difeps \\
    \stackrel{}{\le}\,& \fr{2}{N_1} + \fr{64}{N_1}\log\del{\fr{KT}{N_1}} + \fr{8}{N_1} \\
    \stackrel{}{=}\,& \fr{2}{N_1} + \fr{32}{N_1}\log\del{\del{\fr{KT}{N_1}}^2} + \fr{8}{N_1} \\
    \stackrel{}{\le}\,& \fr{42}{N_1}\log\del{\del{\fr{KT}{N_1}}^2} \tagcmt{use $1 \le \log(K^2)$ with $K \ge 2$} \\
    \stackrel{}{=}\,& \fr{84}{N_1}\log\del{\fr{KT}{N_1}}.
  \end{align*}
  
\end{proof}

\begin{lemma}\label{lemma:maxlcb_upper}
  MLCB achieves
  \begin{align*}
  &\int_0^{\infty} \PP\del{\hmu_{\hr} - \mu_1 > \eps} \difeps \le \fr{60}{N_1}\log\del{\fr{KT}{N_1}},
  \end{align*}
  where $\hmu_{\hr} = \max_{i \in [K]} \hmu_i - \gam_i$.
\end{lemma}

\begin{proof}
  We have 
  \begin{align*}
    & \int_0^{\infty} \PP\del{\hmu_{\hr} - \mu_1 > \eps} \difeps  \\
    \stackrel{}{\le}\,& \int_0^{\infty} \PP\del{\hr = 1,\, \hmu_{1} - \mu_1 > \eps} \difeps \\
    &+ \sum_{\substack{i: i \ne 1, \\ \gam_i < 2\gam_1}} \int_0^{\infty} \PP\del{\hr = i,\, \hmu_{i} - \mu_1 > \eps} \difeps + \sum_{\substack{i: i \ne 1, \\ \gam_i > 2\gam_1}} \int_0^{\infty} \PP\del{\hr = i,\, \hmu_{i} - \mu_1 > \eps} \difeps. 
  \end{align*}

  We bound the probability and integral of each term separately as follows. For the first term, we have
  \begin{align*}
    &\PP\del{\hr = 1,\, \hmu_{\hr} - \mu_1 > \eps} \\
    \stackrel{}{=}\,& \PP\del{\hr = 1,\, \hmu_{1} - \mu_1 > \eps} \\
    \stackrel{}{\le}\,& \exp\del{-\fr{1}{2}N_1\eps^2}.
  \end{align*}

  We use Lemma~\ref{lemma401:utility_integral_without_eps} (with $q_1 = 1$ and $z_1 = \fr{1}{2}N_1$) to bound the integral
  \begin{align*}
    & \int_0^{\infty} \PP\del{\hr = 1,\, \hmu_{\hr} - \mu_1 > \eps} \difeps \le \fr{2}{N_1}.
  \end{align*}
  
  For the second term, for any $i \ne 1,\, \gam_i < 2\gam_1$ we have
  \begin{align*}
    &\sum_{\substack{i: i \ne 1, \\ \gam_i < 2\gam_1}} \PP\del{\hr = i,\, \hmu_{i} - \mu_1 > \eps} \\
    \stackrel{}{=}\,& \sum_{\substack{i: i \ne 1, \\ \gam_i < 2\gam_1}} \PP\del{\hr = i,\, \hmu_{i} - \mu_i > \eps + \Delta_i} \\
    \stackrel{}{\le}\,& \sum_{\substack{i: i \ne 1, \\ \gam_i < 2\gam_1}} \exp\del{-\fr{1}{2}N_i\del{\eps + \Delta_i}^2} \\
    \stackrel{}{\le}\,& \sum_{\substack{i: i \ne 1, \\ \gam_i < 2\gam_1}} \exp\del{-\fr{1}{2}N_i\eps^2} \\
    \stackrel{}{\le}\,& \sum_{\substack{i: i \ne 1, \\ \gam_i < 2\gam_1}} \exp\del{-\fr{1}{2}\min_{\substack{i: i \ne 1, \\ \gam_i < 2\gam_1}}N_i\eps^2} \\
    \stackrel{}{\le}\,& K\exp\del{-\fr{1}{2}\min_{\substack{i: i \ne 1, \\ \gam_i < 2\gam_1}}N_i\eps^2}.
  \end{align*}

  For this term, let $q_1 = q_2 = 1$, $z_1 = \displaystyle \fr{1}{4}\min_{\substack{i: i \ne 1, \\ \gam_i < 2\gam_1}}N_i$, and $\eps_0 = \displaystyle \sqrt{\max_{\substack{i: i \ne 1, \\ \gam_i < 2\gam_1}}\fr{4\log(K)}{N_i}}$. We claim that, in the regime of $\eps \ge \eps_0$, we have
  \begin{align*}
    \sum_{\substack{i: i \ne 1, \\ \gam_i < 2\gam_1}} \PP\del{\hr = i,\, \hmu_{i} - \mu_1 > \eps} \le \exp\del{-z_1\eps^2} 
  \end{align*}
  and, in the regime of $\eps < \eps_0$, we have
  \begin{align*}
    \sum_{\substack{i: i \ne 1, \\ \gam_i < 2\gam_1}} \PP\del{\hr = i,\, \hmu_{i} - \mu_1 > \eps} \le q_2.
  \end{align*}
  The second claim is trivial. We prove the first claim as follows. In the regime of $\eps \ge \eps_0$, we have
  \begin{align*}
    &\sum_{\substack{i: i \ne 1, \\ \gam_i < 2\gam_1}} \PP\del{\hr = i,\, \hmu_{i} - \mu_1 > \eps} \\
    \stackrel{}{\le}\,& K\exp\del{-\fr{1}{2}\min_{\substack{i: i \ne 1, \\ \gam_i < 2\gam_1}}N_i\eps^2} \\
    \stackrel{}{\le}\,& \exp\del{-\fr{1}{2}\min_{\substack{i: i \ne 1, \\ \gam_i < 2\gam_1}}N_i\eps^2 + \log(K)} \\
    \stackrel{}{\le}\,& \exp\del{-\fr{1}{4}\min_{\substack{i: i \ne 1, \\ \gam_i < 2\gam_1}}N_i\eps^2} \tagcmt{use $\eps \ge \eps_0$}.
  \end{align*}

  With the above claim, we use Lemma~\ref{lemma401:utility_integral_with_eps} to bound the integral
  \begin{align*}
    &\int_{0}^{\infty} \sum_{\substack{i: i \ne 1, \\ \gam_i < 2\gam_1}} \PP\del{\hr = i,\, \hmu_{i} - \mu_1 > \eps} \difeps \\
    \stackrel{}{\le}\,& q_2\eps_0^2 + q_1\fr{1}{z_1} \\
    \stackrel{}{=}\,& \max_{\substack{i: i \ne 1, \\ \gam_i < 2\gam_1}}\fr{4\log(K)}{N_i} + \max_{\substack{i: i \ne 1, \\ \gam_i < 2\gam_1}} \fr{4}{N_i} \\
    \stackrel{}{=}\,& \max_{\substack{i: i \ne 1, \\ \gam_i < 2\gam_1}}\fr{2\log(K^2)}{N_i} + \max_{\substack{i: i \ne 1, \\ \gam_i < 2\gam_1}} \fr{4}{N_i} \\
    \stackrel{}{\le}\,& \max_{\substack{i: i \ne 1, \\ \gam_i < 2\gam_1}}\fr{6\log(K^2)}{N_i} \tagcmt{use $1 \le \log(K^2)$ with $K \ge 2$} \\
    \stackrel{}{=}\,& \max_{\substack{i: i \ne 1, \\ \gam_i < 2\gam_1}}\fr{12\log(K)}{N_i}.
  \end{align*}
  
  Lemma~\ref{lemma:maxlcb_101} states that for any $i \ne 1$ the condition $\gam_i < 2\gam_1$ implies that $N_i > \fr{N_1}{4} \fr{\log\del{K}}{\log\del{\fr{KT}{N_1}}}$. Therefore,
  \begin{align*}
    &\max_{\substack{i: i \ne 1, \\ \gam_i < 2\gam_1}}\fr{12\log(K)}{N_i} \\
    \stackrel{}{\le}\,& \max_{\substack{i: i \ne 1, \\ \gam_i < 2\gam_1}}\fr{12\log(K)}{N_1}\fr{4\log\del{\fr{KT}{N_1}}}{\log\del{K}} \\
    \stackrel{}{=}\,& \fr{48}{N_1}\log\del{\fr{KT}{N_1}}.
  \end{align*}

  From this point, let $i$ be an arm such that $i \ne 1,\, \gam_i > 2\gam_1$.
  For the third term, we have
  \begin{align*}
    &\PP\del{\hr = i,\, \hmu_{i} - \mu_1 > \eps} \\
    \stackrel{}{=}\,& \PP\del{\hr = i,\, \hmu_{i} - \mu_i > \eps + \Delta_i} \\
    \stackrel{}{\le}\,& \exp\del{-\fr{1}{2}N_i\del{\eps + \Delta_i}^2} \\
    \stackrel{}{\le}\,& \exp\del{-\fr{1}{2}N_i\eps^2}.
  \end{align*}

  In addition, we have
  \begin{align*}
    &\PP\del{\hr = i} \\
    \stackrel{}{\le}\,& \PP\del{\hr = i,\, \hmu_i - \gam_i > \hmu_1 - \gam_1} \\
    \stackrel{}{=}\,& \PP\del{\hr = i,\, \hmu_i - \mu_i - \hmu_1 + \mu_1 > \gam_i - \gam_1 + \Delta_i} \\
    \stackrel{}{\le}\,& \PP\del{\hr = i,\, \hmu_i - \mu_i - \hmu_1 + \mu_1 > \fr{1}{2}\gam_i + \Delta_i} \tagcmt{use $\gam_i > 2\gam_1$} \\
    \stackrel{}{\le}\,& \exp\del{-\fr{1}{2}\fr{1}{\del{\fr{1}{N_i} + \fr{1}{N_1}}}\del{\fr{1}{2}\gam_i + \Delta_i}^2} \\
    \stackrel{}{\le}\,& \exp\del{-\fr{1}{8}\fr{1}{\del{\fr{1}{N_i} + \fr{1}{N_1}}}\gam_1^2}. 
  \end{align*}
  
  Lemma~\ref{lemma:maxlcb_102} states that for any $i \ne 1$ the condition $\gam_i > 2\gam_1$ implies that $N_i < N_1$. Therefore,
  \begin{align*}
    &\exp\del{-\fr{1}{8}\fr{1}{\del{\fr{1}{N_i} + \fr{1}{N_1}}}\gam_i^2} \\
    \stackrel{}{\le}\,& \exp\del{-\fr{1}{8}\fr{1}{\del{\fr{1}{N_i} + \fr{1}{N_i}}}\gam_i^2} \tagcmt{use $N_i < N_1$} \\
    \stackrel{}{=}\,& \exp\del{-\fr{1}{16}N_i\gam_i^2} \\
    \stackrel{}{=}\,& \exp\del{-\fr{1}{16}N_i\fr{16}{N_i}\log\del{\del{\fr{KT}{N_i}}^2}} \\
    \stackrel{}{=}\,& \del{\fr{N_i}{KT}}^2.
  \end{align*}

  
  

  Combining the two inequalities, we have
  \begin{align*}
    & \PP\del{\hr = i,\, \hmu_{i} - \mu_1 > \eps} \\
    \stackrel{}{\le}\,& \del{\fr{N_i}{KT}}^2 \wedge \exp\del{-\fr{1}{2}N_i\eps^2} \\
    \stackrel{}{\le}\,& \sqrt{\del{\fr{N_i}{KT}}^2\exp\del{-\fr{1}{2}N_i\eps^2}} \\
    \stackrel{}{=}\,& \fr{N_i}{KT} \exp\del{-\fr{1}{4}N_i\eps^2}. 
  \end{align*}

  We use Lemma~\ref{lemma401:utility_integral_without_eps} (with $q_1 = \fr{N_i}{KT}$ and $z_1 = \fr{1}{4}N_i$) to bound the integral
  \begin{align*}
    &\int_{0}^{\infty} \PP\del{\hr = i,\, \hmu_{i} - \mu_1 > \eps} \difeps \\
    \stackrel{}{\le}\,& q\fr{1}{z_1} \\
    \stackrel{}{\le}\,& \fr{N_i}{KT} \fr{4}{N_i} \\
    \stackrel{}{\le}\,& \fr{4}{KT}.
  \end{align*}
  
  Therefore, for the third term, we obtain the integral
  \begin{align*}
    &\sum_{\substack{i: i \ne 1, \\ \gam_i > 2\gam_1}} \int_{0}^{\infty} \PP\del{\hr = i,\, \hmu_{i} - \mu_1 > \eps} \difeps \\
    \stackrel{}{\le}\,& \sum_{\substack{i: i \ne 1, \\ \gam_i > 2\gam_1}} \fr{4}{K\sum_{j \in [K]}N_j} \\
    \stackrel{}{\le}\,& K\fr{4}{KT} \\
    \stackrel{}{=}\,& \fr{4}{T}.
  \end{align*}

  Combining all the terms, we have
  \begin{align*}
    & \int_0^{\infty} \PP\del{\hmu_{\hr} - \mu_1 > \eps} \difeps  \\
    \stackrel{}{\le}\,& \int_0^{\infty} \PP\del{\hr = 1,\, \hmu_{1} - \mu_1 > \eps} \difeps \\
    &+ \sum_{\substack{i: i \ne 1, \\ \gam_i < 2\gam_1}} \int_0^{\infty} \PP\del{\hr = i,\, \hmu_{i} - \mu_1 > \eps} \difeps + \sum_{\substack{i: i \ne 1, \\ \gam_i > 2\gam_1}} \int_0^{\infty} \PP\del{\hr \ne i,\, \hmu_{i} - \mu_1 > \eps} \difeps \\
    \stackrel{}{\le}\,& \fr{2}{N_1} + \fr{48}{N_1}\log\del{\fr{KT}{N_1}} + \fr{4}{S} \\
    \stackrel{}{=}\,& \fr{2}{N_1} + \fr{48}{N_1}\log\del{\fr{KT}{N_1}} + \fr{4}{\sum_{j \in [K]}N_j} \tagcmt{use $T = \sum_{j \in [K]}N_j$} \\
    \stackrel{}{=}\,& \fr{2}{N_1} + \fr{24}{N_1}\log\del{\del{\fr{KT}{N_1}}^2} + \fr{4}{\sum_{j \in [K]}N_j} \\
    \stackrel{}{\le}\,& \fr{30}{N_1}\log\del{\del{\fr{KT}{N_1}}^2} \tagcmt{use $1 \le \log(K^2)$ with $K \ge 2$} \\
    \stackrel{}{=}\,& \fr{60}{N_1}\log\del{\fr{KT}{N_1}}. 
  \end{align*}

\end{proof}

\begin{lemma}\label{lemma:maxlcb_101}
  Let $\gam_i = \sqrt{\fr{a}{N_i}\log\del{\del{\fr{KT}{N_i}}^{s}}}$ where $T \ge N_i$ and $s \ge 1$. For any $i \ne 1$, the condition $\gam_i < b\gam_1$ implies that $N_i > \fr{N_1}{b^2} \fr{\log\del{K}}{\log\del{\fr{KT}{N_1}}}$.
\end{lemma}

\begin{proof}
  
  
  We have
  \begin{align*}
    &\gam_i < b\gam_1 \\
    \stackrel{}{\Leftrightarrow}\,& \fr{a}{N_i}\log\del{\del{\fr{KT}{N_i}}^{s}} < b^2\fr{a}{N_1}\log\del{\del{\fr{KT}{N_1}}^{s}} \\
    \stackrel{}{\Leftrightarrow}\,& \fr{1}{N_i}\log\del{\fr{KT}{N_i}} < \fr{b^2}{N_1}\log\del{\fr{KT}{N_1}} \\
    \stackrel{}{\Leftrightarrow}\,& N_i > \fr{N_1}{b^2} \fr{\log\del{\fr{KT}{N_i}}}{\log\del{\fr{KT}{N_1}}} \\
    \stackrel{}{\Rightarrow}\,& N_i > \fr{N_1}{b^2} \fr{\log\del{K}}{\log\del{\fr{KT}{N_1}}} \tagcmt{use $N_i \le T$}.
  \end{align*}
\end{proof}

\begin{lemma}\label{lemma:maxlcb_102}
  Let $\gam_i = \sqrt{\fr{a}{N_i}\log\del{\del{\fr{KT}{N_i}}^{s}}}$ where $T \ge N_i$ and $s \ge 1$. For $b \ge 2$, for any $i \ne 1$, the condition $\gam_i > b\gam_1$ implies that $N_i < N_1$.
\end{lemma}

\begin{proof}
  
  
  We have
  \begin{align*}
    &\gam_i > b\gam_1 \\
    \stackrel{}{\Leftrightarrow}\,& \fr{a}{N_i}\log\del{\del{\fr{KT}{N_i}}^s} > b^2\fr{a}{N_1}\log\del{\del{\fr{KT}{N_1}}^s} \\
    \stackrel{}{\Leftrightarrow}\,& \fr{1}{N_i}\log\del{\fr{KT}{N_i}} > \fr{b^2}{N_1}\log\del{\fr{KT}{N_1}} \\
    \stackrel{}{\Leftrightarrow}\,& N_i < \fr{N_1}{b^2} \fr{\log\del{\fr{KT}{N_i}}}{\log\del{\fr{KT}{N_1}}}.
  \end{align*}

  For any $c \ge 0$, there are two cases: (1) $N_i \ge c$ and (2) $N_i < c$.

  The case $N_i \ge c$ implies that
  \begin{align*}
    N_i < \fr{N_1}{b^2} \fr{\log\del{\fr{KT}{N_i}}}{\log\del{\fr{KT}{N_1}}} < \fr{N_1}{b^2} \fr{\log\del{\fr{KT}{c}}}{\log\del{\fr{KT}{N_1}}}.
  \end{align*}

  Therefore, for any $c \ge 0$, we have
  \begin{align*}
    N_i < \fr{N_1}{b^2} \fr{\log\del{\fr{KT}{c}}}{\log\del{\fr{KT}{N_1}}} \vee c.
  \end{align*}

  Choosing $c = N_1$, we have
  \begin{align*}
    N_i &< \fr{N_1}{b^2} \fr{\log\del{\fr{KT}{N_1}}}{\log\del{\fr{KT}{N_1}}} \vee N_1 < \fr{N_1}{b^2} \vee N_1 \le N_1. 
  \end{align*}
  
\end{proof}


\section{LEM's Lemma ~\ref{thm:lem_bound}}
\label{thm_proof:lem_bound}

\begin{lemma}\label{lemma:lem_lower_bound_twoarm_equalmean}
  Assuming the arms follow Gaussian distributions, in two-arm instances with equal means (i.e., $\mu_1 = \mu_2$), there exists an absolute constant $c_1$ such that 
  \begin{align*}
    \MSE(\hmume) \ge c_1\del{\fr{1}{\min_{i \in [2]}N_{i}}}~.
  \end{align*}
\end{lemma}

\begin{proof}
  The proof naturally follows Lemma~\ref{lemma:lem_lower_bound_twoarm} with $\Delta_2 = 0$.

\end{proof}

\begin{lemma}\label{lemma:lem_lower_bound_twoarm}
  Assuming the arms follow Gaussian distributions, in two-arm instances, there exists an absolute constant $c_1$ such that 
  \begin{align*}
    \MSE(\hmume) \ge c_1\del{\fr{1}{N_{1}} + \fr{1}{N_{2}}\exp\del{-14N_2\Delta_2^2}} ~.
  \end{align*}
\end{lemma}

\begin{proof}
  We denote by $\ha = \argmax_{i \in [2]}\hmu_i$ the arm with the larger empirical mean between the two arms.
  
  We decompose the MSE as follows
  \begin{align*}
    &\MSE(\hmume) \\
    \stackrel{}{=}\,&\EE\sbr{\del{\hmu_{\ha} - \mu_1}^2} \\
    \stackrel{}{=}\,&\int_{0}^{\infty} \PP\del{\del{\hmu_{\ha} - \mu_1}^2 > \eps^2 } \dif \eps \\
    \stackrel{}{=}\,&\int_{0}^{\infty} \PP\del{\abs{\hmu_{\ha} - \mu_1} > \eps } \difeps \tagcmt{change of variable} \\
    \stackrel{}{=}\,&\int_{0}^{\infty} \del{\PP\del{\hmu_{\ha} - \mu_1 > \eps } + \PP\del{\hmu_{\ha} - \mu_1 < -\eps } }\difeps \\
    \stackrel{}{\ge}\,&\int_{0}^{\infty} \PP\del{\hmu_{\ha} - \mu_1 > \eps } \difeps \\
    \stackrel{}{=}\,& \int_{0}^{\infty} \PP\del{\ha = 1,\, \muhat_1 - \mu_1 > \eps} \difeps + \int_{0}^{\infty} \PP\del{\ha = 2,\, \muhat_2 - \mu_1 > \eps} \difeps.
  \end{align*}

  We bound the first probability as follows. We have 
  \begin{align*}
    &\PP\del{\ha = 1,\, \muhat_1 - \mu_1 > \eps} \\
    \stackrel{}{=}\,&\PP\del{\muhat_1 - \muhat_2 \ge 0,\, \muhat_1 - \mu_1 > \eps} \tagcmt{use $\hmu_{1} = \hmu_{\ha} \ge \hmu_{2}$} \\
    \stackrel{}{=}\,&\PP\del{\muhat_1 - \mu_1 - \muhat_2 + \mu_2 \ge -\Delta_2,\, \muhat_1 - \mu_1 > \eps}.
  \end{align*}

  Let $X_1 = \hmu_1 - \mu_1$ and $X_2 = \muhat_2 - \mu_2$, we further have
  \begin{align*}
    &\PP\del{X_1 - X_2 \ge -\Delta_2,\, X_1 > \eps} \\
    \stackrel{}{=}\,& \int_{x_1,x_2} \idt{x_1 - x_2 \ge -\Delta_2,\, x_1 > \eps}\dif \PP(x_2) \dif \PP(x_1) \\
    \stackrel{}{=}\,& \int_{x_1=\eps}^{\infty} \int_{x_2=-\infty}^{x_1 + \Delta_2} 1 \dif \PP(x_1)\dif \PP(x_2) \\
    \stackrel{}{\ge}\,& \int_{x_1=\eps}^{\infty} \int_{x_2=-\infty}^{0} 1 \dif \PP(x_1)\dif \PP(x_2) \tagcmt{since $x_1 + \Delta_2 \ge \eps + \Delta_2 \ge 0$} \\
    \stackrel{}{=}\,&\int_{x_1=\eps}^{\infty} \fr{1}{2} \dif \PP(x_2) \tagcmt{$x_1$ follows Gaussian distribution with zero mean} \\
    \stackrel{}{=}\,& \fr{1}{2}\PP\del{X_1 \ge \eps} \\
    \stackrel{}{=}\,& \fr{1}{2}\PP\del{\hmu_1 -  \mu_1 \ge \eps} \\
    \stackrel{}{\ge}\,& \fr{1}{2}\fr{1}{4\sqrt{\pi}}\exp\del{-\fr{7}{2}N_1\eps^2}  \tagcmt{use Lemma ~\ref{xlemma:anti_concentration_inequality} with variance $\sigma^2 = \fr{1}{N_1}$} \\
    \stackrel{}{=}\,& \fr{1}{8\sqrt{\pi}}\exp\del{-\fr{7}{2}N_{1}\eps^2}.
  \end{align*}

  Thus, we have
  \begin{align*}
    &\int_{0}^{\infty} \PP\del{\ha = 1,\, \muhat_1 - \mu_1 > \eps} \difeps \\
    \stackrel{}{\ge}\,&  \int_{0}^{\infty} \fr{1}{8\sqrt{\pi}} \exp\del{-\fr{7}{2}N_{1}\eps^2} \difeps \\
    \stackrel{}{=}\,& \fr{1}{8\sqrt{\pi}} \del{- \fr{2}{7N_{1}}\exp\del{-\fr{7}{2}N_{1}\eps^2}}\Big|_{0}^{\infty} \\
    \stackrel{}{=}\,& \fr{2}{56\sqrt{\pi}} \fr{1}{N_{1}}.
  \end{align*}

  Next, we bound the second probability term,
  \begin{align*}
    &\PP\del{\ha = 2,\, \muhat_{2} - \mu_1 > \eps} \\
    \stackrel{}{=}\,& \PP\del{\muhat_{2} - \muhat_1 \ge 0,\, \muhat_{2} - \mu_{1} > \eps } \tagcmt{use $\hmu_{2} = \hmu_{\ha} \ge \hmu_{1}$} \\
    \stackrel{}{=}\,& \PP\del{\muhat_{2} - \mu_{2} - \muhat_1 + \mu_1 \ge \Delta_{2},\, \muhat_{2} - \mu_{2} > \eps + \Delta_{2}}. 
  \end{align*}

  Let $X_{2} = \hmu_{2} - \mu_{2}$ and $X_1 = \muhat_1 - \mu_1$, we further have
  \begin{align*}
    &\PP\del{X_{2} - X_1 \ge \Delta_{2},\, X_{2} > \eps + \Delta_{2}} \\
    \stackrel{}{=}\,& \int_{x_1,x_{2}} \idt{x_{2} - x_1 \ge \Delta_{2},\, x_{2} > \eps + \Delta_{2}}\dif \PP(x_1)\dif \PP(x_{2}) \\
    \stackrel{}{=}\,& \int_{x_{2}=\eps+\Delta_{2}}^{\infty} \int_{x_1=-\infty}^{x_{2} - \Delta_{2}} 1 \dif \PP(x_1)\dif \PP(x_{2}) \\
    \stackrel{}{\ge}\,& \int_{x_{2}=\eps+\Delta_{2}}^{\infty} \int_{x_1=-\infty}^{0} 1 \dif \PP(x_1)\dif \PP(x_{2}) \tagcmt{since $x_{2} - \Delta_{2} \ge \eps + \Delta_{2} - \Delta_{2} \ge 0$} \\
    \stackrel{}{=}\,& \int_{x_{2}=\eps+\Delta_{2}}^{\infty} \fr{1}{2} \dif \PP(x_{2}) \tagcmt{$x_{2}$ follows Gaussian distribution with zero mean} \\
    \stackrel{}{=}\,& \fr{1}{2}\PP\del{X_{2} > \eps + \Delta_{2}} \\
    \stackrel{}{=}\,& \fr{1}{2}\PP\del{\hmu_{2} - \mu_{2} > \eps + \Delta_{2}} \\
    \stackrel{}{\ge}\,& \fr{1}{2}\fr{1}{4\sqrt{\pi}}\exp\del{-\fr{7}{2}N_{2}\del{\eps+\Delta_2}^2} \tagcmt{use Lemma ~\ref{xlemma:anti_concentration_inequality} with variance $\sigma^2 = \fr{1}{N_{2}}$} \\
    \stackrel{}{=}\,& \fr{1}{8\sqrt{\pi}}\exp\del{-\fr{7}{2}N_{2}\del{\eps + \Delta_2}^2}.
  \end{align*}

  In the regime of $\eps \ge \Delta_2$, we have
  \begin{align*}
    &\PP\del{\ha = 2,\, \muhat_{2} - \mu_1 > \eps} \\
    \stackrel{}{\ge}\,& \fr{1}{8\sqrt{\pi}}\exp\del{-\fr{7}{2}N_{2}\del{\eps + \Delta_2}^2} \\
    \stackrel{}{\ge}\,& \fr{1}{8\sqrt{\pi}}\exp\del{-\fr{7}{2}N_{2}\del{2\eps}^2} \tagcmt{use $\eps \ge \Delta_2$} \\
    \stackrel{}{=}\,& \fr{1}{8\sqrt{\pi}}\exp\del{-14N_{2}\eps^2}.
  \end{align*}

  Thus, we have
  \begin{align*}
    &\int_{0}^{\infty} \PP\del{\ha = 2,\, \muhat_{2} - \mu_1 > \eps} \difeps \\
    \stackrel{}{=}\,& \int_{0}^{\Delta_2} \PP\del{\ha = 2,\, \muhat_{2} - \mu_1 > \eps} \difeps + \int_{\Delta_2}^{\infty} \PP\del{\ha = 2,\, \muhat_{2} - \mu_1 > \eps} \difeps \\
    \stackrel{}{\ge}\,& \int_{\Delta_2}^{\infty} \PP\del{\ha = 2,\, \muhat_{2} - \mu_1 > \eps} \difeps \\                   
    \stackrel{}{\ge}\,& \int_{\Delta_2}^{\infty} \fr{1}{8\sqrt{\pi}} \exp\del{-14N_{2}\eps^2} \difeps \\
    \stackrel{}{=}\,& \fr{1}{8\sqrt{\pi}} \del{- \fr{1}{14N_{2}}\exp\del{-14N_{2}\eps^2}}\Big|_{\Delta_2}^{\infty} \\
    \stackrel{}{=}\,& \fr{2}{112\sqrt{\pi}} \fr{1}{N_{2}}\exp\del{-14N_2\Delta_2^2}.
  \end{align*}

  Combining the results, we have
  \begin{align*}
    &\int_{0}^{\infty} \PP\del{\hmu_{\ha} - \mu_1 > \eps } \difeps \\
    \stackrel{}{=}\,& \int_{0}^{\infty} \PP\del{\ha = 1,\, \muhat_1 - \mu_1 > \eps} \difeps + \int_{0}^{\infty} \PP\del{\ha = 2,\, \muhat_2 - \mu_1 > \eps} \difeps \\
    \stackrel{}{\ge}\,& \fr{2}{56\sqrt{\pi}} \fr{1}{N_{1}} + \fr{2}{112\sqrt{\pi}} \fr{1}{N_{2}}\exp\del{-14N_2\Delta_2^2}.
  \end{align*}
  which concludes our proof. 
\end{proof}

\begin{lemma}[Stirling's formula]
  \label{xlemma:stirling_formula}
  Let $k$ and $n$ be two positive integers such that $1 \le k \le n$. Then,
  \begin{align*}
    \del{\fr{n}{k}}^k \le \binom{n}{k} \le \del{\fr{en}{k}}^k.
  \end{align*}
\end{lemma}

\begin{lemma}[Anti-concentration inequality \citep{abramowitz1968handbook}]
  \label{xlemma:anti_concentration_inequality}
  For a Gaussian random variable $X \sim \mathcal{N}(\mu, \sigma^2)$ and any $\eps > 0$, we have
  \begin{align*}
    \PP\del{\abs{X - \mu} > \eps} > \fr{1}{4\sqrt{\pi}}\ep{-\fr{7\eps^2}{2\sigma^2}}.
  \end{align*}
\end{lemma}

\end{document}